\documentclass[]{cit_lab_mfr}

\usepackage{hyperref}
\usepackage{cleveref}
\usepackage{verbatim}

\usepackage{wrapfig}  
\usepackage{graphicx}
\usepackage{subcaption}
\usepackage{listings}
\usepackage{algorithm}
\usepackage{bm}
\usepackage{colortbl}
\usepackage{tocloft} 

\usepackage{xcolor}
\usepackage[dvipsnames]{xcolor}
\usepackage[table]{xcolor}
\usepackage{multirow}
\usepackage{booktabs}
\usepackage{adjustbox}
\usepackage{tabularx}   
\usepackage{siunitx}    
\usepackage{enumitem}

\newcommand{\textBlue}[1]{\textcolor{NavyBlue}{#1}}
\newcommand{\textRed}[1]{\textcolor{BrickRed}{#1}}

\usepackage{subcaption} %

\usepackage[toc,page,header]{appendix}

\usepackage{minitoc}

\title{Forging a Dynamic Memory: Retrieval-Guided Continual Learning for Generalist Medical Foundation Models}

\author{%
\parbox{\textwidth}{\centering
Zizhi Chen$^{1,2,*}$, Yizhen Gao$^{3,*}$, Minghao Han$^{1,2}$, Yizhou Liu$^{1,2}$, Zhaoyu Chen$^{1}$, \\
Dingkang Yang$^{1,2,\dagger,\S}$, Lihua Zhang$^{1,2,\S}$
}}

\affiliation{%
\parbox{\textwidth}{\centering\small
$^1$College of Intelligent Robotics and Advanced Manufacturing, Fudan University\\[1mm]
$^2$Fysics Intelligence Technologies Co., Ltd. (Fysics AI)\\[1mm]
$^3$School of Computer Science and Engineering , Central South University\\[1mm]

}}

\contribution[*]{Equal contribution}
\contribution[\dagger]{Team lead}
\contribution[\S]{Corresponding author}

\abstract{
Multimodal biomedical Vision-Language Models (VLMs) exhibit immense potential in the field of Continual Learning (CL). However, they confront a core dilemma: how to preserve fine-grained intra-modality features while bridging the significant domain gap across different modalities. To address this challenge, we propose a comprehensive framework. Leveraging our 18-million multimodal and comprehensive medical retrieval database derived from PubMed scientific papers, we pioneer the integration of Retrieval-Augmented Generation (RAG) into CL. Specifically, we employ a multi-modal, multi-layer RAG system that provides real-time guidance for model fine-tuning through dynamic, on-demand knowledge retrieval. Building upon this, we introduce a dynamic knowledge distillation framework. This framework precisely resolves the aforementioned core dilemma by dynamically modulating the importance of the parameter space, the granularity of the distilled knowledge, and the data distribution of the reference dataset in accordance with the required level of detail. To thoroughly validate the clinical value of our strategy, we have designed a more rigorous \textbf{M}edical \textbf{G}eneralist \textbf{T}ask \textbf{I}ncremental \textbf{L}earning \textbf{(MGTIL)} benchmark. This benchmark is engineered to simultaneously evaluate the model's capacity for adaptation to significant domain shifts, retention of subtle intra-domain features, and real-time learning of novel and complex medical tasks. Extensive experimental results demonstrate that our proposed method achieves state-of-the-art (SOTA) performance across all metrics. The code is provided in the supplementary materials.
}
\date{\today}
\checkdata[Corresponding]{\url{dicken@fysics.ai,chenzz24@m.fudan.edu.cn}}
\checkdata[Project Page]{\url{https://github.com/CZZZZZZZZZZZZZZZZZ/PRIMED}}

\begin{document}
\maketitle


\addtocontents{toc}{\protect\setcounter{tocdepth}{-1}}

\section{Introduction}
\label{sec:intro}

Vision-Language Models (VLMs) such as CLIP~\cite{CLIP} have succeeded in zero-shot and fine-tuning. In medicine, PMC-CLIP~\cite{Pmc-clip}, trained on million level image-text pairs from scientific articles, created the first Generalist Medical Foundation Model, enabling multimodal AI to aid clinical diagnosis. Yet updating pre-trained VLMs and applying them to many downstream tasks across diverse medical domains and difficulty levels is hard. Retraining from scratch or keeping a separate fine-tuned model per task is often impractical due to onerous data curation and consolidation, scarce data for many diseases, and ethical limits. These approaches also incur prohibitive compute and fit poorly with today’s medical data ecosystem. Continual Learning (CL)~\cite{lwf,icarl,zscl,MOE_CL}, which learns incrementally from task sequences, offers a viable way to handle these constraints.

\begin{wrapfigure}{r}{0.5\textwidth}
  \vspace{-10pt} 
  \begin{center}
    \includegraphics[width=1\linewidth]{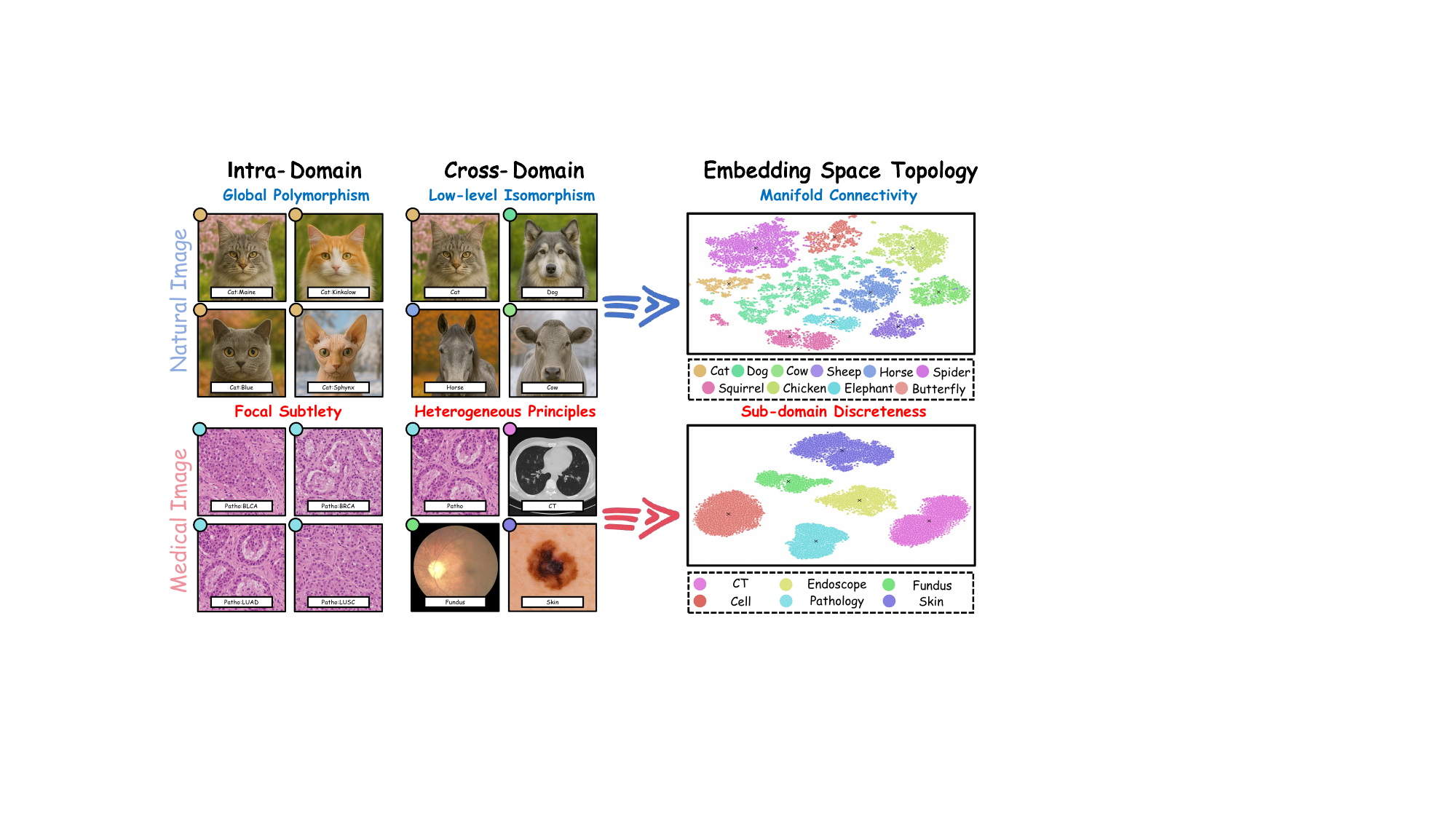}
  \end{center}
  \caption{\textbf{\textcolor{blue}{Natural}}\:\textit{vs.}\:\textbf{\textcolor{red}{Medical}}: Natural images are intra-domain diverse yet cross-domain similar. Medical images are intra-domain uniform but cross-domain distinct.}
  \label{fig1}
  \vspace{-10pt}
\end{wrapfigure}

However, CL in medical scenarios presents a more complex situation. As shown in Fig.~\ref{fig1}, natural images and medical images are fundamentally different. In the natural image, intra-domain samples exhibit significant morphological variability, while inter-domain samples still share correlations at the level of low-level features. Consequently, the overall data distribution tends to form a mesh-like structure. In contrast, medical images display minimal intra-domain variation but are governed by drastically different imaging principles across domains, resulting in a clustered distribution. This implies that in medical CL, we must simultaneously achieve finer-grained intra-domain transfer and bridge a more challenging inter-domain gap.

In addition, CL can lead to \textit{catastrophic forgetting}~\cite{Der,Dualprompt,BiC}. For continuously fine-tuned VLMs, this manifests not only as forgetting previously learned downstream tasks but also as degradation of pretrained knowledge, which significantly impairs their zero-shot capabilities. A straightforward approach to mitigate forgetting is to replay a small subset of stored historical samples~\cite{icarl}. However, since pretrained data are typically inaccessible, this strategy is impractical for VLMs. Consequently, CL for VLMs is often simulated by sampling~\cite{zscl,snd,gnsp} or synthesizing data~\cite{ddgr,diffclass,GIFT} based on ImageNet~\cite{imagenet} labels to mimic the replay process. Nevertheless, constructing reference datasets in a label-based manner is not straightforward for medical VLMs and their large-scale ecosystems, built around contextual or caption-based supervision.  Meanwhile, Retrieval-Augmented Generation (RAG)~\cite{RAG} has become an essential component in modern LLM systems, improving reasoning performance by incorporating retrieved information through keyword matching~\cite{BM25} or embedding-based similarity search~\cite{bge,qwen3,SAIL-Embedding}. This insight inspires a new approach: we can construct reference datasets by generating pseudo-labeled content in bulk, using questions as carriers and employing a RAG-based framework to enable contextual reference during learning. More importantly, compared with previous approaches that rely on sampling or generation, RAG enables more fine-grained and dynamic querying. It not only better captures the clustered distribution patterns of large-scale medical data but also dynamically adapts to domain-internal shifts, cross-domain transitions, and complex scenarios involving difficult tasks.


Based on this, we propose the \textbf{P}recision \textbf{R}etrieval-\textbf{I}nfused model for \textbf{MED}ical \textbf{(PRIMED)} framework. First, we construct an \textbf{18-million} multimodal medical image retrieval database from PubMed scientific papers using a data-cleaning pipeline and the Qwen3-embedding-8B model~\cite{qwen3}. To achieve intra-modal fine-grained feature memory distillation, we collect and organize approximately \textbf{3,000} general medical label entries as question pool, hierarchically refined from perspectives such as domain, lesion location, and disease type, enabling precise disentanglement of clustered medical knowledge across domains. During training, we introduce a dynamic distribution regulator that shifts the reference dataset from uniform to real-time multimodal, updating the RAG query vector database’s content and domain mix based on past and ongoing fine-tuning to create a personalized “review plan” that enables fine-grained intra-domain specialization and precise inter-domain recall. To preserve the global structure of the embedding space and maintain the model’s zero-shot capability, we introduce \textbf{C}ontrastive \textbf{K}nowledge \textbf{T}ransfer \textbf{(CKT)} and \textbf{C}ross-\textbf{M}odality \textbf{C}onsistency \textbf{(CMC)} loss strategies. Finally, by applying a dynamic model weight importance feedback, we achieve compatibility between global structural preservation and fine-grained personalized feature retention. To better evaluate our approach, we propose \textbf{M}edical \textbf{G}eneralist \textbf{T}ask \textbf{I}ncremental \textbf{l}earning \textbf{(MGTIL)} benchmark. Previous medical CL studies have focused always on single-domain~\cite{conslide,QPMIL,CSSL} or support for base model training~\cite{biomedica,MedCoss}, lacking comprehensive cross-domain benchmarks. We design two evaluations: \textbf{HieraMedTransfer}, simultaneously simulating intra and inter domain incremental learning across diverse tasks; and \textbf{MedXtreme}, evaluating the model’s continual memory after fine-tuning on ultra-challenging medical classification tasks. Our contributions are summarized as follows:
\begingroup
\setlist[itemize]{noitemsep, topsep=0pt, left=0pt}
\begin{itemize}
    \item We present an 18M multimodal retrieval database and a 3,000 fine-grained medical question pool that enable real-time fine-grained retrieval.
    \item We propose MGTIL, a comprehensive continual learning benchmark for the medical general domain, covering numerous datasets across diverse fields. It supports three evaluation scenarios: intra-domain transfer, inter-domain transfer, and high-difficulty task retention.
    \item We propose PRIMED, which dynamically adjusts the reference-data distribution and knowledge ratio to predict model-weight importance in real time for complex medical continual learning, preserving fine-grained intra-modal features and bridging significant inter-modal domain gaps. It achieves state-of-the-art performance across all evaluations on MGTIL.  
\end{itemize}
\endgroup

\section{Related Work}
\label{sec:related_work}

\begin{figure*}[tp]
 \includegraphics[width=1\linewidth]{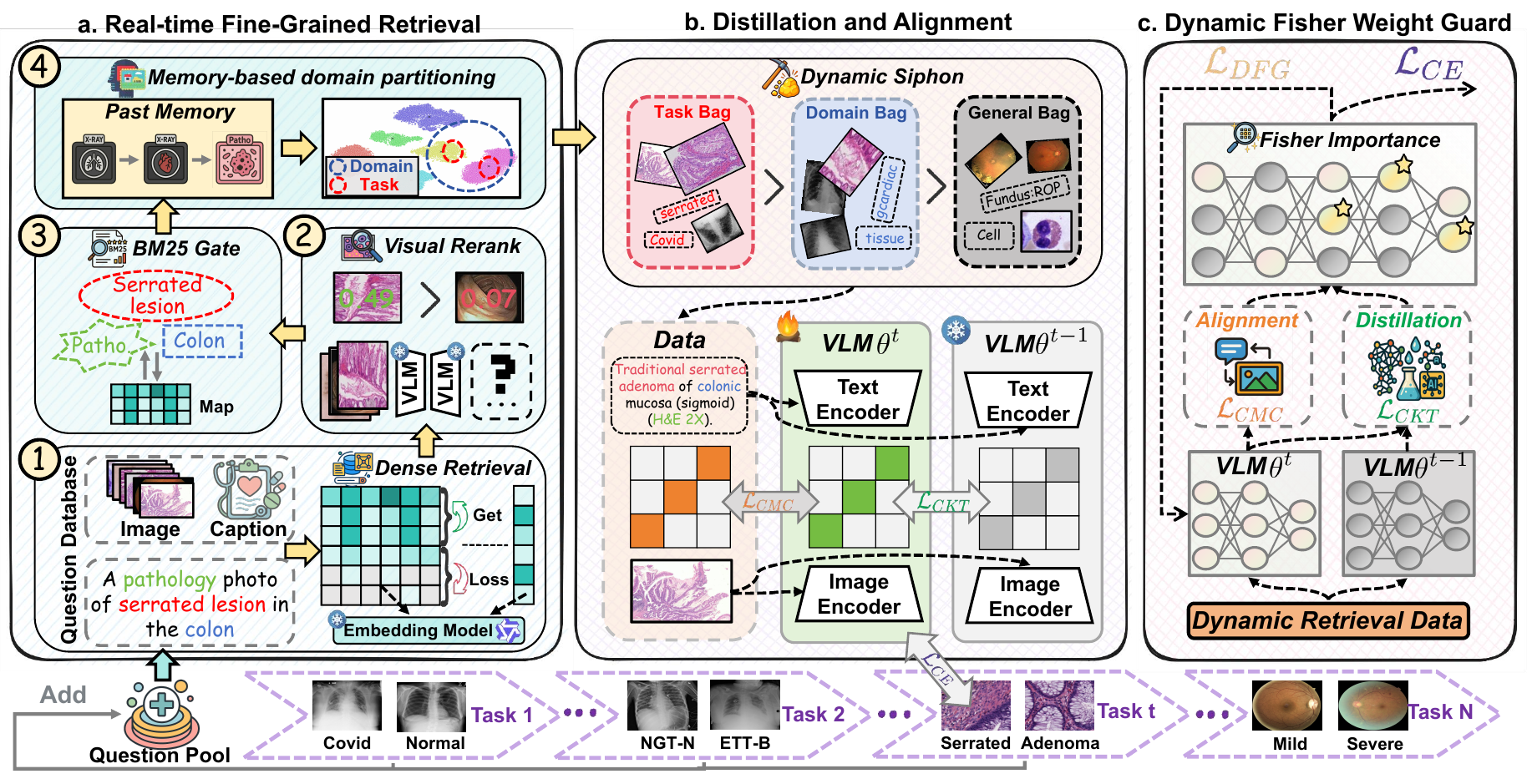}
\vspace{-1.2em}
\caption{Framework overview of PRIMED. (a) \textbf{Real-time Fine-Grained Retrieval}. Perform multi-level retrieval between the real-time query repository and the pre-constructed 18-million multimodal retrieval database, and obtain specialized image-caption pairs through a Dynamic Siphon mechanism. (b) \textbf{Distillation and Alignment}. Contrastive Knowledge Transfer $\mathcal{L}_{CKT}$ is used for matching
CLIP $\theta ^{t} $ and CLIP $\theta ^{t-1} $ on our  Retrieval data, while using CMC loss $\mathcal{L}_{CMC}$ to preserve the modality alignment of CLIP $\theta ^{t}$. (c) \textbf{Dynamic Fisher Weight Guard}. Guided by the $\mathcal{L}_{CKT}$ and $\mathcal{L}_{CMC}$ losses, we employ Fisher Importance Mapping (FIM) to dynamically assess weight importance, resulting in an adaptively enhanced L2 regularization loss, $\mathcal{L}_{DFG}$. }
\label{fig2}
\vspace{-1em}
\end{figure*}

\noindent\textbf{Medical Continual Learning.} Continual learning (CL) is an extensively researched topic in machine learning and computer vision, encompassing various scenarios like class-incremental, task-incremental, and domain-incremental learning. Strategies to address this challenge are generally categorized as regularization-based~\cite{CL_non-stationary_data,MAS,lwf,wise_ft}, distillation-based~\cite{Dark_distill,Self-supervised_distill,CL_non-stationary_data,icarl,zscl,snd,GIFT}, architecture-based~\cite{Dytox_arch,dualnet_arch,DIKI,MOE_prompts}, or rehearsal-based~\cite{Dark_distill,BiC,Co2l}. In recent years, several studies have explored the application of CL in medical image analysis, including works on MRI segmentation and classification~\cite{3distill_MIA_MRI,NC_MRI,CLMS_MIA_MRI}, X-ray analysis~\cite{Biaspruner_MICCAI_CT,CL_MICCAIw_X_ray}, pathology image analysis~\cite{CoMEL,conslide,QPMIL}, skin image classification~\cite{Continual-GEN_MICCAI_skin} and robotic surgery~\cite{ICRA_surgery,ICRA_surgery_2}. In response to the escalating size of models and their corresponding data needs, regularization-based~\cite{biomedica} and LoRA-based~\cite{MOE_CL,MoE-Adapters++,med_lora_llama} CL strategies are playing a crucial role in the effective pre-training of medical foundation models. To date, the literature notably lacks a systematic methodology and a corresponding benchmark to investigate the CL performance of medical Vision Language Models (VLMs) and concurrently optimize their task-incremental and domain-incremental capabilities.

\vspace{0.3pt}\noindent\textbf{Continual Learning for VLMs.} VLMs exhibit strong generalization and zero-shot capabilities~\cite{E2ECL,representation_with_noisy_text_supervision,lwf}. While extensive research has focused on finetuning general-domain VLMs~\cite{bert,vis_trans,lora} to boost downstream performance while retaining generalization, finetuning for complex medical scenarios remains underexplored. Parameter-efficient methods like prompt tuning~\cite{CoOp,DIKI,Visual_prompt_tuning} and adapters~\cite{MOE_CL,MoE-Adapters++,MOE_prompts} are often suboptimal for complex tasks due to limited learnable parameters, prompting exploration into robust backbone finetuning to balance stability and plasticity. In CL scenarios with domain shifts, ZSCL~\cite{zscl} preserves zero-shot capabilities via distillation from external datasets, which SND~\cite{snd} enhances using multi-teacher distillation and GIFT~\cite{GIFT} optimizes by replacing the dataset with a generative model to reduce storage. Our approach, however, employs a multi-level, multi-modal Retrieval-Augmented Generation (RAG) framework to construct a dynamic medical external dataset, thereby enhancing the granularity and dynamism of replayed content.

\vspace{0.3pt}\noindent\textbf{Generalist Medical Foundation Models.} As medical expert language models~\cite{BioBERT,PMC-LLaMA,PubmedBert} have advanced, large-scale image–text pretrained VLMs have emerged in the medical field. Contrastive language–vision pretraining aligns images and text within a unified embedding space, delivering strong performance in visual tasks like classification and prognosis. Among generalist models, PMC-CLIP~\cite{Pmc-clip} leverages one million PMC-OA image–caption pairs, trained with dual contrastive learning and masked language modeling. BiomedCLIP~\cite{biomedclip} expands this to PMC-15M pairs and adopts a ViT~\cite{VIT} visual encoder. UniMed-CLIP~\cite{Unimed-clip} extends to more diverse tasks, including generative modeling. Ye et al.~\cite{MedCoss} introduced continual learning across medical domains to improve efficiency and preserve data privacy. Lozano et al.~\cite{biomedica} used DINOv2~\cite{dinov2} to recluster PUBMED data, forming the BIOMEDICA-22M dataset and applying weighted continual pretraining. Recently, MMKD-CLIP~\cite{MMKD-CLIP} combined strengths of prior medical contrastive models via multi-teacher knowledge distillation, achieving superior overall results.

\section{Methodology}

\subsection{Preliminaries}\label{sec:RAG}

\noindent\textbf{Continual Learning.} Given a sequence $[\mathcal{T}^1, \mathcal{T}^2, \cdots, \mathcal{T}^n]$ of $n$ tasks, continual training is performed sequentially on each task $\mathcal{T}^i=(\mathcal{D}^i, C^i)$, where $i=1,\dots,n$. Here, $\mathcal{D}^i$ denotes the dataset of task $i$, represented as $\{\bm{x}^i_j, \bm{y}^i_j\}_{j=1}^{N_i}$, where $\bm{x}^i_j$ is an image, $\bm{y}^i_j$ is a one-hot vector representing its ground truth label, and $N_i$ is the total number of images in the dataset. The class names $C^i=\{c^i_j\}_{j=1}^{m_i}$ map each label to its corresponding object name, where $m_i$ denotes the number of classes in task $\mathcal{T}^i$. The goal of continual  training is to maintain strong performance across all tasks.

We focus on task-incremental and domain-incremental learning in medical continual learning (CL). During inference, the image $\bm{x}$ is given with its task identity $t$, so the model only distinguishes classes within $C^t$. The domain shift is induced by the task order defined in the benchmark.

\vspace{0.3pt}\noindent\textbf{Medical CLIP.}
The CLIP~\cite{CLIP} model includes an image encoder $f_{image}$ and a text encoder $f_{text}$. For image classification, each class $c$ in task $\mathcal{T}^i$ is converted into a sentence like “a photo of {$c$}”. Then $f_{text}$ encodes them into text embeddings ${\bm{t}^i_j}_{j=1}^{m_i}$. The image encoder encodes an input image $\bm{x}_k$. The cosine similarity between the image and text embeddings is $\bm{s}^i_{k,j} = \langle \bm{t}^i_j, f_{image}(\bm{x}_k)\rangle$. The class with the highest score is taken as the prediction. The CLIP architecture typically uses classification losses such as Cross-Entropy ($\mathcal{L} _{CE}$) or Binary Cross-Entropy ($\mathcal{L} _{BCE}$) loss to perform fine-tuning on downstream tasks.

Although Medical CLIP slightly differs from the original CLIP in its training corpus~\cite{PubmedBert,BioBERT} and architecture~\cite{coca,BMC-CLIP-long,CoCa-CXR}, it retains the same zero-shot and fine-tuning protocol. Following ZSCL’s model selection strategy~\cite{zscl}, we adopt ViT-B/16~\cite{VIT} for experiments. To maintain architectural consistency, strong performance, and prevent data leakage, we use BiomedCLIP~\cite{biomedclip} as the backbone. \textbf{Appendix F.1} and \textbf{F.2} contains the rationale for our baseline model selection and demonstrates that our method achieves superior performance on other backbones as well.

\vspace{0.3pt}\noindent\textbf{Offline Multimodal Retrieval Database.} Rapid advancements in Generalist Medical Foundation Models~\cite{Pmc-clip,biomedclip,biomedica,PMC2.2M,Open-PMC-18M} have refined PubMed multimodal data collection. Following BIOMEDICA~\cite{biomedica}, we gathered and processed all literature data, performing segmentation and pseudo-labeling of images, captions, and text. However, much data is unrelated to medical imaging or uses multi-image formats, which are suboptimal for CL. To enhance retrieval efficiency, we compressed textual information and applied a pipeline to decompose multi-image entries, constructing an 18M-scale retrieval database with Qwen3-Embedding-8B~\cite{qwen3}. Database construction details are in the \textbf{Appendix D.1}. We also curated a 3,000-question pool from various sources (literature, datasets, QA reformulation, Wikipedia), refined by domains and anatomical sites.

\subsection{Dynamic Multi-stage Retrieval Mechanism}

\noindent\textbf{Dense Retrieval via Embedding.} Based on the definitions from Sec.~\ref{sec:RAG}, we use the retrieval database $S$ and the medical question pool $M^0$. $S$ contains tuples $(\hat{\mathbf{s}}, \mathbf{c}, \mathbf{i})$, where $\hat{\mathbf{s}}$ is an embedding vector, $\mathbf{c}$ the caption, and $\mathbf{i}$ the image. For CL over $n$ tasks $[\mathcal{T}^1, \dots, \mathcal{T}^n]$, we dynamically add previous task labels. Before task $\mathcal{T}^i$'s training, the pool $M^i$ is:
{
\small
\begin{equation}
M^i = M^0 \cup \left( \bigcup_{k=1}^{i-1} D^k \right).\tag{1}\label{eq1}
\end{equation}
}

\noindent For each query $m \in M^i$, let $S_e = \{ \hat{\mathbf{s}} \mid (\hat{\mathbf{s}}, \mathbf{c}, \mathbf{i}) \in S \}$ be the set of all normalized vectors from the database. We compute its cosine similarity with all $\hat{\mathbf{s}} \in S_e$. The query $m$ is first embedded by model $\Phi_E$~\cite{qwen3} and then normalized:
{
\small
\begin{align}
\langle m, \hat{\mathbf{s}} \rangle &= \frac{\Phi_E(m)}{\|\Phi_E(m)\|_2} \cdot \hat{\mathbf{s}}, \quad \forall \hat{\mathbf{s}} \in S_e, \tag{2} \\
\mathcal{U}_m^i &= \{ \langle m, \hat{\mathbf{s}}\rangle \mid \hat{\mathbf{s}} \in S_e \}. \tag{3}
\end{align}
}

\noindent To ensure $C_m^i$ is robust to numerous tied scores, we avoid simple top-$k$ selection, instead defining a dynamic threshold $\tau_m$ using the $k'$-th largest unique similarity score:
{
\small
\begin{align}
k' = \min(k, |\mathcal{U}_m^i|), \:\: \tau_m = (\text{sort}{\downarrow}(\mathcal{U}_m^i))[k'] .\tag{4,5}
\end{align}
}

\noindent Here, $k$ is the target retrieval size. The final candidate set $C_m^i$ includes all entries scoring $\ge \tau_m$, for reranking:
{
\small
\begin{equation}
C_m^i = \{ (\hat{\mathbf{s}}, \mathbf{c}, \mathbf{i}) \in S \mid \langle m, \hat{\mathbf{s}}\rangle \ge \tau_m \}. \tag{6}
\end{equation}
}

\vspace{0.3pt}\noindent\textbf{Rerank and Gate.} To refine the initial candidate set $C_m^i$, we first rerank all candidates using a VLM~\cite{biomedclip} encoder ($\Phi_{V}$) to compute a cross-modal score $S_{v}$. We then mitigate redundancy by grouping by caption and retaining only the top $k_{v}$ images per group, creating an intermediate set $C_m^{i*}$. This set is processed by a lexical gate, which uses BM25~\cite{BM25} to filter the set down to $k$ candidates if its size $C_m^{i*}$ exceeds the target $k$. Let $R_{topk}(C, S, k)$ be the top-$k$ selection operator. The final set $C$ is:
{
\small
\begin{equation}
C_r^{i} = \begin{cases} C_m^{i*} & \text{if } |C_m^{i*}| \le k \\ R_{topk}(C_m^{i*}, S_{bm25}, k) & \text{if } |C_m^{i*}| > k \end{cases}.\tag{7}
\end{equation}
}

\noindent Finally, the $k$ candidates in $C_r^{i}$ are sorted by $S_{v}$.

\vspace{0.3pt}\noindent\textbf{Dynamic Siphon.} To enhance knowledge retention, we divide the $M^i$ into three mutually exclusive subsets: $M_{\text{task}}$, $M_{\text{domain}}$, and $M_{\text{gen}}$, which correspond to task-specific, domain-related, and general-purpose queries, respectively. The formal definition is given as follows:
{
\small
\begin{gather}
M_{task} = \bigcup_{k=1}^{i-1} D^k,\:\: C_{task} = \bigcup_{k=1}^{i-1} \{ \Phi_{D}(x) \mid x \in D^k \} , \tag{8,9} \\
M_{domain} = \{ m \in M^0 \mid \Phi_{D}(m) \in C_{task} \}, \tag{10} \\
M_{gen} = \mathcal{S}(M^0 \setminus M_{domain}, n). \tag{11}
\end{gather}
}

Let $\Phi_D$~\cite{biomedclip} map $x$ to its domain category, $C_{\text{task}}$ be the set of $i-1$ previous task domains, $\mathcal{S}$ be a random sampler, and $n$ be the number of samples. The final tuple set $C^i$ is generated via top-$k$ sampling with distinct parameters $a, b, c$ at different levels.
{
\small
\begin{equation}
C^i = \begin{cases}
R_{topk}(C_r^i, S_v, a) & \forall  m \in M_{task} \\
R_{topk}(C_r^i, S_v, b) & \forall  m \in M_{domain} \\
R_{topk}(C_r^i, S_v, c) & \forall  m \in M_{gen}
\end{cases}
\tag{12}\label{eq:siphon}
\end{equation}
}

\subsection{Distillation, Alignment and Guard}

\vspace{0.3pt}\noindent\textbf{Contrastive Knowledge Transfer.} For a batch of $B$ image-text pairs from $C^i$, we use the BiomedCLIP~\cite{biomedclip} image encoder $f_{image}^i$ and text encoder $f_{text}^i$ to compute the $B \times B$ cross-modal similarity matrix $M^i = (\mathbf{m}^i_{k,j})$ as:

{
\small
\begin{equation}
\mathbf{m}^i_{k,j} = \langle f_{text}^i(\mathbf{c}^i_{k,j}),f_{image}^i(\mathbf{i}^i_{k,j}) \rangle.
\tag{13}
\end{equation}
}

\noindent Then, the teacher model from the task $\mathcal{T}^{i-1}$, with encoders $f_{image}^{i-1}$ and $f_{text}^{i-1}$, processes the same batch to get its similarity matrix $M^{i-1}$. Logits are similarities scaled by $\tau $:
{
\small
\begin{equation}
Z^i=\frac{M^i}{\tau} ,\:\:Z^{i-1}=\frac{M^{i-1}}{\tau}.
\tag{14}
\end{equation}
}

To compute image-to-text and text-to-image alignments, we convert similarity scores to probability distributions using $\sigma$ and measure their $\mathcal{KL}$ divergence.
{
\small
\begin{align}
\mathcal{L}_{Distill-i2t} &=  \sum_{i=1}^{B} \mathcal{KL} \left( \sigma(Z^{i-1}_{k,:}) \,\Vert\,\sigma(Z^{i}_{k,:}) \right), \tag{15} \\
\mathcal{L}_{Distill-t2i} &=  \sum_{j=1}^{B} \mathcal{KL} \left( \sigma(Z^{i-1}_{:,j}) \,\Vert\, \sigma(Z^{i}_{:,j}) \right). \tag{16}
\end{align}
}
The final Contrastive Knowledge Transfer loss $\mathcal{L}_{CKT}$ is:
{
\small
\begin{align}
\mathcal{L}_{CKT} = \mathcal{L}_{Distill-i2t} + \mathcal{L}_{Distill-t2i}. \tag{17}
\end{align}
}

\vspace{0.3pt}\noindent\textbf{Cross-Modality Consistency.} In CL knowledge distillation, the teacher model (one task behind the student) also suffers from catastrophic forgetting, notably decoupling cross-modal knowledge. We thus perform contrastive learning with $M^i$ against the identity matrix for $C^i$ to get Cross-Modality Consistency loss $\mathcal{L}_{CMC}$:
{
\small
\begin{align}
\mathcal{L}_{CMC} = \mathcal{L}_{Align-i2t} + \mathcal{L}_{Align-t2i}. \tag{18}
\end{align}
}

\noindent Combining the Cross-Entropy ($\mathcal{L}_{CE}$) classification loss for task $\mathcal{T}^i$, the total training loss for the model is defined as:
{
\small
\begin{equation} 
    \mathcal{L}_{Train} = \mathcal{L}_{CE} + \alpha \mathcal{L}_{CKT} + \beta \mathcal{L}_{CMC} \,,\tag{19}
\end{equation}
}

\noindent where $\alpha$ and $\beta$ are trade-off hyperparameters.

\vspace{0.3pt}\noindent\textbf{Dynamic Fisher Weight Guard.} Applying a MSE~\cite{MSE} penalty between the weights of the fine-tuned VLM and its original pre-trained state~\cite{si} can effectively mitigate overfitting induced by the cross-entropy gradient and address the catastrophic forgetting problem in the foundational model. Elastic Weight Consolidation (EWC)~\cite{ocf} is a typical weight consolidation method that imposes a parameter importance-weighted $l_2$ loss~\cite{l2_norm}, as follows:
{
\small
\begin{equation}
    \mathcal{L}_{EWC} = \sum_{i} \mathcal{W}_{\theta_{i}^{t-1}} \cdot \left( \theta^{t}_i - \theta^{t-1}_i \right)^2 \,.\tag{20}
\end{equation}
}

\noindent Parameter importance $\,\smash{\mathcal{W}_{\theta_{i}^{t-1}}}$ is the Fisher Information Matrix diagonal. We further propose: if fine-tuning interference causes forgetting, shouldn't $\mathcal{L}_{CE}$ also dynamically adapt to distillation and alignment shifts? The Fisher Information Matrix evolving via gradient backpropagation is a natural mechanism for this:
{
\small
\begin{align}
W_i^{(j)} &= \left( \frac{\partial \left( \mathcal{L}_{\text{Train}}^{(j)}-\mathcal{L}_{\text{CE}}^{(j)} \right)}{\partial \theta_{i}^t} \right) ^2,\tag{21}\\
\mathcal{L}_{\text{DFG}}^{(j)} &= \sum_{i} W_i^{(j)} \cdot \left( \theta^{t(j)}_i - \theta^{t-1}_i \right)^2 .\tag{22}
\end{align}
}

\noindent where $\smash{\mathcal{W}_{\theta_{i}^t}^{(j)}}$ denotes the diagonal Fisher information of model parameter $\theta_{i}^t$ at the $j^{th}$ optimization step.

{\renewcommand{\arraystretch}{0.9}
\begin{table*}[t]
\small
\centering
\caption{Comparison of SOTA methods on HieraMedTransfer Order I and II. \colorbox{gray!8}{{\textbf{Gray Background}}} indicates baseline. \colorbox{RedOrange!8}{\textbf{Red Background}} \& \textbf{bold} indicate best results. \textRed{\textit{Red}}/\textBlue{\textit{blue}} fonts indicate \textit{increase/decrease} relative to baseline.}
\vspace{-1.5ex}
\begin{tabular}{@{}>{\centering\arraybackslash}m{1.8cm}|>{\centering\arraybackslash}m{1.7cm}|cc|cc|cc!{\vrule width 1.2pt}cc|cc|cc@{}}
\toprule
\multirow{2}{*}{\textbf{Method}} & \multirow{2}{*}{\textbf{Publication}} 
& \multicolumn{6}{c!{\vrule width 1.2pt}}{\textbf{HieraMedTransfer Order I}} 
& \multicolumn{6}{c}{\textbf{HieraMedTransfer Order II}} \\
\cmidrule(lr){3-8} \cmidrule(lr){9-14}
&& Transfer & $\Delta$ & Avg. & $\Delta$ & Last & $\Delta$
& Transfer & $\Delta$ & Avg. & $\Delta$ & Last & $\Delta$ \\
\midrule
Zero-shot  &-& 57.5 &  - & 52.9 & - & 53.0 & - & 57.5 &  - & 52.9 & - & 53.0 & - \\
Continual FT &-& 51.4 & - & 67.6 & - & 70.8 & - & 47.7 & - & 61.9 & - & 58.6 & - \\
\midrule
\rowcolor{gray!8}
$l_2$ baseline &-& 53.6 & {\textit{\footnotesize 0.0}} & 68.5 & {\textit{\footnotesize 0.0}} & 74.1 & {\textit{\footnotesize 0.0}}
& 45.8 & {\textit{\footnotesize 0.0}} & 63.9 & {\textit{\footnotesize 0.0}} & 66.8 & {\textit{\footnotesize 0.0}} \\
\midrule
LwF~\cite{lwf} &TPAMI 2017& 43.7 & \textBlue{\textit{\footnotesize -9.9}} & 54.3 & \textBlue{\textit{\footnotesize -14.2}} & 65.1 & \textBlue{\textit{\footnotesize -9.0}}
& 47.0 & \textRed{\textit{\footnotesize +1.2}} & 61.7 & \textBlue{\textit{\footnotesize -2.2}} & 61.7 & \textBlue{\textit{\footnotesize -5.1}} \\
iCaRL~\cite{icarl} &CVPR 2017& 52.8 & \textBlue{\textit{\footnotesize -0.8}} & 70.7 & \textRed{\textit{\footnotesize +2.2}} & 77.8 & \textRed{\textit{\footnotesize +3.7}}
& 48.1 & \textRed{\textit{\footnotesize +2.3}} & 66.0 & \textRed{\textit{\footnotesize +2.1}} & 73.9 & \textRed{\textit{\footnotesize +7.1}} \\
WiSE-FT~\cite{wise_ft} &CVPR 2022& 53.2 & \textBlue{\textit{\footnotesize -0.4}} & 69.4 & \textRed{\textit{\footnotesize +0.9}} & 75.2 & \textRed{\textit{\footnotesize +1.1}}
& 48.3 & \textRed{\textit{\footnotesize +2.5}} & 64.8 & \textRed{\textit{\footnotesize +0.9}} & 65.9 & \textBlue{\textit{\footnotesize -0.9}} \\
ZSCL~\cite{zscl} &ICCV 2023& 57.7 & \textRed{\textit{\footnotesize +4.1}} & 70.5 & \textRed{\textit{\footnotesize +2.0}} & 77.6 & \textRed{\textit{\footnotesize +3.5}}
& 45.2 & \textBlue{\textit{\footnotesize -0.6}} & 65.0 & \textRed{\textit{\footnotesize +1.1}} & 76.9 & \textRed{\textit{\footnotesize +10.1}} \\
MoE-CL~\cite{MOE_CL} &CVPR 2024& 56.8 & \textRed{\textit{\footnotesize +3.2}} & 70.7 & \textRed{\textit{\footnotesize +2.2}} & 76.1 & \textRed{\textit{\footnotesize +2.0}}
& 47.6 & \textRed{\textit{\footnotesize +1.8}} & 66.1 & \textRed{\textit{\footnotesize +2.2}} & 74.2 & \textRed{\textit{\footnotesize +7.4}} \\
SND~\cite{snd} &ECCV 2024& 52.0 & \textBlue{\textit{\footnotesize -1.6}} & 67.5 & \textBlue{\textit{\footnotesize -1.0}} & 70.1 & \textBlue{\textit{\footnotesize -4.0}}
& 46.9 & \textRed{\textit{\footnotesize +1.1}} & 61.0 & \textBlue{\textit{\footnotesize -2.9}} & 53.3 & \textBlue{\textit{\footnotesize -13.5}} \\
DIKI~\cite{DIKI} &ECCV 2024& 56.3 & \textRed{\textit{\footnotesize +2.7}} & 70.9 & \textRed{\textit{\footnotesize +2.4}} & 77.4 & \textRed{\textit{\footnotesize +3.3}}
& 46.6 & \textRed{\textit{\footnotesize +0.8}} & 65.7 & \textRed{\textit{\footnotesize +1.8}} & 76.7 & \textRed{\textit{\footnotesize +9.9}} \\
GIFT~\cite{GIFT} &CVPR 2025& 53.4 & \textBlue{\textit{\footnotesize -0.2}} & 69.8 & \textRed{\textit{\footnotesize +1.3}} & 75.2 & \textRed{\textit{\footnotesize +1.1}}
& 46.8 & \textRed{\textit{\footnotesize +1.0}} & 65.0 & \textRed{\textit{\footnotesize +1.1}} & 71.3 & \textRed{\textit{\footnotesize +4.5}} \\
\midrule

$\text{PRIMED}_{uni} $ &-& 57.1 & \textRed{\textit{\footnotesize +3.5}} & 72.7 & \textRed{\textit{\footnotesize +4.2}} & 81.7 & \textRed{\textit{\footnotesize +7.6}}
& 48.0 & \textRed{\textit{\footnotesize +2.2}} & 67.7 & \textRed{\textit{\footnotesize +3.8}} & 77.0 & \textRed{\textit{\footnotesize +10.2}} \\
\rowcolor{RedOrange!8}
$\text{PRIMED}_{dyn}$ &-& \textbf{58.3} & \textbf{\textRed{\textit{\footnotesize +4.7}}} & \textbf{73.1} & \textbf{\textRed{\textit{\footnotesize +4.6}}} & \textbf{82.1} & \textbf{\textRed{\textit{\footnotesize +8.0}}}
& \textbf{48.5} & \textbf{\textRed{\textit{\footnotesize +2.7}}} & \textbf{68.0} & \textbf{\textRed{\textit{\footnotesize +4.1}}} & \textbf{81.2} & \textbf{\textRed{\textit{\footnotesize +14.4}}} \\

\bottomrule

\end{tabular}%
\vspace{-1ex}
\label{tab1}
\end{table*}
}

\subsection{MGTIL Benchmark}\label{sec:mgtil_benchmark}

\begin{wrapfigure}{r}{0.5\textwidth} 
  \vspace{-10pt} 
  \begin{center}
    \includegraphics[width=1\linewidth]{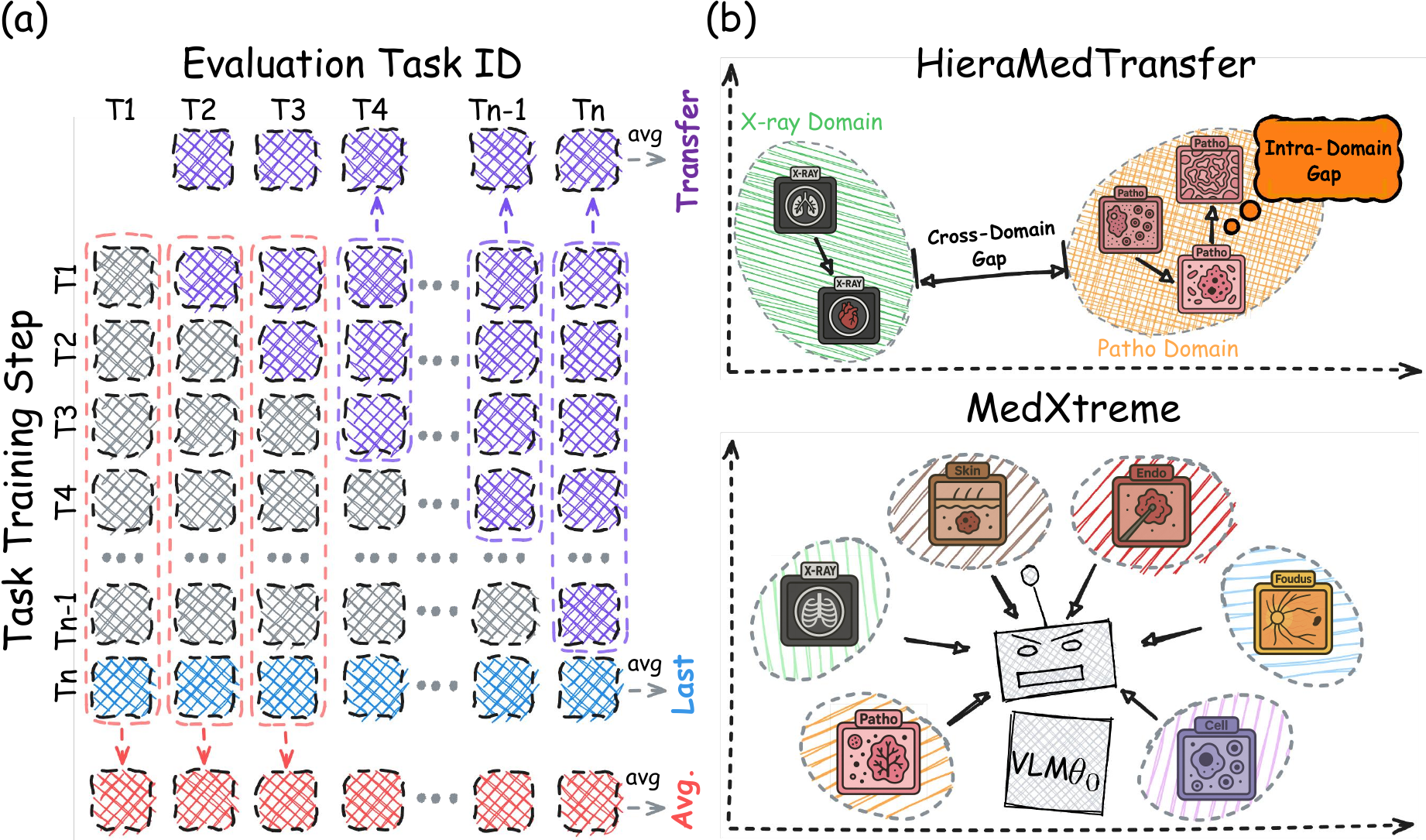}
  \end{center}
  \caption{(a) Illustration of calculating CL metrics: Transfer, Avg., and Last. (b) Overview of MGTIL's two tasks, evaluating CL scenarios across intra-domain, cross-domain, and task retention.}
  \label{fig3}
  \vspace{-10pt} 
\end{wrapfigure}

As shown in Fig.~\ref{fig3}(b), MGTIL features two distinct tasks to evaluate the CL performance of VLMs at different levels.

\vspace{0.3pt}\noindent\textbf{HieraMedTransfer Benchmark.} Intra-domain medical images are highly similar~\cite{imagenet_classifier,face,patho_finegrain,culprit}, requiring continual discrimination to memorize fine-grained features. Concurrently, retaining memory across large cross-domain gaps is a severe challenge~\cite{long_task}. We propose HieraMedTransfer to address both challenges. We curated nine datasets~\cite{breakhis,chaoyang,Nucls,PD,CheXchoNet,ranzcr,eyepacs,AIROGS,FARFUM-RoP} from three domains (X-ray, pathology, and fundus) covering diverse resolutions, regions, and involvement. We designed two sequences: OrderI trains sequentially by domain to simulate pre-training, while OrderII uses a randomized alphabetical task arrangement to simulate a realistic clinical CL scenario.

\vspace{0.3pt}\noindent\textbf{MedXtreme Benchmark.} MedXtreme targets the challenging tasks~\cite{NCT,NIH-Chest-ray,AOD,Pitvis,isic-2024,BMC} across 6 medical domains: X-ray, fundus, pathology, endoscopy, dermoscopy and cell, each involving up to 33 classes. These domains and tasks exhibit insufficient training data and limited exploration. Experiments were conducted with alphabetical (OrderI) and random (OrderII) task sequences. Detailed benchmark datasets are presented in the \textbf{Appendix D.2}.

\vspace{0.3pt}\noindent\textbf{Evaluation Metrics.} We perform the multi-task evaluation following the ZSCL, as depicted in Fig.~\ref{fig3}(a). Zero-shot transfer ability enables predictions on all datasets. The Avg metric is the average accuracy across all datasets and timestamps. The Last metric is the average performance of all tasks after CL. The Transfer metric is the average task performance in the upper-right triangle of the matrix, measuring the preservation of zero-shot transfer ability. Tasks are first averaged to ensure equal dataset weighting. Before learning task $i$, tasks $j \ge i$ are not fine-tuned; their performance thus indicates zero-shot ability. However, even Medical VLMs exhibit insufficient zero-shot performance on the MedXtreme. Therefore, we adopt traditional Task-CL metrics, such as ACC, AUC, BWT~\cite{BWT} and Forgetting~\cite{forget}.

\section{Experiments}

{\renewcommand{\arraystretch}{0.9}
\begin{table*}[t]
\small
\centering
\caption{Comparison of SOTA methods on MedXtreme Order I and II. \colorbox{gray!8}{{\textbf{Gray Background}}} indicates baseline. \colorbox{RedOrange!8}{\textbf{Red Background}} \& \textbf{bold} indicate best results. \textRed{\textit{Red}}/\textBlue{\textit{blue}} fonts indicate \textit{increase/decrease} relative to baseline.}
\vspace{-1.5ex}
\begin{tabular}{@{}>{\centering\arraybackslash}m{1.8cm}|>{\centering\arraybackslash}m{1.7cm}|cc|cc|cc!{\vrule width 1.2pt}cc|cc|cc@{}}
\toprule
\multirow{2}{*}{\textbf{Method}} & \multirow{2}{*}{\textbf{Publication}} 
& \multicolumn{6}{c!{\vrule width 1.2pt}}{\textbf{MedXtreme Order I}} 
& \multicolumn{6}{c}{\textbf{MedXtreme Order II}} \\
\cmidrule(lr){3-8} \cmidrule(lr){9-14}
&& ACC & $\Delta$ & AUC & $\Delta$ & BWT & $\Delta$
& ACC & $\Delta$ & AUC & $\Delta$ & BWT & $\Delta$ \\
\midrule
Zero-shot &-& 9.9 &  - & 61.2 & - & - & - & 9.9 &  - & 61.2 & - & - & - \\
Continual FT &-& 61.0 & - & 86.7 & - & -15.1 & - & 60.0 & - & 84.1 & - & -16.0 & - \\
\midrule
\rowcolor{gray!8}
$l_2$ baseline &-& 61.1 & \textit{\footnotesize 0.0} & 82.9 & \textit{\footnotesize 0.0} & -10.0 & \textit{\footnotesize 0.0} 
& 57.3 & \textit{\footnotesize 0.0} & 81.2 & \textit{\footnotesize 0.0} & -14.5 & \textit{\footnotesize 0.0} \\
\midrule
LwF~\cite{lwf} &TPAMI 2017& 51.5 & \textBlue{\textit{\footnotesize -9.6}} & 81.4 & \textBlue{\textit{\footnotesize -1.5}} & -8.3 & \textRed{\textit{\footnotesize +1.7}} 
& 44.2 & \textBlue{\textit{\footnotesize -13.1}} & 81.0 & \textBlue{\textit{\footnotesize -0.2}} & -11.3 & \textRed{\textit{\footnotesize +3.2}} \\
iCaRL~\cite{icarl} &CVPR 2017& 65.9 & \textRed{\textit{\footnotesize +4.8}} & 84.8 & \textRed{\textit{\footnotesize +1.9}} & -3.1 & \textRed{\textit{\footnotesize +6.9}}
& 64.2 & \textRed{\textit{\footnotesize +6.9}} & 84.3 & \textRed{\textit{\footnotesize +3.1}} & -5.1 & \textRed{\textit{\footnotesize +9.4}} \\
WiSE-FT~\cite{wise_ft} &CVPR 2022& 65.1 & \textRed{\textit{\footnotesize +4.0}} & 86.5 & \textRed{\textit{\footnotesize +3.6}} & -8.5 & \textRed{\textit{\footnotesize +1.5}}
& 64.1 & \textRed{\textit{\footnotesize +6.8}} & 85.6 & \textRed{\textit{\footnotesize +4.4}} & -9.2 & \textRed{\textit{\footnotesize +5.3}} \\
ZSCL~\cite{zscl} &ICCV 2023& 53.7 & \textBlue{\textit{\footnotesize -7.4}} & 79.9 & \textBlue{\textit{\footnotesize -3.0}} & -6.5 & \textRed{\textit{\footnotesize +3.5}}
& 48.3 & \textBlue{\textit{\footnotesize -9.0}} & 78.6 & \textBlue{\textit{\footnotesize -2.6}} & -13.7 & \textRed{\textit{\footnotesize +0.8}} \\
MoE-CL~\cite{MOE_CL} &CVPR 2024& 65.3 & \textRed{\textit{\footnotesize +4.2}} & 84.9 & \textRed{\textit{\footnotesize +2.0}} & -4.1 & \textRed{\textit{\footnotesize +5.9}}
& 64.7 & \textRed{\textit{\footnotesize +7.4}} & 84.5 & \textRed{\textit{\footnotesize +3.3}} & -4.4 & \textRed{\textit{\footnotesize +10.1}} \\
SND~\cite{snd} &ECCV 2024& 61.7 & \textRed{\textit{\footnotesize +0.6}} & 85.3 & \textRed{\textit{\footnotesize +2.4}} & -13.9 & \textBlue{\textit{\footnotesize -3.9}}
& 57.6 & \textRed{\textit{\footnotesize +0.3}} & 84.3 & \textRed{\textit{\footnotesize +3.1}} & -18.8 & \textBlue{\textit{\footnotesize -4.3}} \\
DIKI~\cite{DIKI} &ECCV 2024& 64.8 & \textRed{\textit{\footnotesize +3.7}} & 86.7 & \textRed{\textit{\footnotesize +3.8}} & -9.2 & \textRed{\textit{\footnotesize +0.8}}
& 63.1 & \textRed{\textit{\footnotesize +5.8}} & 85.2 & \textRed{\textit{\footnotesize +4.0}} & -10.3 & \textRed{\textit{\footnotesize +4.2}} \\
GIFT~\cite{GIFT} &CVPR 2025& 66.0 & \textRed{\textit{\footnotesize +4.9}} & 86.6 & \textRed{\textit{\footnotesize +3.7}} & -3.7 & \textRed{\textit{\footnotesize +6.3}}
& 65.7 & \textRed{\textit{\footnotesize +8.4}} & 85.2 & \textRed{\textit{\footnotesize +4.0}} & -4.1 & \textRed{\textit{\footnotesize +10.4}} \\
\midrule
$\text{PRIMED}_{uni}  $ &-& 66.2 & \textRed{\textit{\footnotesize +5.1}} & 86.7 & \textRed{\textit{\footnotesize +3.8}} & -4.4 & \textRed{\textit{\footnotesize +5.6}}
& 64.5 & \textRed{\textit{\footnotesize +7.2}} & 85.4 & \textRed{\textit{\footnotesize +4.2}} & -6.6 & \textRed{\textit{\footnotesize +7.9}} \\
\rowcolor{RedOrange!8}
$\text{PRIMED}_{dyn}$ &-& \textbf{68.6} & \textbf{\textRed{\textit{\footnotesize +7.5}}} & \textbf{87.4} & \textbf{\textRed{\textit{\footnotesize +4.5}}} & \textbf{-2.7} & \textbf{\textRed{\textit{\footnotesize +7.3}}}
& \textbf{68.1} & \textbf{\textRed{\textit{\footnotesize +10.8}}} & \textbf{86.3} & \textbf{\textRed{\textit{\footnotesize +5.1}}} & \textbf{-3.4} & \textbf{\textRed{\textit{\footnotesize +11.1}}} \\
\bottomrule
\end{tabular}%

\vspace{-1ex}
\label{tab2}
\end{table*}}

\subsection{Experimental Setting}

\noindent\textbf{Implementation Details.} We conduct all experiments on a 4-NVIDIA A6000 GPU workstation, using BiomedCLIP~\cite{biomedclip} as the backbone. Each MGTIL task is trained for 1,000 iterations with a batch size of 64 and a $1 \times 10^{-5}$ learning rate. Full retrieval and model training settings and hyperparameters are detailed in Tab.~\ref{tab4} and the \textbf{Appendix E}.

\vspace{0.3pt}\noindent\textbf{Datasets and Task Sequence.} Following Sec.~\ref{sec:mgtil_benchmark}'s design, we detail the datasets and order. HieraMedTransfer is evaluated on RANZCR~\cite{ranzcr}, CheXchoNet~\cite{CheXchoNet}, PD~\cite{PD}, Breakhis~\cite{breakhis}, Chaoyang~\cite{chaoyang}, NuCLS~\cite{Nucls}, Eyepacs~\cite{eyepacs}, AIROGS~\cite{AIROGS}, and FARFUM-RoP~\cite{FARFUM-RoP}. The two task orders are: Order I (as introduced above) and Order II (alphabetical). MedXtreme comprises AOD~\cite{AOD}, NCT100K~\cite{NCT}, PITVIS~\cite{Pitvis}, ISIC2024~\cite{isic-2024}, NIH-Chest-Xray~\cite{NIH-Chest-ray}, and BMC~\cite{BMC}. Similarly, the task orders are Order I (alphabetical) and Order II (random, as introduced above). The \textbf{Appendix D.2} details the dataset composition, distribution, and volume, confirming its isolation to prevent data leakage. 

\vspace{0.3pt}\noindent\textbf{Reference Dataset Construction.} Distillation-based methods~\cite{zscl,snd} require a reference dataset. Since prior work focused on natural images, we adapted these methods to the medical domain for a fair comparison. While existing methods use uniform sampling from coarse-grained labels, we construct an equivalently sized reference dataset to ZSCL's via proportional retrieval using 30 medical domain keywords. Text-to-image methods~\cite{GIFT} generate data from customized labels; we utilize our fine-grained question pool as generation prompts. Notably, while our method expectedly excels in dynamic multi-stage retrieval, it also outperforms all reference-set distillation methods under the uniform retrieval setting showing in Tab.~\ref{tab1} and Tab.~\ref{tab2}.

\subsection{Comparison with State-of-the-art Methods}

The average performance of different methods on Sequence I and Sequence II of the MGTIL benchmark is presented in Tab.~\ref{tab1} and Tab.~\ref{tab2}, respectively (more granular and comprehensive numerical results are available in the \textbf{Appendix D.2}). At the top of the table, the Zero-shot method is used to obtain the logical upper bound for transfer, while continual FT without any strategy provides the logical lower bounds for Avg. and Last. $l_2$ regularization, which effectively constrains the magnitude of fine-tuning, yields a more balanced result and is treated as our baseline.

\begin{wrapfigure}{r}{0.5\textwidth} 
  \vspace{-20pt} 
  \begin{center}
    \includegraphics[width=1\linewidth]{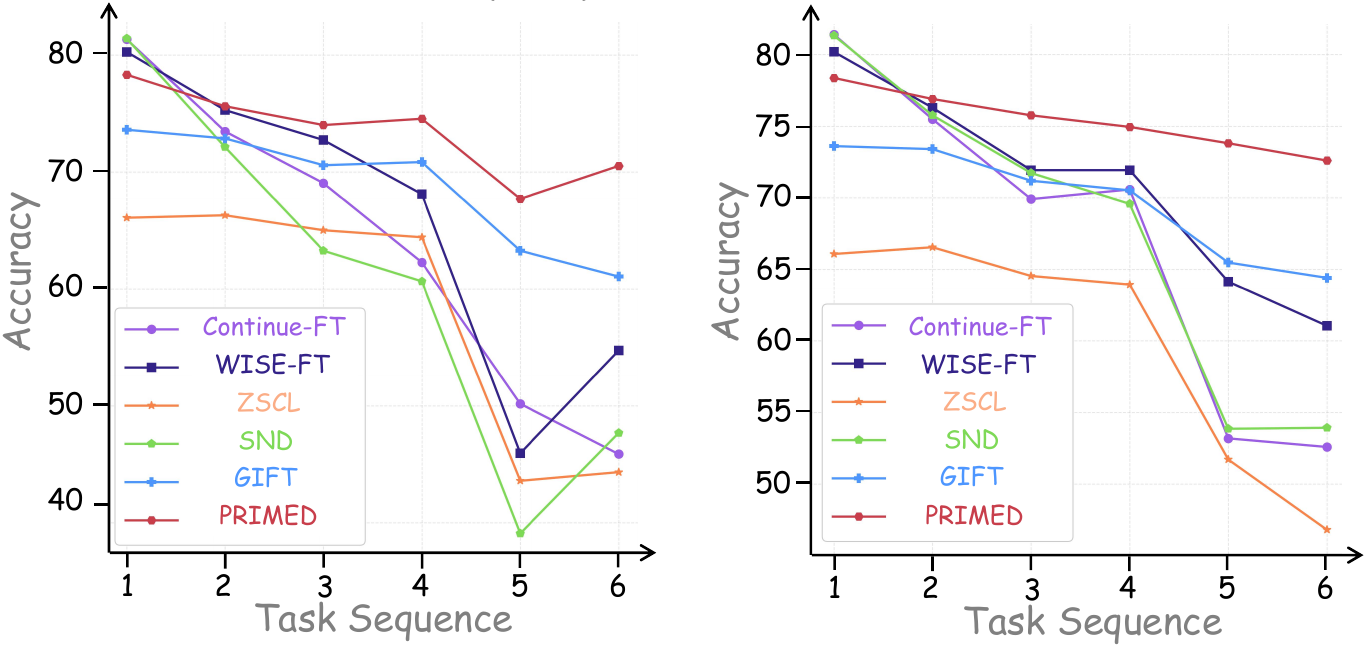}
  \end{center}
  \caption{ACC of 1st task (AOD) in MedXtreme Order I and II}
  \label{fig4}
  \vspace{-10pt} 
\end{wrapfigure}

\vspace{0.3pt}\noindent\textbf{Results on HieraMedTransfer Benchmark.} Order I uses sequential fine-tuning on uniform domains, creating a regular sequence. Thus, ZSCL~\cite{zscl} with reference data and sequential MOE-CL~\cite{MOE_CL} adapt better here. Conversely, Order II randomizes inter-task domain distance. This greater challenge explains its lower average performance versus Order I. Here, rehearsal, reference data, and prompt tuning methods~\cite{DIKI} performed well. $\text{PRIMED}_{uni}$ using ZSCL's reference data beat all comparators, showing our loss function's suitability for complex medical tasks. Moreover, $\text{PRIMED}_{dyn}$ using dynamic multi-stage retrieval achieved superior results. It reached state-of-the-art (SOTA) on all metrics with up to a 5.6\% improvement and avoids data contamination from rehearsal or generative methods.

\vspace{0.3pt}\noindent\textbf{Results on MedXtreme Benchmark.} SOTA performance on both task sequences is achieved by $\text{PRIMED}_{dyn}$. As peak performance typically occurs post-fine-tuning, BWT and Forgetting exhibit an inverse relationship. We therefore report BWT scores exclusively. Tab.~\ref{tab2} demonstrates that dynamic multi-stage retrieval is essential for constructing the granular memory required to master complex tasks. Consequently, replay-based~\cite{icarl} and generative-based~\cite{GIFT} methods, while competitive, are outperformed by our approach. Crucially, beyond its quantitative superiority, $\text{PRIMED}_{dyn}$ obviates the need for original training data, positioning it as a more practical method for real-world medical scenarios. Fig.~\ref{fig4} shows the first task (AOD) ACC variation during training. Our method avoids abrupt catastrophic forgetting and exhibits the lowest overall forgetting.

\subsection{Ablation Study}

Our method is validated on all MGTIL sequences. Tab.\ref{tab3} and Tab.\ref{tab4} provide ablations for HieraMedTransfer Order I. Full results are in the \textbf{Appendix E}.

\noindent\textbf{Modular Level Analysis.} As shown in Tab.\ref{tab3}, we decouple the three modules via distinct losses and conduct a comprehensive ablation. Contrastive Knowledge Transfer \textbf{(CKT)} is the foundational component; its removal causes a drastic performance degradation across all levels. Cross-Modality Consistency \textbf{(CMC)} leverages mini-batch contrastive learning to preserve zero-shot capabilities, preventing excessive knowledge distribution shifts. Finally, the Dynamic Fisher Weight Guard \textbf{(DFG)} provides dynamic knowledge refinement, comprehensively improving overall performance.

\begin{wraptable}{r}{0.5\textwidth} 
    \vspace{-10pt} 
    \centering
    \caption{Ablation study of different modules.}
    \resizebox{1\linewidth}{!}{%
    \begin{tabular}{@{}>{\centering\arraybackslash}m{1.2cm}
                      >{\centering\arraybackslash}m{1.2cm}
                      >{\centering\arraybackslash}m{1.2cm}|c|c|c@{}}
        \toprule
        +CKT & +CMC & +DFG & Transfer & Avg. & Last \\ 
        \midrule
        $\checkmark$ & & & 56.2 & 71.8 & \textbf{82.6 }\\ 
         & $\checkmark$ & & 54.3 & 68.9 & 78.9 \\ 
        $\checkmark$ & $\checkmark$ & & 56.9 & 71.6 & 82.2 \\ 
        $\checkmark$ & & $\checkmark$ & 57.2 & 72.8 & 82.5 \\ 
         & $\checkmark$& $\checkmark$ & 56.7 & 70.2 & 77.1 \\ 
        \cellcolor{RedOrange!8}{$\checkmark$} & \cellcolor{RedOrange!8}{$\checkmark$} & \cellcolor{RedOrange!8}{$\checkmark$} 
        & \cellcolor{RedOrange!8}{\textbf{58.3}} 
        & \cellcolor{RedOrange!8}{\textbf{73.1}} 
        & \cellcolor{RedOrange!8}{82.1} \\ 
        \bottomrule
    \end{tabular}%
    }
    \vspace{-10pt} 
    \label{tab3}
\end{wraptable}

\vspace{0.3pt}\noindent\textbf{Component and Hyperparameter Analysis.} As shown in Tab.~\ref{tab4}, we conduct an ablation study to validate the feasibility and necessity of the components and hyperparameters used in our training process. The best-performing results are demonstrated on the HieraMedTransfer Order I. To demonstrate the robustness of our method, all hyperparameters in the aforementioned experiments are uniformly set, using a random seed of 42. The settings are divided into 4 main categories. For the teacher model, we compare three variants: the initial model, the model from the previous fine-tuning task (Last), and an ensemble model based on the WISE~\cite{wise_ft} weight-averaging strategy. It is demonstrated that for medical VLMs, the 'Last' model achieves the best performance, likely due to discrepancies in knowledge distribution. This finding also holds true for the highly challenging data scenarios represented by MedXtreme. Regarding the training loss hyperparameters, we utilized the settings that achieved peak performance. At the regularization constraint level, we demonstrate the value of our dynamic approach, which outperforms standard $l_2$ and EWC. In terms of dynamic retrieval, our dynamic multi-stage retrieval shows significant performance gains. Furthermore, we present an analysis in the \textbf{Appendix E.4} on the impact of the optimal class-wise retrieval ratio and the total volume of retrieved data within our Dynamic Siphon module.

\begin{table*}[t]
    \centering
    \caption{\textbf{Ablation experiments.} Our method uses Last CLIP and Hierarchical Retrieval for reference data, applies simultaneous image-text distillation, and leverages the DFG algorithm for dynamic weighting. Default settings are marked in \colorbox{RedOrange!8}{\textbf{Red Background}}.}
    \vspace{-0.5em}

    \resizebox{1\linewidth}{!}{%
    \begin{minipage}{0.32\linewidth}
        \centering
        \caption*{(a) \textbf{Teacher Model.}}
        \vspace{-2ex}
        \begin{tabular}{@{}c|c|c|c@{}}
            \toprule
            Teacher & Transfer & Avg. & Last \\
            \midrule
            Initial CLIP & 57.9 & 70.7 & 78.2 \\
            \rowcolor{RedOrange!8}
            Last CLIP & \textbf{58.3} & \textbf{73.1} & \textbf{82.1} \\
            WISE(0.5) & 58.1 & 71.4 & 79.1 \\
            \bottomrule
        \end{tabular}
    \end{minipage}%
    \hspace{0.05\linewidth}%
    \begin{minipage}{0.32\linewidth}
        \centering
        \caption*{(b) \textbf{Distillation Loss.}}
        \vspace{-2ex}
        \begin{tabular}{@{}c|c|c|c@{}}
            \toprule
            Loss & Transfer & Avg. & Last \\
            \midrule
            Image-only & 57.7 & 72.4 & 81.3 \\
            Text-only & 58.2 & 73.0 & 80.5 \\
            \rowcolor{RedOrange!8}
            Contrastive & \textbf{58.3} & \textbf{73.1} & \textbf{82.1} \\
            \bottomrule
        \end{tabular}
    \end{minipage}%
    \hspace{0.05\linewidth}%
    \begin{minipage}{0.32\linewidth}
        \centering
        \caption*{(c) \textbf{Scale of Distillation.}}
        \vspace{-2ex}
        \begin{tabular}{@{}c|c|c|c@{}}
            \toprule
            CKT Scale & Transfer & Avg. & Last \\
            \midrule
            $\alpha=0.5$ & 58.3 & 72.8 & 81.3 \\
            \rowcolor{RedOrange!8}
            $\alpha=1$ & \textbf{58.3} & \textbf{73.1} & \textbf{82.1} \\
            $\alpha=1.5$ & 58.3 & 73.0 & 81.9 \\
            \bottomrule
        \end{tabular}
    \end{minipage}%
    }

    \vspace{0.5ex}

    \resizebox{1\linewidth}{!}{%
    \begin{minipage}{0.32\linewidth}
        \centering
        \caption*{(d) \textbf{Scale of Image-Text Alignment.}}
        \vspace{-2ex}
        \begin{tabular}{@{}c|c|c|c@{}}
            \toprule
            CMC Scale & Transfer & Avg. & Last \\
            \midrule
            $\beta=0.0$ & 57.2 & 72.8 & \textbf{82.5 }\\
            \rowcolor{RedOrange!8}
            $\beta=0.25$ & \textbf{58.3} & \textbf{73.1} & 82.1 \\
            $\beta=0.5$ & 58.1 & 72.5 & 81.4 \\
            \bottomrule
        \end{tabular}
    \end{minipage}%
    \hspace{0.05\linewidth}%
    \begin{minipage}{0.32\linewidth}
        \centering
        \caption*{(e) \textbf{Regularization Term.}}
        \vspace{-2ex}
        \begin{tabular}{@{}c|c|c|c@{}}
            \toprule
            Method & Transfer & Avg. & Last \\
            \midrule
            $l_2$ & 57.2 & 72.9 & 81.8 \\
            EWC & 57.2 & 71.3 & 80.6 \\
            \rowcolor{RedOrange!8}
            DFG & \textbf{58.3} & \textbf{73.1} & \textbf{82.1} \\
            \bottomrule
        \end{tabular}
    \end{minipage}%
    \hspace{0.05\linewidth}%
    \begin{minipage}{0.32\linewidth}
        \centering
        \caption*{(f) \textbf{Retrieval Method.}}
        \vspace{-2ex}
        \begin{tabular}{@{}c|c|c|c@{}}
            \toprule
            Method & Transfer & Avg. & Last \\
            \midrule
            BM25 & 55.1 & 72.2 & 81.8 \\
            Embedding & 56.1 & 72.5 & 81.5 \\
            \rowcolor{RedOrange!8}
            Hierarchical & \textbf{58.3} & \textbf{73.1} & \textbf{82.1} \\
            \bottomrule
        \end{tabular}
    \end{minipage}%
    }
    \label{tab4}
    \vspace{-0em}
\end{table*}

\begin{figure*}[tp]
\centering
 \includegraphics[width=0.96\linewidth]{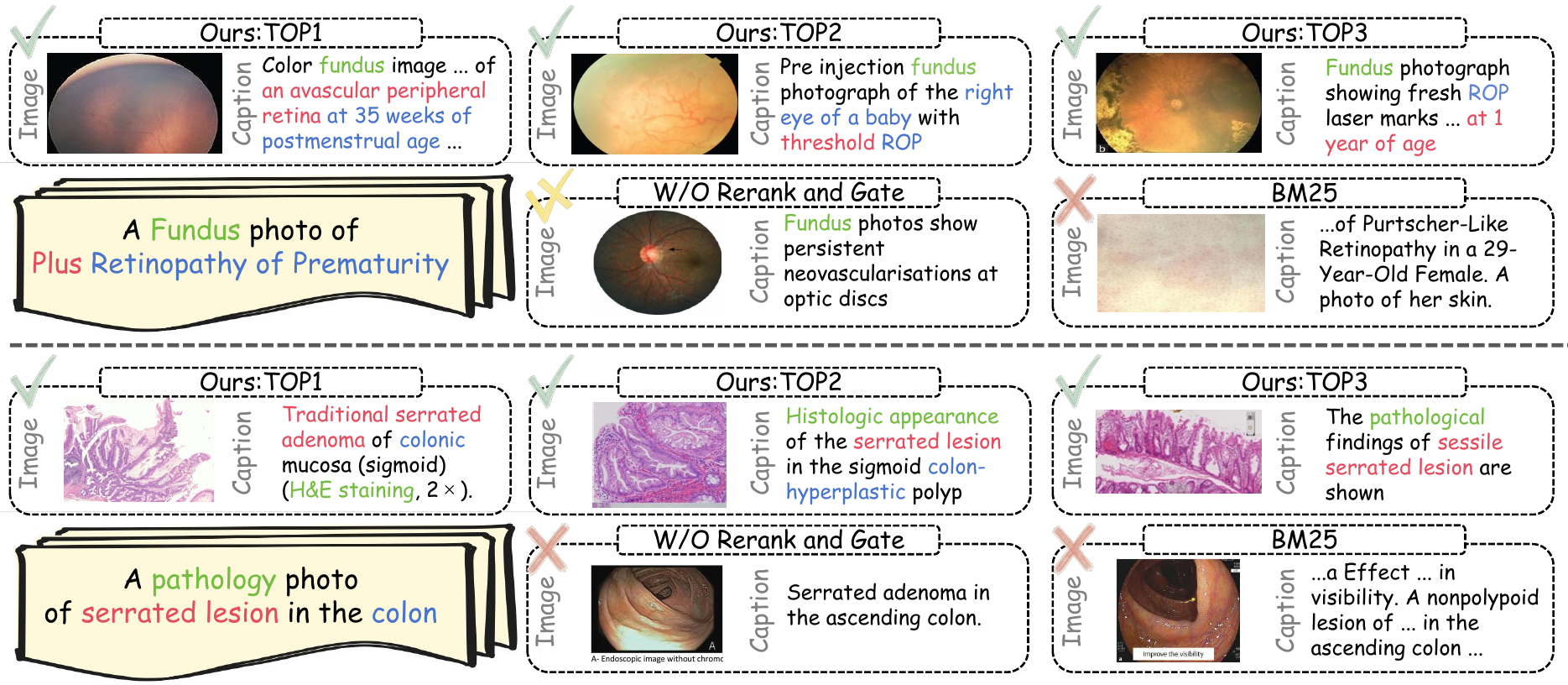}
\caption{In our qualitative visualization for retrieval using two prompts, we showcase the \textbf{Top-3 results} of our proposed method, comparing them against an \textbf{ablation variant} without the Rerank and Gate way and the pure \textbf{BM25} baseline. Our approach offers two significant advantages: 1. It avoids spurious matches that focus merely on keywords while failing to capture the correct domain context. 2. It is capable of retrieving more fine-grained matches related to lesion characteristics and severity, moving beyond superficial textual similarity.}
\label{fig5}
\end{figure*}

\subsection{Retrieval Visualization}

Our method demonstrates two significant retrieval advantages. First, it mitigates spurious textual correlations, preventing false pairings. For example, a naive keyword-based method might erroneously pair pathology and endoscopy images via the term \textbf{``colon''}—a fundamental domain mismatch that precludes effective learning. Second, our approach transcends simple keyword matching. Although \textbf{``Retinopathy of Prematurity''} is a pediatric condition, the term lacks explicit pediatric keywords. Our method nonetheless retrieves the semantically correct match, proving robust in the absence of direct keyword overlap.

\section{Conclusion}
\label{sec:conclusion}

In this paper, we propose PRIMED, a novel Continual Learning (CL) framework for Generalist Medical Foundation Models. To facilitate dataset curation, we have developed a new retrieval library and question pool. A complete pipeline encompassing retrieval, distillation, alignment, and guarding is dynamically implemented by PRIMED, enhancing its adaptability to the true data distribution of medical imaging and the realities of clinical practice. Comprehensive experiments demonstrate that our method achieves optimal performance in domain spanning, micro-level feature capture, and hard task retention. Methodologically, it explores the relationship between retrieval and distillation as two distinct forms of model memory. Clinically, this method can be utilized for the real-time updating of medical diagnostic systems. From a data ethics perspective, we have audited the relevant data for ethical and copyright compliance, and will release all associated content upon secondary confirmation. We provide a further discussion on limitations and ethical considerations in the \textbf{Appendix}.

\clearpage

\bibliographystyle{plainnat}
\bibliography{main}

\begin{thebibliography}{93}
\providecommand{\natexlab}[1]{#1}
\providecommand{\url}[1]{\texttt{#1}}
\expandafter\ifx\csname urlstyle\endcsname\relax
  \providecommand{\doi}[1]{doi: #1}\else
  \providecommand{\doi}{doi: \begingroup \urlstyle{rm}\Url}\fi

\bibitem[Akbari et~al.(2024)Akbari, Pourreza, Khalili~Pour, Dastjani~Farahani, Bazvand, Ebrahimiadib, Imani~Fooladi, and Ramazani~K]{FARFUM-RoP}
Morteza Akbari, Hamid-Reza Pourreza, Elias Khalili~Pour, Afsar Dastjani~Farahani, Fatemeh Bazvand, Nazanin Ebrahimiadib, Marjan Imani~Fooladi, and Fereshteh Ramazani~K.
\newblock Farfum-rop, a dataset for computer-aided detection of retinopathy of prematurity.
\newblock \emph{Scientific Data}, 11\penalty0 (1):\penalty0 1176, 2024.

\bibitem[Aljundi et~al.(2018)Aljundi, Babiloni, Elhoseiny, Rohrbach, and Tuytelaars]{MAS}
Rahaf Aljundi, Francesca Babiloni, Mohamed Elhoseiny, Marcus Rohrbach, and Tinne Tuytelaars.
\newblock Memory aware synapses: Learning what (not) to forget.
\newblock In \emph{Proceedings of the European conference on computer vision (ECCV)}, pages 139--154, 2018.

\bibitem[Asraf and Islam(2021)]{PD}
A~Asraf and Z~Islam.
\newblock Covid19, pneumonia and normal chest x-ray pa dataset. mendeley data v1 (2021), 2021.

\bibitem[Baghbanzadeh et~al.(2025{\natexlab{a}})Baghbanzadeh, Ashkezari, Dolatabadi, and Afkanpour]{Open-PMC-18M}
Negin Baghbanzadeh, Sajad Ashkezari, Elham Dolatabadi, and Arash Afkanpour.
\newblock Open-pmc-18m: A high-fidelity large scale medical dataset for multimodal representation learning.
\newblock \emph{arXiv preprint arXiv:2506.02738}, 2025{\natexlab{a}}.

\bibitem[Baghbanzadeh et~al.(2025{\natexlab{b}})Baghbanzadeh, Fallahpour, Parhizkar, et~al.]{PMC2.2M}
Negin Baghbanzadeh, Adibvafa Fallahpour, Yasaman Parhizkar, et~al.
\newblock Advancing medical representation learning through high-quality data.
\newblock In \emph{International Conference on Medical Image Computing and Computer-Assisted Intervention}, pages 24--33. Springer, 2025{\natexlab{b}}.

\bibitem[Bayasi et~al.(2021)Bayasi, Hamarneh, and Garbi]{culprit}
Nourhan Bayasi, Ghassan Hamarneh, and Rafeef Garbi.
\newblock Culprit-prune-net: Efficient continual sequential multi-domain learning with application to skin lesion classification.
\newblock In \emph{International Conference on Medical Image Computing and Computer-Assisted Intervention}, pages 165--175. Springer, 2021.

\bibitem[Bayasi et~al.(2023)Bayasi, Du, Hamarneh, and Garbi]{Continual-GEN_MICCAI_skin}
Nourhan Bayasi, Siyi Du, Ghassan Hamarneh, and Rafeef Garbi.
\newblock Continual-gen: Continual group ensembling for domain-agnostic skin lesion classification.
\newblock In \emph{International Conference on Medical Image Computing and Computer-Assisted Intervention}, pages 3--13. Springer, 2023.

\bibitem[Bayasi et~al.(2024)Bayasi, Fayyad, Bissoto, Hamarneh, and Garbi]{Biaspruner_MICCAI_CT}
Nourhan Bayasi, Jamil Fayyad, Alceu Bissoto, Ghassan Hamarneh, and Rafeef Garbi.
\newblock Biaspruner: Debiased continual learning for medical image classification.
\newblock In \emph{International Conference on Medical Image Computing and Computer-Assisted Intervention}, pages 90--101. Springer, 2024.

\bibitem[Bhave et~al.(2024)Bhave, Rodriguez, Poterucha, Mutasa, Aberle, et~al.]{CheXchoNet}
Shreyas Bhave, Victor Rodriguez, Timothy Poterucha, Simukayi Mutasa, Dwight Aberle, et~al.
\newblock Deep learning to detect left ventricular structural abnormalities in chest x-rays.
\newblock \emph{European heart journal}, 45\penalty0 (22):\penalty0 2002--2012, 2024.

\bibitem[Buzzega et~al.(2020)Buzzega, Boschini, Porrello, Abati, and Calderara]{Dark_distill}
Pietro Buzzega, Matteo Boschini, Angelo Porrello, Davide Abati, and Simone Calderara.
\newblock Dark experience for general continual learning: a strong, simple baseline.
\newblock \emph{Advances in neural information processing systems}, 33:\penalty0 15920--15930, 2020.

\bibitem[Castro et~al.(2018)Castro, Mar{\'\i}n-Jim{\'e}nez, Guil, Schmid, and Alahari]{E2ECL}
Francisco~M Castro, Manuel~J Mar{\'\i}n-Jim{\'e}nez, Nicol{\'a}s Guil, Cordelia Schmid, and Karteek Alahari.
\newblock End-to-end incremental learning.
\newblock In \emph{Proceedings of the European conference on computer vision (ECCV)}, pages 233--248, 2018.

\bibitem[Cha et~al.(2021)Cha, Lee, and Shin]{Co2l}
Hyuntak Cha, Jaeho Lee, and Jinwoo Shin.
\newblock Co2l: Contrastive continual learning.
\newblock In \emph{Proceedings of the IEEE/CVF International conference on computer vision}, pages 9516--9525, 2021.

\bibitem[Chakraborti et~al.(2021)Chakraborti, Gleeson, and Rittscher]{patho_finegrain}
Tapabrata Chakraborti, Fergus Gleeson, and Jens Rittscher.
\newblock Contrastive representations for continual learning of fine-grained histology images.
\newblock In \emph{International Workshop on Machine Learning in Medical Imaging}, pages 1--9. Springer, 2021.

\bibitem[Chaudhry et~al.(2018)Chaudhry, Dokania, Ajanthan, and Torr]{forget}
Arslan Chaudhry, Puneet~K Dokania, Thalaiyasingam Ajanthan, and Philip~HS Torr.
\newblock Riemannian walk for incremental learning: Understanding forgetting and intransigence.
\newblock In \emph{Proceedings of the European conference on computer vision (ECCV)}, pages 532--547, 2018.

\bibitem[Chen et~al.(2024{\natexlab{a}})Chen, Xiao, Zhang, Luo, Lian, and Liu]{bge}
Jianlv Chen, Shitao Xiao, Peitian Zhang, Kun Luo, Defu Lian, and Zheng Liu.
\newblock Bge m3-embedding: Multi-lingual, multi-functionality, multi-granularity text embeddings through self-knowledge distillation.
\newblock \emph{arXiv preprint arXiv:2402.03216}, 2024{\natexlab{a}}.

\bibitem[Chen et~al.(2024{\natexlab{b}})Chen, Du, You, Islam, Guo, Jin, Chen, and Heng]{ICRA_surgery}
Kexin Chen, Yuyang Du, Tao You, Mobarakol Islam, Ziyu Guo, Yueming Jin, Guangyong Chen, and Pheng-Ann Heng.
\newblock Llm-assisted multi-teacher continual learning for visual question answering in robotic surgery.
\newblock In \emph{2024 IEEE International Conference on Robotics and Automation (ICRA)}, pages 10772--10778. IEEE, 2024{\natexlab{b}}.

\bibitem[Chen et~al.(2025)Chen, Xu, Sellergren, Matias, Hassidim, Shetty, Golden, Yuille, and Yang]{CoCa-CXR}
Yixiong Chen, Shawn Xu, Andrew Sellergren, Yossi Matias, Avinatan Hassidim, Shravya Shetty, Daniel Golden, Alan~L Yuille, and Lin Yang.
\newblock Coca-cxr: Co ntrastive ca ptioners learn strong temporal structures for chest x-ray vision-language understanding.
\newblock In \emph{International Conference on Medical Image Computing and Computer-Assisted Intervention}, pages 78--88. Springer, 2025.

\bibitem[Chicco et~al.(2021)Chicco, Warrens, and Jurman]{MSE}
Davide Chicco, Matthijs~J Warrens, and Giuseppe Jurman.
\newblock The coefficient of determination r-squared is more informative than smape, mae, mape, mse and rmse in regression analysis evaluation.
\newblock \emph{Peerj computer science}, 7:\penalty0 e623, 2021.

\bibitem[de~Vente et~al.(2024)de~Vente, Vermeer, and I.]{AIROGS}
Coen de~Vente, Koenraad~A.Sánchez Vermeer, and Clara I.
\newblock Airogs: Artificial intelligence for robust glaucoma screening challenge.
\newblock \emph{IEEE Transactions on Medical Imaging}, 43\penalty0 (1):\penalty0 542--557, 2024.
\newblock \doi{10.1109/TMI.2023.3313786}.

\bibitem[Deng et~al.(2009)Deng, Dong, Socher, Li, Li, and Fei-Fei]{imagenet}
Jia Deng, Wei Dong, Richard Socher, Li-Jia Li, Kai Li, and Li~Fei-Fei.
\newblock Imagenet: A large-scale hierarchical image database.
\newblock In \emph{2009 IEEE conference on computer vision and pattern recognition}, pages 248--255. Ieee, 2009.

\bibitem[Devlin et~al.(2019)Devlin, Chang, Lee, and Toutanova]{bert}
Jacob Devlin, Ming-Wei Chang, Kenton Lee, and Kristina Toutanova.
\newblock Bert: Pre-training of deep bidirectional transformers for language understanding.
\newblock In \emph{Proceedings of the 2019 conference of the North American chapter of the association for computational linguistics: human language technologies, volume 1 (long and short papers)}, pages 4171--4186, 2019.

\bibitem[D{\'\i}az-Rodr{\'\i}guez et~al.(2018)D{\'\i}az-Rodr{\'\i}guez, Lomonaco, Filliat, and Maltoni]{BWT}
Natalia D{\'\i}az-Rodr{\'\i}guez, Vincenzo Lomonaco, David Filliat, and Davide Maltoni.
\newblock Don't forget, there is more than forgetting: new metrics for continual learning.
\newblock \emph{arXiv preprint arXiv:1810.13166}, 2018.

\bibitem[Dosovitskiy(2020)]{VIT}
Alexey Dosovitskiy.
\newblock An image is worth 16x16 words: Transformers for image recognition at scale.
\newblock \emph{arXiv preprint arXiv:2010.11929}, 2020.

\bibitem[Douillard et~al.(2022)Douillard, Ram{\'e}, Couairon, and Cord]{Dytox_arch}
Arthur Douillard, Alexandre Ram{\'e}, Guillaume Couairon, and Matthieu Cord.
\newblock Dytox: Transformers for continual learning with dynamic token expansion.
\newblock In \emph{Proceedings of the IEEE/CVF conference on computer vision and pattern recognition}, pages 9285--9295, 2022.

\bibitem[Fang et~al.(2024)Fang, Liu, Liu, Wang, Su, Zhang, Kong, Yi, and Wu]{ICRA_surgery_2}
Senlin Fang, Yiwen Liu, Chengliang Liu, Jingnan Wang, Yuanzhe Su, Yupo Zhang, Hoiio Kong, Zhengkun Yi, and Xinyu Wu.
\newblock Probabilistic spiking neural network for robotic tactile continual learning.
\newblock In \emph{2024 IEEE International Conference on Robotics and Automation (ICRA)}, pages 530--536. IEEE, 2024.

\bibitem[Fini et~al.(2022)Fini, Da~Costa, Alameda-Pineda, Ricci, Alahari, and Mairal]{Self-supervised_distill}
Enrico Fini, Victor G~Turrisi Da~Costa, Xavier Alameda-Pineda, Elisa Ricci, Karteek Alahari, and Julien Mairal.
\newblock Self-supervised models are continual learners.
\newblock In \emph{Proceedings of the IEEE/CVF conference on computer vision and pattern recognition}, pages 9621--9630, 2022.

\bibitem[Gao and Liu(2023)]{ddgr}
Rui Gao and Weiwei Liu.
\newblock Ddgr: Continual learning with deep diffusion-based generative replay.
\newblock In \emph{International Conference on Machine Learning}, pages 10744--10763. PMLR, 2023.

\bibitem[Gou et~al.(2025)Gou, Ji, Liu, and Ye]{QPMIL}
Jiaxiang Gou, Luping Ji, Pei Liu, and Mao Ye.
\newblock Queryable prototype multiple instance learning with vision-language models for incremental whole slide image classification.
\newblock In \emph{Proceedings of the AAAI Conference on Artificial Intelligence}, volume~39, pages 3158--3166, 2025.

\bibitem[Gu et~al.(2021)Gu, Tinn, Cheng, Lucas, Usuyama, Liu, Naumann, Gao, and Poon]{PubmedBert}
Yu~Gu, Robert Tinn, Hao Cheng, Michael Lucas, Naoto Usuyama, Xiaodong Liu, Tristan Naumann, Jianfeng Gao, and Hoifung Poon.
\newblock Domain-specific language model pretraining for biomedical natural language processing.
\newblock \emph{ACM Transactions on Computing for Healthcare (HEALTH)}, 3\penalty0 (1):\penalty0 1--23, 2021.

\bibitem[Gulshan et~al.(2016)Gulshan, Peng, et~al.]{eyepacs}
Varun Gulshan, Lily Peng, et~al.
\newblock Development and validation of a deep learning algorithm for detection of diabetic retinopathy in retinal fundus photographs.
\newblock \emph{jama}, 316\penalty0 (22):\penalty0 2402--2410, 2016.

\bibitem[Hendrycks et~al.(2021)Hendrycks, Basart, Mu, Kadavath, Wang, et~al.]{face}
Dan Hendrycks, Steven Basart, Norman Mu, Saurav Kadavath, Frank Wang, et~al.
\newblock The many faces of robustness: A critical analysis of out-of-distribution generalization.
\newblock In \emph{Proceedings of the IEEE/CVF international conference on computer vision}, pages 8340--8349, 2021.

\bibitem[Hoerl and Kennard(1970)]{l2_norm}
Arthur~E Hoerl and Robert~W Kennard.
\newblock Ridge regression: Biased estimation for nonorthogonal problems.
\newblock \emph{Technometrics}, 12\penalty0 (1):\penalty0 55--67, 1970.

\bibitem[Hu et~al.(2022{\natexlab{a}})Hu, Shen, Wallis, et~al.]{lora}
Edward~J Hu, Yelong Shen, Phillip Wallis, et~al.
\newblock Lora: Low-rank adaptation of large language models.
\newblock \emph{ICLR}, 1\penalty0 (2):\penalty0 3, 2022{\natexlab{a}}.

\bibitem[Hu et~al.(2022{\natexlab{b}})Hu, Li, Li, et~al.]{Nucls}
Weiming Hu, Chen Li, Xiaoyan Li, et~al.
\newblock Gashissdb: A new gastric histopathology image dataset for computer aided diagnosis of gastric cancer.
\newblock \emph{Computers in biology and medicine}, 142:\penalty0 105207, 2022{\natexlab{b}}.

\bibitem[Huang et~al.(2023)Huang, Zhao, Wang, Fu, Jiang, and Yu]{conslide}
Yanyan Huang, Weiqin Zhao, Shujun Wang, Yu~Fu, Yuming Jiang, and Lequan Yu.
\newblock Conslide: Asynchronous hierarchical interaction transformer with breakup-reorganize rehearsal for continual whole slide image analysis.
\newblock In \emph{Proceedings of the IEEE/CVF International Conference on Computer Vision}, pages 21349--21360, 2023.

\bibitem[Jia et~al.(2021)Jia, Yang, Xia, Chen, Parekh, Pham, Le, Sung, Li, and Duerig]{representation_with_noisy_text_supervision}
Chao Jia, Yinfei Yang, Ye~Xia, Yi-Ting Chen, Zarana Parekh, Hieu Pham, Quoc Le, Yun-Hsuan Sung, Zhen Li, and Tom Duerig.
\newblock Scaling up visual and vision-language representation learning with noisy text supervision.
\newblock In \emph{International conference on machine learning}, pages 4904--4916. PMLR, 2021.

\bibitem[Jia et~al.(2022)Jia, Tang, Chen, Cardie, Belongie, Hariharan, and Lim]{Visual_prompt_tuning}
Menglin Jia, Luming Tang, Bor-Chun Chen, Claire Cardie, Serge Belongie, Bharath Hariharan, and Ser-Nam Lim.
\newblock Visual prompt tuning.
\newblock In \emph{European conference on computer vision}, pages 709--727. Springer, 2022.

\bibitem[Kather et~al.(2018)Kather, Halama, and Marx]{NCT}
Jakob~Nikolas Kather, Niels Halama, and Alexander Marx.
\newblock {100,000 histological images of human colorectal cancer and healthy tissue}, April 2018.
\newblock URL \url{https://doi.org/10.5281/zenodo.1214456}.

\bibitem[Khattak et~al.(2024)Khattak, Kunhimon, Naseer, Khan, and Khan]{Unimed-clip}
Muhammad~Uzair Khattak, Shahina Kunhimon, Muzammal Naseer, Salman Khan, and Fahad~Shahbaz Khan.
\newblock Unimed-clip: Towards a unified image-text pretraining paradigm for diverse medical imaging modalities.
\newblock \emph{arXiv preprint arXiv:2412.10372}, 2024.

\bibitem[Kirkpatrick et~al.(2017)Kirkpatrick, Pascanu, Rabinowitz, Veness, Desjardins, Rusu, Milan, Quan, Ramalho, Grabska-Barwinska, et~al.]{ocf}
James Kirkpatrick, Razvan Pascanu, Neil Rabinowitz, Joel Veness, Guillaume Desjardins, Andrei~A Rusu, Kieran Milan, John Quan, Tiago Ramalho, Agnieszka Grabska-Barwinska, et~al.
\newblock Overcoming catastrophic forgetting in neural networks.
\newblock \emph{Proceedings of the national academy of sciences}, 114\penalty0 (13):\penalty0 3521--3526, 2017.

\bibitem[Kurle et~al.(2019)Kurle, Cseke, Klushyn, Van Der~Smagt, and G{\"u}nnemann]{CL_non-stationary_data}
Richard Kurle, Botond Cseke, Alexej Klushyn, Patrick Van Der~Smagt, and Stephan G{\"u}nnemann.
\newblock Continual learning with bayesian neural networks for non-stationary data.
\newblock In \emph{International Conference on Learning Representations}, 2019.

\bibitem[Kurtansky et~al.(2024)Kurtansky, Rotemberg, Gillis, Kose, Reade, and Chow]{isic-2024}
Nicholas Kurtansky, Veronica Rotemberg, Maura Gillis, Kivanc Kose, Walter Reade, and Ashley Chow.
\newblock Isic 2024 - skin cancer detection with 3d-tbp, 2024.
\newblock URL \url{https://kaggle.com/competitions/isic-2024-challenge}.

\bibitem[Le et~al.(2024)Le, Nguyen, Nguyen, Nguyen, Pham, Van~Ngo, and Ho]{MOE_prompts}
Minh Le, An~Nguyen, Huy Nguyen, Trang Nguyen, Trang Pham, Linh Van~Ngo, and Nhat Ho.
\newblock Mixture of experts meets prompt-based continual learning.
\newblock \emph{Advances in Neural Information Processing Systems}, 37:\penalty0 119025--119062, 2024.

\bibitem[Lee et~al.(2025)Lee, Jeong, Han, Lee, and Chun]{CoMEL}
Byung~Hyun Lee, Wongi Jeong, Woojae Han, Kyoungbun Lee, and Se~Young Chun.
\newblock Continual multiple instance learning with enhanced localization for histopathological whole slide image analysis.
\newblock \emph{arXiv preprint arXiv:2507.02395}, 2025.

\bibitem[Lee et~al.(2020)Lee, Yoon, Kim, Kim, Kim, So, and Kang]{BioBERT}
Jinhyuk Lee, Wonjin Yoon, Sungdong Kim, Donghyeon Kim, Sunkyu Kim, Chan~Ho So, and Jaewoo Kang.
\newblock Biobert: a pre-trained biomedical language representation model for biomedical text mining.
\newblock \emph{Bioinformatics}, 36\penalty0 (4):\penalty0 1234--1240, 2020.

\bibitem[Lewis et~al.(2020)Lewis, Perez, et~al.]{RAG}
Patrick Lewis, Ethan Perez, et~al.
\newblock Retrieval-augmented generation for knowledge-intensive nlp tasks.
\newblock \emph{Advances in neural information processing systems}, 33:\penalty0 9459--9474, 2020.

\bibitem[Li et~al.(2025)Li, Zhang, Zhou, Yang, Xie, and He]{CLMS_MIA_MRI}
Weilu Li, Yun Zhang, Hao Zhou, Wenhan Yang, Zhi Xie, and Yao He.
\newblock Clms: Bridging domain gaps in medical imaging segmentation with source-free continual learning for robust knowledge transfer and adaptation.
\newblock \emph{Medical Image Analysis}, 100:\penalty0 103404, 2025.

\bibitem[Li and Hoiem(2017)]{lwf}
Zhizhong Li and Derek Hoiem.
\newblock Learning without forgetting.
\newblock \emph{IEEE transactions on pattern analysis and machine intelligence}, 40\penalty0 (12):\penalty0 2935--2947, 2017.

\bibitem[Lin et~al.(2025)Lin, Long, Wan, Wang, Yang, Yang, Yao, Chen, Guo, Li, et~al.]{SAIL-Embedding}
Lin Lin, Jiefeng Long, Zhihe Wan, Yuchi Wang, Dingkang Yang, Shuang Yang, Yueyang Yao, Xu~Chen, Zirui Guo, Shengqiang Li, et~al.
\newblock Sail-embedding technical report: Omni-modal embedding foundation model.
\newblock \emph{arXiv preprint arXiv:2510.12709}, 2025.

\bibitem[Lin et~al.(2023)Lin, Zhao, Zhang, Wu, Zhang, Wang, and Xie]{Pmc-clip}
Weixiong Lin, Ziheng Zhao, Xiaoman Zhang, Chaoyi Wu, Ya~Zhang, Yanfeng Wang, and Weidi Xie.
\newblock Pmc-clip: Contrastive language-image pre-training using biomedical documents.
\newblock In \emph{International Conference on Medical Image Computing and Computer-Assisted Intervention}, pages 525--536. Springer, 2023.

\bibitem[Liu et~al.(2022)Liu, Li, Zhang, Yang, Qi, Su, Zhu, and Zhang]{Dab-detr}
Shilong Liu, Feng Li, Hao Zhang, Xiao Yang, Xianbiao Qi, Hang Su, Jun Zhu, and Lei Zhang.
\newblock Dab-detr: Dynamic anchor boxes are better queries for detr.
\newblock \emph{arXiv preprint arXiv:2201.12329}, 2022.

\bibitem[Lozano et~al.(2025)Lozano, Sun, et~al.]{biomedica}
Alejandro Lozano, Min~Woo Sun, et~al.
\newblock Biomedica: An open biomedical image-caption archive, dataset, and vision-language models derived from scientific literature.
\newblock In \emph{Proceedings of the Computer Vision and Pattern Recognition Conference}, pages 19724--19735, 2025.

\bibitem[Matek et~al.(2021)Matek, Krappe, M{\"u}nzenmayer, Haferlach, and Marr]{BMC}
Christian Matek, Sebastian Krappe, Christian M{\"u}nzenmayer, Torsten Haferlach, and Carsten Marr.
\newblock Highly accurate differentiation of bone marrow cell morphologies using deep neural networks on a large image data set.
\newblock \emph{Blood, The Journal of the American Society of Hematology}, 138\penalty0 (20):\penalty0 1917--1927, 2021.

\bibitem[Meng et~al.(2024)Meng, Zhang, Yang, Zhan, Zhao, and Wang]{diffclass}
Zichong Meng, Jie Zhang, Changdi Yang, Zheng Zhan, Pu~Zhao, and Yanzhi Wang.
\newblock Diffclass: Diffusion-based class incremental learning.
\newblock In \emph{European Conference on Computer Vision}, pages 142--159. Springer, 2024.

\bibitem[Oquab et~al.(2023)Oquab, Darcet, Moutakanni, Vo, Szafraniec, Khalidov, Fernandez, Haziza, Massa, El-Nouby, et~al.]{dinov2}
Maxime Oquab, Timoth{\'e}e Darcet, Th{\'e}o Moutakanni, Huy Vo, Marc Szafraniec, Vasil Khalidov, Pierre Fernandez, Daniel Haziza, Francisco Massa, Alaaeldin El-Nouby, et~al.
\newblock Dinov2: Learning robust visual features without supervision.
\newblock \emph{arXiv preprint arXiv:2304.07193}, 2023.

\bibitem[Peng et~al.(2025)Peng, Liu, Yang, Hong, and Tian]{gnsp}
Tiantian Peng, Yuyang Liu, Shuo Yang, Qiuhe Hong, and YongHong Tian.
\newblock Gnsp: Gradient null space projection for preserving cross-modal alignment in vlms continual learning.
\newblock \emph{arXiv preprint arXiv:2507.19839}, 2025.

\bibitem[Perkonigg et~al.(2021)Perkonigg, Hofmanninger, Herold, Brink, Pianykh, Prosch, and Langs]{NC_MRI}
Matthias Perkonigg, Johannes Hofmanninger, Christian~J Herold, James~A Brink, Oleg Pianykh, Helmut Prosch, and Georg Langs.
\newblock Dynamic memory to alleviate catastrophic forgetting in continual learning with medical imaging.
\newblock \emph{Nature communications}, 12\penalty0 (1):\penalty0 5678, 2021.

\bibitem[Pham et~al.(2021)Pham, Liu, and Hoi]{dualnet_arch}
Quang Pham, Chenghao Liu, and Steven Hoi.
\newblock Dualnet: Continual learning, fast and slow.
\newblock \emph{Advances in Neural Information Processing Systems}, 34:\penalty0 16131--16144, 2021.

\bibitem[Radford et~al.(2021)Radford, Kim, Hallacy, Ramesh, Goh, Agarwal, et~al.]{CLIP}
Alec Radford, Jong~Wook Kim, Chris Hallacy, Aditya Ramesh, Gabriel Goh, Sandhini Agarwal, et~al.
\newblock Learning transferable visual models from natural language supervision.
\newblock In \emph{International conference on machine learning}, pages 8748--8763. PmLR, 2021.

\bibitem[Rashid et~al.(2024)Rashid, Sharmin, Khatun, Hasan, and Uddin]{AOD}
Mohammad~Riadur Rashid, Shayla Sharmin, Tania Khatun, Md~Zahid Hasan, and Mohammad~Shorif Uddin.
\newblock {Eye Disease Image Dataset}, 2024.
\newblock URL \url{https://data.mendeley.com/datasets/s9bfhswzjb/1}.

\bibitem[Rebuffi et~al.(2017)Rebuffi, Kolesnikov, Sperl, and Lampert]{icarl}
Sylvestre-Alvise Rebuffi, Alexander Kolesnikov, Georg Sperl, and Christoph~H Lampert.
\newblock icarl: Incremental classifier and representation learning.
\newblock In \emph{Proceedings of the IEEE conference on Computer Vision and Pattern Recognition}, pages 2001--2010, 2017.

\bibitem[Recht et~al.(2019)Recht, Roelofs, Schmidt, and Shankar]{imagenet_classifier}
Benjamin Recht, Rebecca Roelofs, Ludwig Schmidt, and Vaishaal Shankar.
\newblock Do imagenet classifiers generalize to imagenet?
\newblock In \emph{International conference on machine learning}, pages 5389--5400. PMLR, 2019.

\bibitem[Robertson et~al.(2009)Robertson, Zaragoza, et~al.]{BM25}
Stephen Robertson, Hugo Zaragoza, et~al.
\newblock The probabilistic relevance framework: Bm25 and beyond.
\newblock \emph{Foundations and Trends{\textregistered} in Information Retrieval}, 3\penalty0 (4):\penalty0 333--389, 2009.

\bibitem[{Royal Australian and New Zealand College of Radiologists (RANZCR) and Kaggle}(2021)]{ranzcr}
{Royal Australian and New Zealand College of Radiologists (RANZCR) and Kaggle}.
\newblock {RANZCR CLiP - Catheter and Line Position Challenge}.
\newblock Kaggle Competition, 2021.
\newblock URL \url{https://www.kaggle.com/competitions/ranzcr-clip-catheter-line-classification}.

\bibitem[Spanhol et~al.(2015)Spanhol, Oliveira, Petitjean, and Heutte]{breakhis}
Fabio~A Spanhol, Luiz~S Oliveira, Caroline Petitjean, and Laurent Heutte.
\newblock A dataset for breast cancer histopathological image classification.
\newblock \emph{Ieee transactions on biomedical engineering}, 63\penalty0 (7):\penalty0 1455--1462, 2015.

\bibitem[Speidel et~al.(2023)Speidel, Maier-Hein, Stoyanov, and Giannarou]{Pitvis}
Stefanie Speidel, Lena Maier-Hein, Danail Stoyanov, and Stamatia Giannarou.
\newblock Endoscopic vision challenge 2023, September 2023.
\newblock URL \url{https://doi.org/10.5281/zenodo.8315050}.

\bibitem[Srivastava et~al.(2021)Srivastava, Yaqub, Nandakumar, Ge, and Mahapatra]{CL_MICCAIw_X_ray}
Shikhar Srivastava, Mohammad Yaqub, Karthik Nandakumar, Zongyuan Ge, and Dwarikanath Mahapatra.
\newblock Continual domain incremental learning for chest x-ray classification in low-resource clinical settings.
\newblock In \emph{MICCAI Workshop on Domain Adaptation and Representation Transfer}, pages 226--238. Springer, 2021.

\bibitem[Sun et~al.(2025)Sun, Lozano, Tejero, Nath, et~al.]{BMC-CLIP-long}
Min~Woo Sun, Alejandro Lozano, Javier~Gamazo Tejero, Vishwesh Nath, et~al.
\newblock No tokens wasted: Leveraging long context in biomedical vision-language models.
\newblock \emph{arXiv preprint arXiv:2510.03978}, 2025.

\bibitem[Tang et~al.(2024)Tang, Tian, Li, He, Zhou, Zhao, Li, and Jia]{DIKI}
Longxiang Tang, Zhuotao Tian, Kai Li, Chunming He, Hantao Zhou, Hengshuang Zhao, Xiu Li, and Jiaya Jia.
\newblock Mind the interference: Retaining pre-trained knowledge in parameter efficient continual learning of vision-language models.
\newblock In \emph{European conference on computer vision}, pages 346--365. Springer, 2024.

\bibitem[Tasai et~al.(2025)Tasai, Li, Togo, Tang, Yoshimura, Sugimori, Hirata, Ogawa, Kudo, and Haseyama]{CSSL}
Ren Tasai, Guang Li, Ren Togo, Minghui Tang, Takaaki Yoshimura, Hiroyuki Sugimori, Kenji Hirata, Takahiro Ogawa, Kohsuke Kudo, and Miki Haseyama.
\newblock Continual self-supervised learning considering medical domain knowledge in chest ct images.
\newblock In \emph{ICASSP 2025-2025 IEEE International Conference on Acoustics, Speech and Signal Processing (ICASSP)}, pages 1--5. IEEE, 2025.

\bibitem[Wang et~al.(2025)Wang, Jin, Hu, Safari, Zhao, Chang, Qiu, Roper, Yu, and Yang]{MMKD-CLIP}
Shansong Wang, Zhecheng Jin, Mingzhe Hu, Mojtaba Safari, Feng Zhao, Chih-Wei Chang, Richard~LJ Qiu, Justin Roper, David~S Yu, and Xiaofeng Yang.
\newblock Unifying biomedical vision-language expertise: Towards a generalist foundation model via multi-clip knowledge distillation.
\newblock \emph{arXiv preprint arXiv:2506.22567}, 2025.

\bibitem[Wang et~al.(2017)Wang, Peng, Lu, Lu, Bagheri, and Summers]{NIH-Chest-ray}
Xiaosong Wang, Yifan Peng, Le~Lu, Zhiyong Lu, Mohammadhadi Bagheri, and Ronald~M Summers.
\newblock Chestx-ray8: Hospital-scale chest x-ray database and benchmarks on weakly-supervised classification and localization of common thorax diseases.
\newblock In \emph{Proceedings of the IEEE conference on computer vision and pattern recognition}, pages 2097--2106, 2017.

\bibitem[Wang et~al.(2022{\natexlab{a}})Wang, Shen, Duan, Zhan, Fang, and Gao]{long_task}
Zhenyi Wang, Li~Shen, Tiehang Duan, Donglin Zhan, Le~Fang, and Mingchen Gao.
\newblock Learning to learn and remember super long multi-domain task sequence.
\newblock In \emph{Proceedings of the IEEE/CVF Conference on Computer Vision and Pattern Recognition}, pages 7982--7992, 2022{\natexlab{a}}.

\bibitem[Wang et~al.(2022{\natexlab{b}})Wang, Zhang, Ebrahimi, Sun, Zhang, et~al.]{Dualprompt}
Zifeng Wang, Zizhao Zhang, Sayna Ebrahimi, Ruoxi Sun, Han Zhang, et~al.
\newblock Dualprompt: Complementary prompting for rehearsal-free continual learning.
\newblock In \emph{European conference on computer vision}, pages 631--648. Springer, 2022{\natexlab{b}}.

\bibitem[Wortsman et~al.(2022)Wortsman, Ilharco, Kim, Li, Kornblith, Roelofs, Lopes, Hajishirzi, Farhadi, Namkoong, et~al.]{wise_ft}
Mitchell Wortsman, Gabriel Ilharco, Jong~Wook Kim, Mike Li, Simon Kornblith, Rebecca Roelofs, Raphael~Gontijo Lopes, Hannaneh Hajishirzi, Ali Farhadi, Hongseok Namkoong, et~al.
\newblock Robust fine-tuning of zero-shot models.
\newblock In \emph{Proceedings of the IEEE/CVF conference on computer vision and pattern recognition}, pages 7959--7971, 2022.

\bibitem[Wu et~al.(2025)Wu, Shi, Wang, and Ye]{GIFT}
Bin Wu, Wuxuan Shi, Jinqiao Wang, and Mang Ye.
\newblock Synthetic data is an elegant gift for continual vision-language models.
\newblock In \emph{Proceedings of the Computer Vision and Pattern Recognition Conference}, pages 2813--2823, 2025.

\bibitem[Wu et~al.(2024)Wu, Lin, Zhang, Zhang, Xie, and Wang]{PMC-LLaMA}
Chaoyi Wu, Weixiong Lin, Xiaoman Zhang, Ya~Zhang, Weidi Xie, and Yanfeng Wang.
\newblock Pmc-llama: toward building open-source language models for medicine.
\newblock \emph{Journal of the American Medical Informatics Association}, 31\penalty0 (9):\penalty0 1833--1843, 2024.

\bibitem[Wu et~al.(2019)Wu, Chen, Wang, Ye, Liu, Guo, and Fu]{BiC}
Yue Wu, Yinpeng Chen, Lijuan Wang, Yuancheng Ye, Zicheng Liu, Yandong Guo, and Yun Fu.
\newblock Large scale incremental learning.
\newblock In \emph{Proceedings of the IEEE/CVF conference on computer vision and pattern recognition}, pages 374--382, 2019.

\bibitem[Yan et~al.(2021)Yan, Xie, and He]{Der}
Shipeng Yan, Jiangwei Xie, and Xuming He.
\newblock Der: Dynamically expandable representation for class incremental learning.
\newblock In \emph{Proceedings of the IEEE/CVF conference on computer vision and pattern recognition}, pages 3014--3023, 2021.

\bibitem[Ye et~al.(2024)Ye, Xie, Zhang, Chen, Wu, and Xia]{MedCoss}
Yiwen Ye, Yutong Xie, Jianpeng Zhang, Ziyang Chen, Qi~Wu, and Yong Xia.
\newblock Continual self-supervised learning: Towards universal multi-modal medical data representation learning.
\newblock In \emph{Proceedings of the IEEE/CVF conference on computer vision and pattern recognition}, pages 11114--11124, 2024.

\bibitem[Yosinski et~al.(2014)Yosinski, Clune, Bengio, and Lipson]{vis_trans}
Jason Yosinski, Jeff Clune, Yoshua Bengio, and Hod Lipson.
\newblock How transferable are features in deep neural networks?
\newblock \emph{Advances in neural information processing systems}, 27, 2014.

\bibitem[Yu et~al.(2022)Yu, Wang, Vasudevan, Yeung, Seyedhosseini, and Wu]{coca}
Jiahui Yu, Zirui Wang, Vijay Vasudevan, Legg Yeung, Mojtaba Seyedhosseini, and Yonghui Wu.
\newblock Coca: Contrastive captioners are image-text foundation models.
\newblock \emph{arXiv preprint arXiv:2205.01917}, 2022.

\bibitem[Yu et~al.(2024{\natexlab{a}})Yu, Zhuge, Zhang, Hu, Wang, Lu, and He]{MOE_CL}
Jiazuo Yu, Yunzhi Zhuge, Lu~Zhang, Ping Hu, Dong Wang, Huchuan Lu, and You He.
\newblock Boosting continual learning of vision-language models via mixture-of-experts adapters.
\newblock In \emph{Proceedings of the IEEE/CVF Conference on Computer Vision and Pattern Recognition}, pages 23219--23230, 2024{\natexlab{a}}.

\bibitem[Yu et~al.(2025)Yu, Huang, Zhuge, Zhang, Hu, Wang, Lu, and He]{MoE-Adapters++}
Jiazuo Yu, Zichen Huang, Yunzhi Zhuge, Lu~Zhang, Ping Hu, Dong Wang, Huchuan Lu, and You He.
\newblock Moe-adapters++: Towards more efficient continual learning of vision-language models via dynamic mixture-of-experts adapters.
\newblock \emph{IEEE Transactions on Pattern Analysis and Machine Intelligence}, 2025.

\bibitem[Yu et~al.(2024{\natexlab{b}})Yu, Huang, Chen, Chang, Lai, Yang, and Wang]{snd}
Yu-Chu Yu, Chi-Pin Huang, Jr-Jen Chen, Kai-Po Chang, Yung-Hsuan Lai, Fu-En Yang, and Yu-Chiang~Frank Wang.
\newblock Select and distill: Selective dual-teacher knowledge transfer for continual learning on vision-language models.
\newblock In \emph{European Conference on Computer Vision}, pages 219--236. Springer, 2024{\natexlab{b}}.

\bibitem[Zenke et~al.(2017)Zenke, Poole, and Ganguli]{si}
Friedemann Zenke, Ben Poole, and Surya Ganguli.
\newblock Continual learning through synaptic intelligence.
\newblock In \emph{International conference on machine learning}, pages 3987--3995. PMLR, 2017.

\bibitem[Zhang et~al.(2025{\natexlab{a}})Zhang, Liang, Kuang, Cen, Qu, Cen, Zeng, and Kan]{med_lora_llama}
Haojie Zhang, Yixiong Liang, Hulin Kuang, Lihui Cen, Zhe Qu, Yigang Cen, Min Zeng, and Shichao Kan.
\newblock Contrastive regularization over lora for multimodal biomedical image incremental learning.
\newblock \emph{arXiv preprint arXiv:2508.11673}, 2025{\natexlab{a}}.

\bibitem[Zhang et~al.(2023)Zhang, Xu, Usuyama, Xu, et~al.]{biomedclip}
Sheng Zhang, Yanbo Xu, Naoto Usuyama, Hanwen Xu, et~al.
\newblock Biomedclip: a multimodal biomedical foundation model pretrained from fifteen million scientific image-text pairs.
\newblock \emph{arXiv preprint arXiv:2303.00915}, 2023.

\bibitem[Zhang et~al.(2025{\natexlab{b}})Zhang, Li, Long, Zhang, Lin, Yang, Xie, Yang, Liu, Lin, et~al.]{qwen3}
Yanzhao Zhang, Mingxin Li, Dingkun Long, Xin Zhang, Huan Lin, Baosong Yang, Pengjun Xie, An~Yang, Dayiheng Liu, Junyang Lin, et~al.
\newblock Qwen3 embedding: Advancing text embedding and reranking through foundation models.
\newblock \emph{arXiv preprint arXiv:2506.05176}, 2025{\natexlab{b}}.

\bibitem[Zheng et~al.(2023)Zheng, Ma, Wang, Qin, Yue, and You]{zscl}
Zangwei Zheng, Mingyuan Ma, Kai Wang, Ziheng Qin, Xiangyu Yue, and Yang You.
\newblock Preventing zero-shot transfer degradation in continual learning of vision-language models.
\newblock In \emph{Proceedings of the IEEE/CVF international conference on computer vision}, pages 19125--19136, 2023.

\bibitem[Zhou et~al.(2022)Zhou, Yang, Loy, and Liu]{CoOp}
Kaiyang Zhou, Jingkang Yang, Chen~Change Loy, and Ziwei Liu.
\newblock Conditional prompt learning for vision-language models.
\newblock In \emph{Proceedings of the IEEE/CVF conference on computer vision and pattern recognition}, pages 16816--16825, 2022.

\bibitem[Zhu et~al.(2021)Zhu, Chen, Peng, Wang, and Jin]{chaoyang}
Chuang Zhu, Wenkai Chen, Ting Peng, Ying Wang, and Mulan Jin.
\newblock Hard sample aware noise robust learning for histopathology image classification.
\newblock \emph{IEEE transactions on medical imaging}, 41\penalty0 (4):\penalty0 881--894, 2021.

\bibitem[Zhu et~al.(2024)Zhu, Ma, Wang, Dong, Wang, Wu, Luo, Wang, and Li]{3distill_MIA_MRI}
Zhanshi Zhu, Xinghua Ma, Wei Wang, Suyu Dong, Kuanquan Wang, Lianming Wu, Gongning Luo, Guohua Wang, and Shuo Li.
\newblock Boosting knowledge diversity, accuracy, and stability via tri-enhanced distillation for domain continual medical image segmentation.
\newblock \emph{Medical image analysis}, 94:\penalty0 103112, 2024.

\end{thebibliography}

\clearpage

\beginappendix
\addtocontents{toc}{\protect\setcounter{tocdepth}{2}}

\tableofcontents

\vspace{2em}

\section{Limitations}
\noindent\textbf{Storage Usage.} As with all Retrieval-Augmented Generation (RAG) methodologies, the construction and maintenance of our multimodal retrieval corpus entail additional computational resources and human effort. Furthermore, dynamic retrieval inherently imposes computational and temporal overheads. These challenges motivate us to advocate for a phased implementation of our approach. Given that the initial content of the Question Pool is static, retrieval for this segment can be pre-computed and reused. Consequently, we restrict real-time retrieval operations solely to questions added subsequently.

\vspace{2pt}\noindent\textbf{The Trade-off in Data Sources.} To strictly prevent data leakage and avoid copyright or ethical disputes, we utilized the PubMed scientific literature database as our retrieval corpus. This choice ensures the reliability of our experimental results, the generalizability of the method in real-world deployments, and the overall safety and legality of the model. However, this imposes a limitation: application-level principles, such as fairness and broad ethical considerations, cannot be explicitly enforced at the algorithmic level during retrieval. Since the vast majority of scientific captions do not contain sensitive attributes like gender or race, and inferring such information solely from visual data is challenging, our continual learning method cannot explicitly target these dimensions. Instead, it primarily ensures the absence of fundamental safety hazards.

\section{Ethical Statement}
\label{sec:appendix_ethics}
This research strictly adheres to the relevant ethical guidelines for medical AI research. 

\vspace{2pt}\noindent\textbf{Data Usage and Patient Privacy.} All data used in this study were sourced from publicly available research publications or scientific datasets. All data were fully anonymized and de-identified by the original providers prior to release and contain no Protected Health Information (PHI). Our usage strictly complies with the Data Use Agreement (DUA) for PubMed (consistent with BiomedCLIP~\cite{biomedclip}/BIOMEDICA~\cite{biomedica}) and adheres to the DUAs of all respective open-source classification datasets involved.

\vspace{2pt}\noindent\textbf{Algorithmic Bias.} The performance of our model is dependent upon both the backbone and the continual learning methodology. The data for both components are sourced from scientific literature available on PubMed. Although it is generally assumed that a dataset comprising tens of millions of samples provides sufficient data diversity, it cannot be guaranteed that undiscovered biases (e.g., in demographic representation across race, age, or sex) are not present. These biases may subsequently be learned and amplified by the model. Future work is required to specifically quantify and mitigate such biases.

\section{Future Work}
We will focus on the real-world rollout of PRIMED and making it easier to deploy. 

\vspace{2pt}\noindent\textbf{Data Services.} Acknowledging the difficulties associated with large-scale data management, we will release our retrieval database subject to a secondary ethical audit. We will also establish a cloud service where users can retrieve information by merely uploading questions. Additionally, we support data streaming to streamline local deployment and the augmentation of proprietary retrieval repositories.

\vspace{2pt}\noindent\textbf{Rare Disease Content Enrichment.} To mitigate the data scarcity and heterogeneity of rare diseases, we plan to augment both the retrieval corpus and the Question Pool, thereby extending our framework's generalizability and value in this challenging domain.

\section{Method and Evaluation Details} 
This section details the data cleaning methodologies employed to complement the construction of the retrieval corpus. It further elucidates the logic governing the curation of the benchmark dataset and offers comprehensive supplementary information. Lastly, we provide the disaggregated performance metrics for HieraMedTransfer, noting that the aggregated means of these values constitute the results presented in the main body of the paper.

\subsection{Data Cleansing Approaches}

\noindent\textbf{Data Acquisition.} We adopted the data acquisition methodology outlined in BIOMEDICA~\cite{biomedica} to collect raw image-caption pairs. Since BIOMEDICA has already implemented fine-grained unsupervised clustering based on DINOv2~\cite{dinov2}, we were able to exclude clinically irrelevant content—such as charts and natural images—by simply filtering based on the off-the-shelf pseudo-labels.

\vspace{2pt}\noindent\textbf{Multi-Subgraph Decoupling.} Scientific literature frequently employs multi-panel figures to serve its illustrative purposes. However, given that clinical diagnostic images are predominantly presented in a single-panel format, the abundance of multi-panel content in reference datasets is detrimental to Continual Learning.

\begin{wrapfigure}{r}{0.5\textwidth}
  \vspace{-20pt} 
  \begin{center}
    \includegraphics[width=1\linewidth]{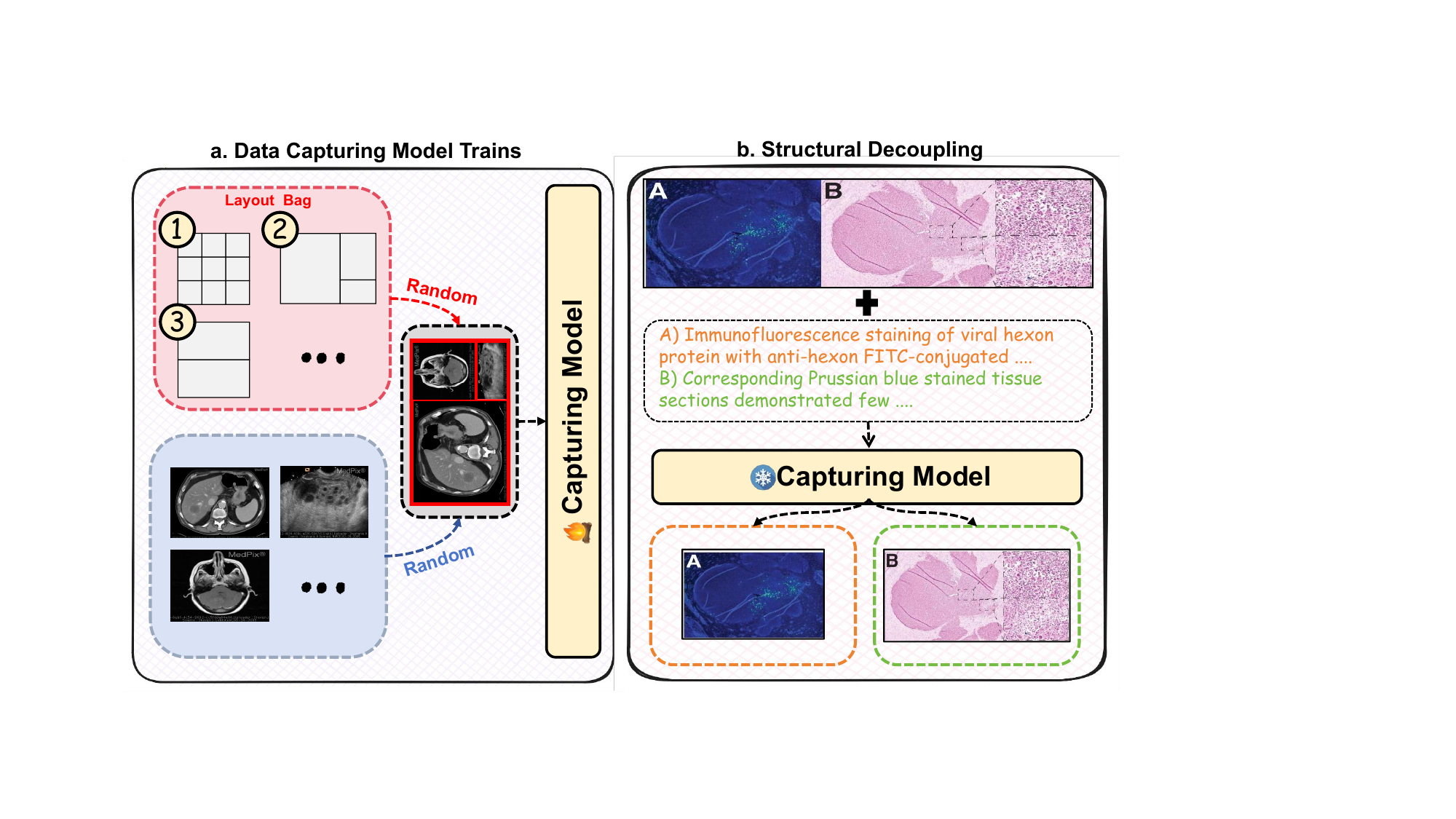}
  \end{center}
  \caption{Overview of the Multi-Subgraph Capture Model: Training Methodology and Application Scenarios.}
  \label{sup_fig1}
  \vspace{-10pt} 
\end{wrapfigure}

Drawing inspiration from the work~\cite{Open-PMC-18M} of Baghbanzadeh et al., we recognize that utilizing object detection models for multi-subgraph partitioning presents a promising approach. Given that subfigures in scientific literature typically consist of content with similar domains or semantics, we utilized single-panel images from our previously collected dataset for synthesis. Specifically, we combined images sharing the same coarse-grained pseudo-labels to generate synthetic multi-panel figures, resulting in a batched training dataset with object detection annotations. As depicted in Fig.\ref{sup_fig1}, we present a Multi-Subgraph Capture Model tailored for the medical domain, leveraging the DAB-DETR~\cite{Dab-detr} architecture.

Ultimately, regular expressions were utilized to identify subfigure indicators (e.g., (1), (a), A). This facilitated the segmentation and realignment of captions based on spatial layout, given that scientific literature typically follows a fixed left-to-right reading sequence.

\begin{table*}[t]
\centering
\small
\caption{Visualization of the nine datasets utilized in HieraMedTransfer. We implemented a design for multi-scale transfer across both in-domain and out-of-domain settings.}
\label{sub_dataset_demo_1}
\begin{adjustbox}{max width=\textwidth}
\resizebox{1\linewidth}{!}{
\begin{tabular}{
    >{\centering\arraybackslash}m{9cm} 
    >{\centering\arraybackslash}m{3cm} 
    >{\centering\arraybackslash}m{2cm} 
    >{\centering\arraybackslash}m{2cm} 
    >{\centering\arraybackslash}m{2cm} 
    >{\centering\arraybackslash}m{1cm}
}
\toprule
\rowcolor{gray!10} \textbf{Dataset Example} & \textbf{Dataset Name} & \textbf{Domain} & \textbf{Region/Type}&\textbf{Number}&\textbf{Classes} \\

\includegraphics[width=8cm, height=1cm]{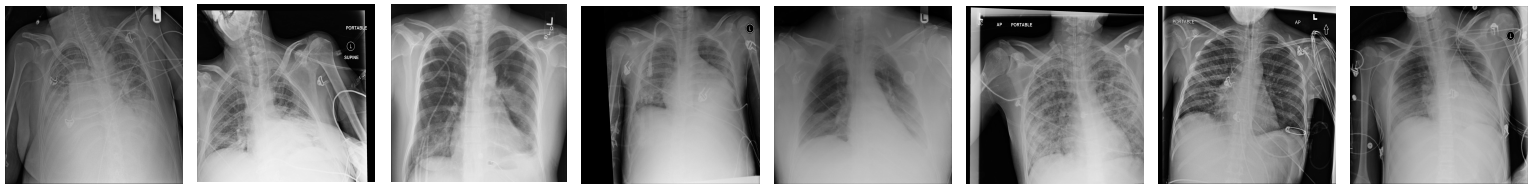} & RANZCR~\cite{ranzcr} & X-ray & Blood Vessel& 33665 & 11\\

\includegraphics[width=8cm, height=1cm]{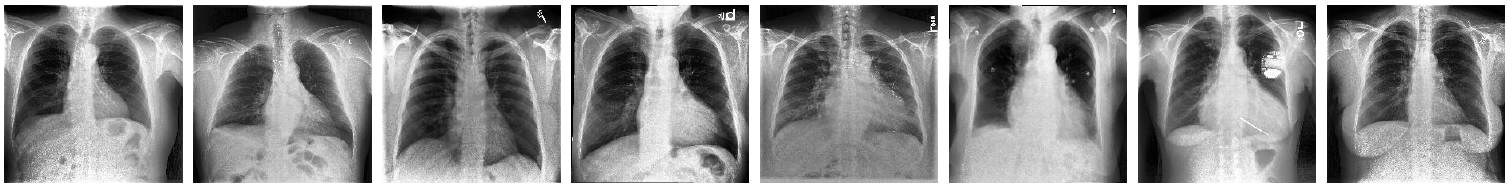} & CheXchoNet~\cite{CheXchoNet} & X-ray & Chest & 71589 & 4\\

\includegraphics[width=8cm, height=1cm]{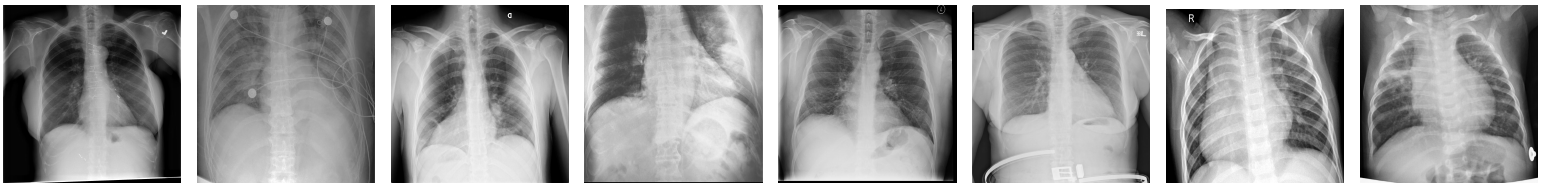} & PD~\cite{PD} & X-ray & Lung& 4575 & 3\\

\includegraphics[width=8cm, height=1cm]{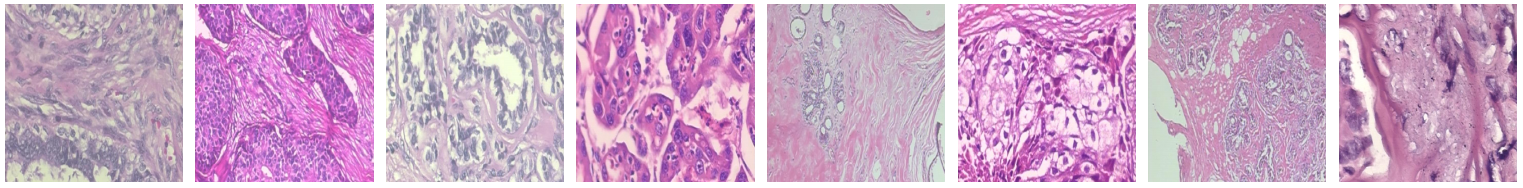} & Breakhis~\cite{breakhis} & Patho. & Breast& 7909 & 2\\

\includegraphics[width=8cm, height=1cm]{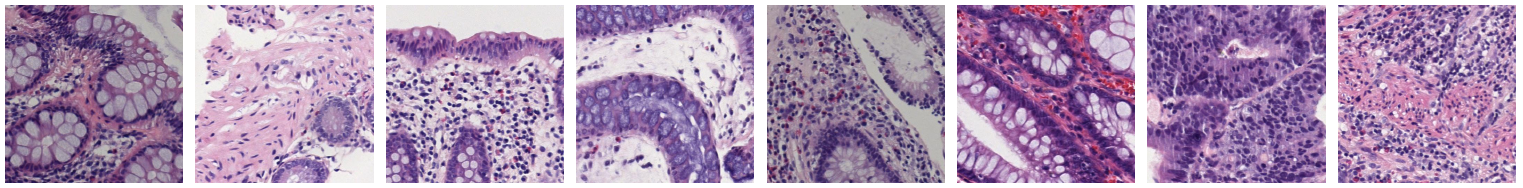} & Chaoyang~\cite{chaoyang} & Patho. & Colonic& 6160 & 4\\

\includegraphics[width=8cm, height=1cm]{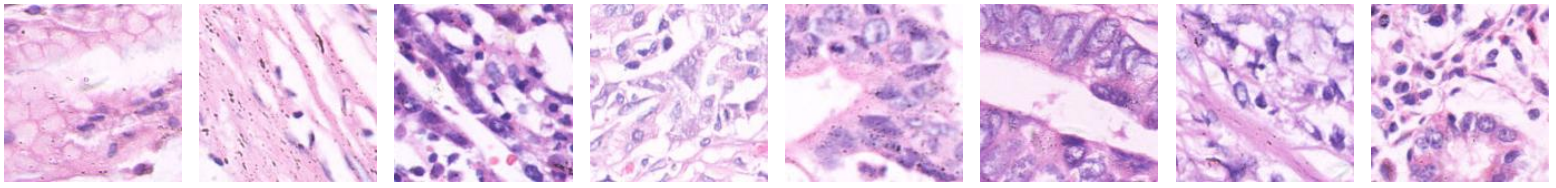} & Nucls~\cite{Nucls} & Patho. & Gastric& 33284 & 2\\

\includegraphics[width=8cm, height=1cm]{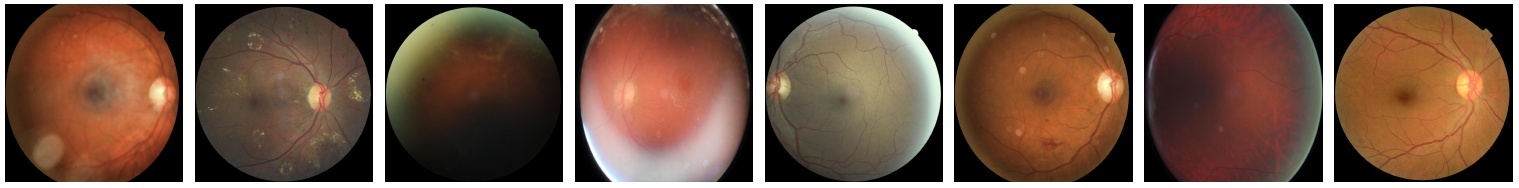} & Eyepacs~\cite{eyepacs} & Fundus & Diabetes& 35126 & 5\\

\includegraphics[width=8cm, height=1cm]{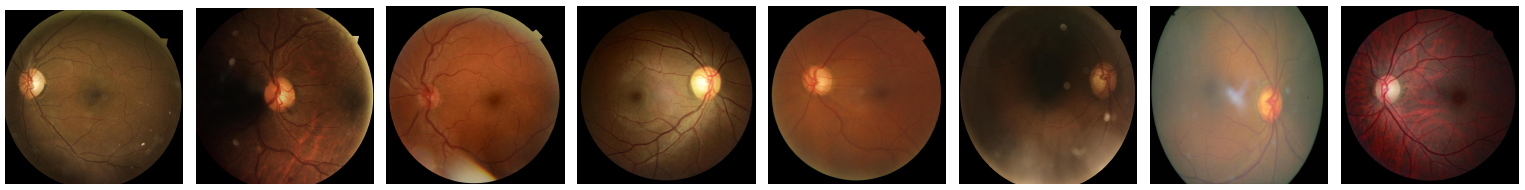} & AIROGS~\cite{AIROGS} & Fundus & Glaucoma& 101442 & 2\\

\includegraphics[width=8cm, height=1cm]{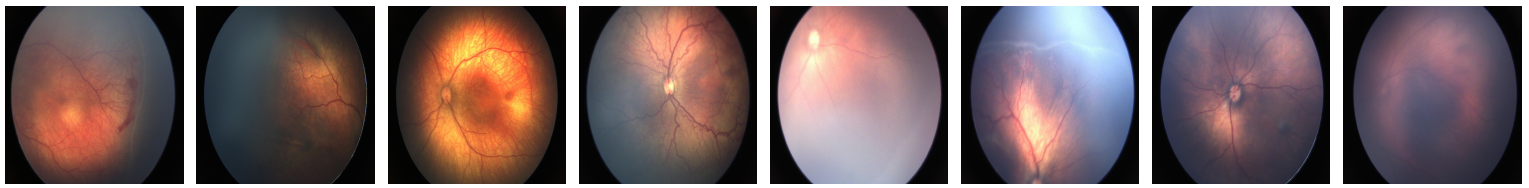} & FARFUM-RoP~\cite{FARFUM-RoP} & Fundus& ROP& 1533 & 3\\

\bottomrule
\end{tabular}
}
\end{adjustbox}
\end{table*}

\begin{table*}[t]
\centering
\small
\caption{Visualization of the six datasets utilized in MedXtreme. This collection was curated to maximize classification difficulty while satisfying the data requirements for fine-tuning.}
\label{sub_dataset_demo_2}
\begin{adjustbox}{max width=\textwidth}
\resizebox{1\linewidth}{!}{
\begin{tabular}{
    >{\centering\arraybackslash}m{9cm} 
    >{\centering\arraybackslash}m{3cm} 
    >{\centering\arraybackslash}m{2cm} 
    >{\centering\arraybackslash}m{2cm} 
    >{\centering\arraybackslash}m{2cm} 
    >{\centering\arraybackslash}m{1cm}
}
\toprule
\rowcolor{gray!10} \textbf{Dataset Example} & \textbf{Dataset Name} & \textbf{Domain} & \textbf{Region/Type}&\textbf{Number}&\textbf{Classes} \\

\includegraphics[width=8cm, height=1cm]{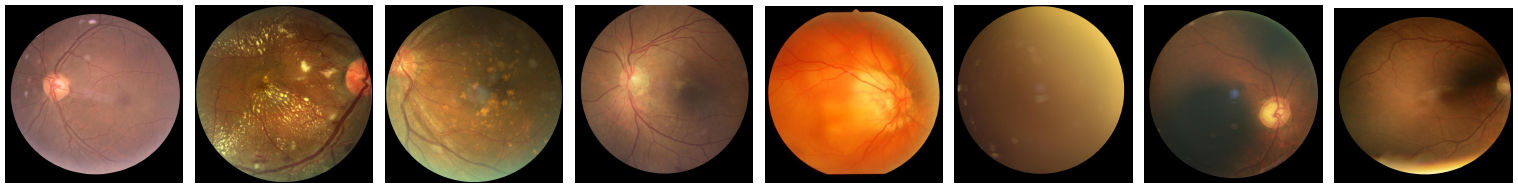} & AOD~\cite{AOD} & Fundus & Eye& 10000 & 8 \\

\includegraphics[width=8cm, height=1cm]{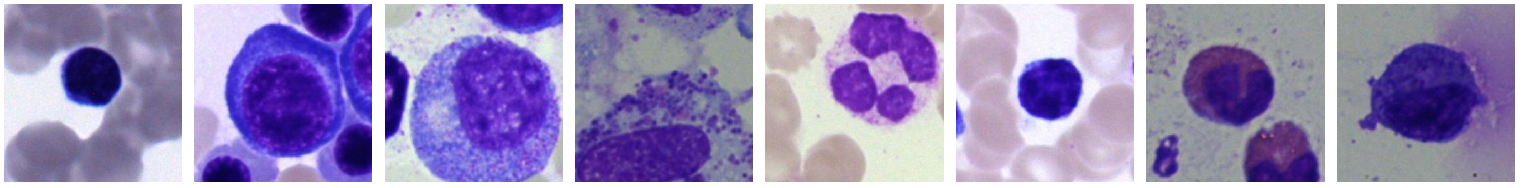} & BMC~\cite{BMC} & Cell & Bone Marrow& 171375 & 21 \\

\includegraphics[width=8cm, height=1cm]{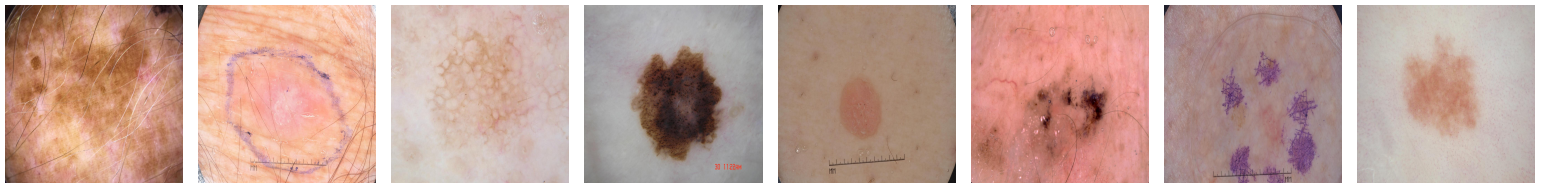} & ISIC2024~\cite{isic-2024} & Skin & Skin& 81722 & 33 \\

\includegraphics[width=8cm, height=1cm]{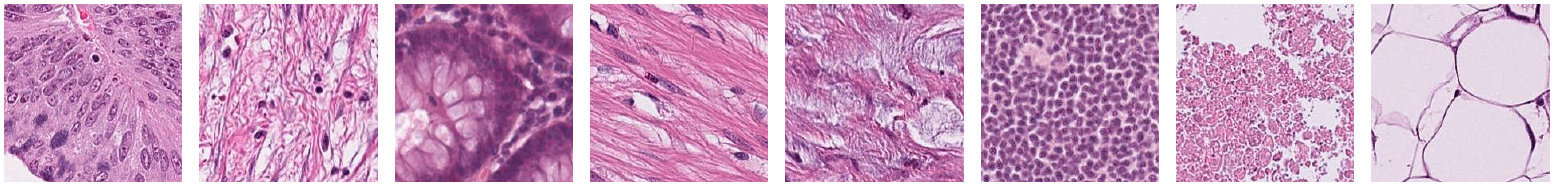} & NCT100K~\cite{NCT} & Patho. & Colorectal& 100000 & 9 \\

\includegraphics[width=8cm, height=1cm]{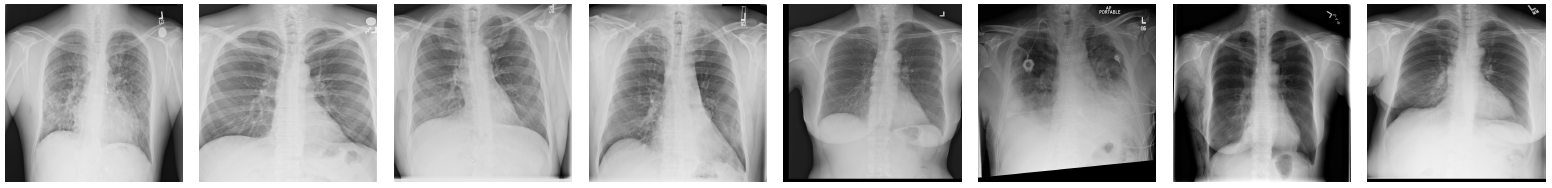} & NIH-Chest-Xray~\cite{NIH-Chest-ray} & X-ray & Chest& 112120 & 15 \\

\includegraphics[width=8cm, height=1cm]{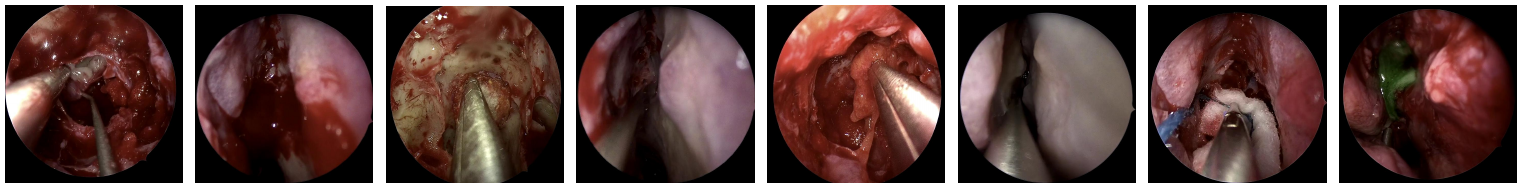} & PITVIS~\cite{Pitvis} & Endo. & Pituitary& 120024 & 15\\

\bottomrule
\end{tabular}
}
\end{adjustbox}

\end{table*}

\begin{figure*}[t]
\centering
 \includegraphics[width=1\linewidth]{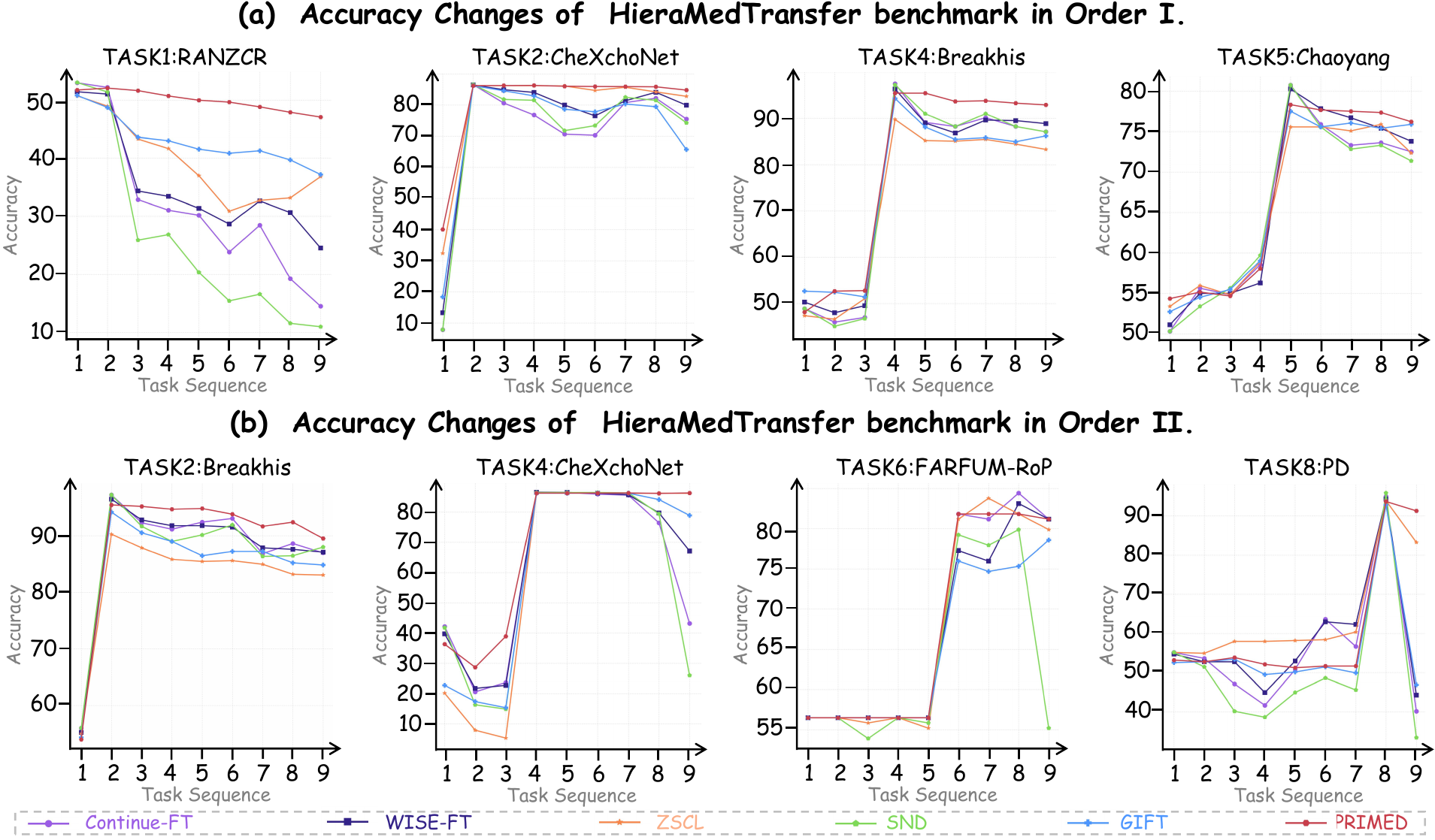}
\caption{Illustration of classification accuracy changes as tasks are learned on the HieraMedTransfer benchmark in two orders. Our method consistently exhibits a mirrored Z-shaped pattern.}
\label{sup_fig2}
\end{figure*}

\subsection{Benchamrk Database Construction}

We constructed the MGTIL benchmark based on three key principles:
\begingroup
\setlist[itemize]{noitemsep, topsep=0pt, left=0pt}
\begin{itemize}
    \item There are substantial discrepancies across domains, which vary significantly in terms of imaging modalities and spatial resolutions.
    \item The dataset preserves fine-grained intra-domain variations, such as diverse spatial regions and varying object scales, thereby avoiding severe homogeneity.
    \item These challenging datasets typify the common hurdles in medical classification tasks, being characterized by a large number of categories, high task complexity, and the prevalence of few-shot scenarios.  
\end{itemize}
\endgroup

\vspace{2pt}\noindent\textbf{HieraMedTransfer  Construction.} As detailed in Tab. \ref{sub_dataset_demo_1}, our experimental evaluation encompasses three distinct modalities: X-ray, pathological, and fundus images. The intra-domain datasets exhibit multi-level discrepancies, including variations in anatomical regions, resolutions, and label granularity, which effectively mirrors the complex data heterogeneity inherent to real-world medical scenario.

\vspace{2pt}\noindent\textbf{MedXtreme  Construction.} MedXtreme encompasses six medical datasets across distinct domains, characterized by large label spaces, high task complexity, and significant domain shifts. It is designed to evaluate a model's capacity for learning and memory retention on challenging tasks within a continual fine-tuning setting. Notably, the inclusion of a substantial number of few-shot classes effectively simulates the dilemma of diagnosing rare diseases in clinical practice. Further details on the dataset are provided in Tab.\ref{sub_dataset_demo_2}.

\vspace{2pt}\noindent\textbf{Detailed Performance Analysis.} Tab.\ref{tab:Sup_HieraMedTransfer_order_I} and Tab.\ref{tab:Sup_HieraMedTransfer_order_II} detail the results on HieraMedTransfer for Order I and II. Task-specific metrics follow the ZSCL~\cite{zscl} protocol, with averages listed in the "Average" column as in the main text. Notably, we supplement these tables with "Fine-tune" results—representing the theoretical upper bound achieved by fine-tuning solely on the target dataset. As shown, our method yields the highest average performance among state-of-the-art competitors and, most notably, performs comparably to the Fine-tune upper bound.

We visualize the accuracy evolution of selected tasks from two distinct orders in Fig.\ref{sup_fig2} (a) and (b). In the context of Vision-Language Models (VLMs), an optimal Continual Learning strategy is characterized by a \textbf{``mirrored Z-shaped''} curve. This signifies the effective maintenance of zero-shot capabilities prior to task acquisition and the mitigation of forgetting subsequent to learning. As demonstrated, our method aligns closely with this ideal profile.

\clearpage

\begin{table*}[t!]
  \centering
  \caption{Detailed Transfer, Avg., and Last scores (\%) of different continue training methods on HieraMedTransfer benchmark in \textbf{Order I}. \colorbox{RedOrange!8}{\textbf{Red Background}} \& \textbf{bold} indicate best results.}
  \label{tab:Sup_HieraMedTransfer_order_I}
  {
  \begin{tabular}{@{}ccccccccccccc@{}}
  \toprule
  Method & \rotatebox{90}{RANZCR \cite{ranzcr}}& \rotatebox{90}{CheXchoNet \cite{CheXchoNet}} & \rotatebox{90}{PD \cite{PD}} & \rotatebox{90}{Breakhis \cite{breakhis}} & \rotatebox{90}{Chaoyang \cite{chaoyang}} & \rotatebox{90}{NuCLS \cite{Nucls}} & \rotatebox{90}{Eyepacs \cite{eyepacs}} & \rotatebox{90}{AIROGS  \cite{AIROGS}} & \rotatebox{90}{FARFUM-RoP  \cite{FARFUM-RoP}}  & Average \\ \midrule
  
   Zero-shot & 16.44 & 28.34 & 52.07 & 53.88 & 52.65 & 57.45 & 62.28 & 96.46 & 56.49 &52.90  \\
   Fine-tune & 53.13 & 86.55 & 95.64 & 97.35 & 80.91 & 97.03 & 78.33 & 97.55 & 79.87 &85.15  \\ \midrule
   \textbf{Transfer} \\
    Continual FT  & & 7.85 & 39.22 & 47.33 & 54.85 & 56.50 & 57.65 & 91.52 & 56.49 & 51.43 \\
    $l_2$ baseline & &  13.09 & 38.24 & 51.41 & 54.49 & 55.68 & 64.34 & 95.40 & 56.48 & 53.64 \\
    LwF~\cite{lwf} & & 31.89 & 43.68 & \textbf{52.51} & 54.13 & 56.49 & 19.96 & 34.30 & 56.49 & 43.68\\
    iCaRL~\cite{icarl} & &7.70 & 32.89 & 54.49 & 55.57 & \textbf{60.60} & 58.26 & 96.67 & 56.41 & 52.82 \\
    WiSE-FT~\cite{wise_ft} & &  13.33 & 40.85 & 49.31 & 54.37 & 55.85 & 60.66 & 94.51 & 56.49 & 53.17 \\
    ZSCL~\cite{zscl} & &34.88 & 47.17 & 51.08 & 54.41 & 57.26 & \textbf{64.43} & 95.68 & 56.49 & 57.68 \\
    MoE-CL~\cite{MOE_CL} & & 32.61 & 51.36 & 50.14 & 53.96 & 56.40 & 61.72 & 92.87 & 55.38 & 56.81\\  
    SND~\cite{snd} & & 7.85 & 40.20 & 46.95 & 54.77 & 58.21 & 58.65 & 92.98 & 56.41 & 52.00\\  
    DIKI~\cite{DIKI}& & 21.87 & 47.51 & 51.03 & 54.96 & 58.36 & 63.82 & 96.65 & 56.48 & 56.34\\  
    GIFT~\cite{GIFT}& & 15.42 & 42.57 & 52.21 & 55.46 & 52.27 & 59.25 & 93.77 & 56.49 & 53.43 \\ 
    $\text{PRIMED}_{uni}$ & &27.46 & \textbf{52.40} & 50.91 & 54.98 & 55.77 & 62.73 & 96.21 & 56.49 & 57.12\\  
    \rowcolor{RedOrange!8}
     $\text{PRIMED}_{dyn}$ & & \textbf{39.09} & 48.38 & 51.24 & \textbf{55.58} & 55.42 & 63.68 & \textbf{96.72} & \textbf{56.49} & \textbf{58.33} \\  
    \midrule
   \textbf{Avg.} \\
    Continual FT   & 31.69 & 70.13 & 80.90 & 75.81 & 66.16 & 72.41 & 59.83 & 91.84 & 59.38 & 67.57 \\
    $l_2$ baseline & 27.71 & 76.31 & 77.25 & 76.05 & 66.25 & 71.94 & 66.82 & 95.35 & 58.94 & 68.51 \\
    LwF~\cite{lwf} &8.55 & 67.76 & 68.70 & 68.84 & 60.52 & 70.07 & 37.36 & 47.07 & 59.45 & 54.26 \\
    iCaRL~\cite{icarl} &49.69 & 77.56 & 71.53 & 76.41 & 64.92 & \textbf{76.00} & 64.42 & \textbf{96.78} & 59.09 & 70.71\\
    WiSE-FT~\cite{wise_ft} &35.33 & 74.52 & 81.38 & 76.46 & 66.83 & 71.93 & 63.60 & 94.79 & 59.45 & 69.37\\
    ZSCL~\cite{zscl} &43.91 & 78.98 & 80.39 & 72.74 & 64.74 & 70.76 & 67.67 & 95.90 & 59.09 & 70.46 \\
    MoE-CL~\cite{MOE_CL} &33.07 & 80.58 & \textbf{84.92} & 77.92 & 65.58 & 70.87 & 67.40 & 96.84 & 59.43&70.73 \\  
    SND~\cite{snd} &  25.77 & 71.23 & 82.21 & 75.96 & 65.88 & 73.43 & 60.05 & 93.50 & 59.02 & 67.45 \\  
    DIKI~\cite{DIKI}&41.84 & 79.50 & 80.74 & 75.70 & 64.81 & 72.58 & 67.61 & 96.28 & 59.41&70.94\\  
    GIFT~\cite{GIFT}&42.97 & 72.68 & 80.37 & 75.73 & 66.92 & 71.39 & 63.57 & 95.95 & 58.29 & 69.76 \\ 
    $\text{PRIMED}_{uni}$ & 50.49 & 79.34 & 83.95 & 78.41 & 66.52 & 72.98 & 66.63 & 96.25 &\textbf{59.45} & 72.67 \\  
    \rowcolor{RedOrange!8}
     $\text{PRIMED}_{dyn}$ &\textbf{50.95} & \textbf{80.83} & 83.35 & \textbf{79.82} & \textbf{67.71} & 72.71 & \textbf{67.73} & 95.61 & 59.30 & \textbf{73.11} \\  
    \midrule
   \textbf{Last} \\
    Continual FT  &14.49 & 75.50 & 90.41 & 87.10 & 72.49 & 85.46 & 40.84 & 88.32 & 82.47 & 70.79 \\
    $l_2$ baseline &15.88 & 81.04 & 89.11 & 87.48 & 72.98 & 85.40 & 63.86 & 93.57 & 77.92 & 74.14 \\
    LwF~\cite{lwf} &1.92 & 75.67 & 70.59 & 72.06 & 64.40 & 67.56 & 64.31 & 86.23 & 83.12 & 65.10 \\
    iCaRL~\cite{icarl} &48.66 & \textbf{86.23} & 79.21 & 73.70 & 65.05 & \textbf{94.62} & \textbf{75.93} & 96.12 & 80.52 & 77.78 \\
    WiSE-FT~\cite{wise_ft} &24.51 & 79.94 & 90.63 & 88.87 & 73.79 & 86.42 & 55.48 & 94.28 & 83.12 & 75.23 \\
    ZSCL~\cite{zscl} & 41.75 & 81.94 & 87.58 & 80.91 & 71.84 & 84.41 & 74.08 & 96.08 & 79.87 & 77.61 \\
    MoE-CL~\cite{MOE_CL} &38.80 & 79.57 & 85.71 & 82.83 & 71.94 & 87.27 & 67.61 & 94.27 & 77.25 &76.14\\  
    SND~\cite{snd} & 11.95 & 74.26 & 91.94 & 87.10 & 71.36 & 86.51 & 35.29 & 92.99 & 79.87 & 70.14\\  
    DIKI~\cite{DIKI}& 41.29& 84.16 & 91.60 & 86.11 & 72.10 & 84.97 & 70.64 & 91.56 & 74.47 &77.43\\  
    GIFT~\cite{GIFT}&37.18 & 65.71 & 84.75 & 86.22 & 75.89 & 92.67 & 64.82 & 96.34 & 72.73 & 75.15\\ 
    $\text{PRIMED}_{uni}$ & \textbf{49.78} & 84.77 & 91.94 & 91.66 & 74.60 & 93.30 & 70.64 & 95.96 & \textbf{83.12} & 81.75 \\  
    \rowcolor{RedOrange!8}
     $\text{PRIMED}_{dyn}$ &47.07 & 84.80 & \textbf{92.37} & \textbf{92.92} & \textbf{76.21} & 92.31 & 74.74 & \textbf{96.40} & 81.82 & \textbf{82.07} \\  
  \bottomrule
  \end{tabular}%
  }
  \end{table*}

\clearpage

\begin{table*}[t!]
  \centering
  \caption{Detailed Transfer, Avg., and Last scores (\%) of different continue training methods on HieraMedTransfer benchmark in \textbf{Order II}. \colorbox{RedOrange!8}{\textbf{Red Background}} \& \textbf{bold} indicate best results.}
  \label{tab:Sup_HieraMedTransfer_order_II}
  {
  \begin{tabular}{@{}ccccccccccccc@{}}
  \toprule
Method & \rotatebox{90}{AIROGS \cite{AIROGS}} & \rotatebox{90}{Breakhis \cite{breakhis}} & \rotatebox{90}{Chaoyang \cite{chaoyang}} & \rotatebox{90}{CheXchoNet \cite{CheXchoNet}} & \rotatebox{90}{Eyepacs \cite{eyepacs}} & \rotatebox{90}{FARFUM-RoP \cite{FARFUM-RoP}} & \rotatebox{90}{NuCLS \cite{Nucls}} & \rotatebox{90}{PD \cite{PD}} & \rotatebox{90}{RANZCR \cite{ranzcr}} & Average \\ \midrule
  
   Zero-shot & 96.46 & 53.88 & 52.65 & 28.34 & 62.28 & 56.49 & 57.45 & 52.07 & 16.44 &52.90 \\
   Fine-tune & 97.55 & 97.35 & 80.91 & 86.55 & 78.33 & 79.87 & 97.03 & 95.64 & 53.13 & 85.15  \\ \midrule
   \textbf{Transfer} \\
    Continual FT  & &56.01 & 56.15 & 28.83 & 64.18 & 56.49 & 55.10 & 52.63 & 11.98 & 47.67 \\
    $l_2$ baseline & &55.12 & 55.82 & 18.58 & 64.80 & 56.62 & 54.15 & 53.52 & 7.40 & 45.75 \\
    LwF~\cite{lwf} & & 57.65 & 55.82 & 19.78 & 61.65 & 55.19 & 57.56 & 52.94 & 14.42 & 46.88 \\
    iCaRL~\cite{icarl} & & 53.73 & 54.69 & 38.39 &\textbf{ 68.51} & 56.49 & \textbf{57.92} & 48.80 & 6.57 & 48.14 \\
    WiSE-FT~\cite{wise_ft} & &55.25 & 55.74 & 28.09 & 67.45 & 56.49 & 54.44 & 54.74 & 14.11 & 48.29 \\
    ZSCL~\cite{zscl} & &55.25 & 55.10 & 9.42 & 67.86 & 51.43 & 54.31 & 56.83 & 11.36 & 45.20 \\
    MoE-CL~\cite{MOE_CL} & &52.87 & 55.50 & 28.78 & 63.12 & 55.93 & 53.29 & 53.09 & 18.25 &47.60\\  
    SND~\cite{snd} & & \textbf{56.01} & 56.40 & 24.34 & 64.42 & 55.84 & 56.04 & 46.31 & 15.96 & 46.92 \\  
    DIKI~\cite{DIKI}& & 52.06 & 55.02 & 20.54 & 64.76 & 55.63 & 54.77 & 49.19 & 20.57 &46.57\\  
    GIFT~\cite{GIFT}& &  54.36 & 54.86 & 18.51 & 67.72 & 56.49 & 55.06 & 51.36 & 15.95 & 46.79 \\ 
    $\text{PRIMED}_{uni}$ & & 53.98 & 54.86 & 18.90 & 66.23 & 56.49 & 53.18 & \textbf{56.92} & \textbf{22.62} & 47.90\\  
    \rowcolor{RedOrange!8}
     $\text{PRIMED}_{dyn}$ & & 53.78 & \textbf{56.40} & \textbf{34.66} & 64.28 & \textbf{56.89} & 52.09 & 53.62 & 16.44 & \textbf{48.52}\\  
    \midrule
   \textbf{Avg.} \\
    Continual FT  &79.46 & 87.18 & 71.84 & 61.17 & 48.53 & 67.59 & 68.67 & 55.94 & 16.55 & 61.88\\
    $l_2$ baseline & 94.25 & 84.67 & 69.83 & 62.98 & 62.75 & 64.79 & 67.26 & 56.77 & 12.18 & 63.94 \\
    LwF~\cite{lwf} &89.64 & 72.64 & 63.22 & 53.50 & 63.82 & 65.87 & 70.31 & 59.04 & 17.45 & 61.72 \\
    iCaRL~\cite{icarl} &96.93 & 87.81 & 68.16 & \textbf{70.34} & \textbf{73.58} & 61.25 & \textbf{70.69} & 54.01 & 11.20 & 66.00 \\
    WiSE-FT~\cite{wise_ft} &92.30 & 86.91 & 72.14 & 64.70 & 57.51 & 66.66 & 66.65 & 57.98 & 18.37 & 64.80 \\
    ZSCL~\cite{zscl} & 96.70 & 82.92 & 69.04 & 60.61 & 71.55 & 56.85 & 66.86 & \textbf{64.66} & 15.92 & 65.01 \\
    MoE-CL~\cite{MOE_CL} &95.88 & 88.51 & 69.98 & 63.57 & 68.41 & 64.76 & 65.11 & 59.62 & 18.73&66.06 \\  
    SND~\cite{snd} & 81.36 & 86.31 & 71.20 & 58.23 & 48.53 & 63.49 & 69.46 & 50.37 & 20.05 & 61.00 \\  
    DIKI~\cite{DIKI}&96.95 & 86.80 & 69.94 & 61.02 & 69.17 & 62.86 & 65.44 & 62.69 & 16.39 &65.70\\  
    GIFT~\cite{GIFT}&95.44 & 84.34 & 72.19 & 62.61 & 61.64 & 65.22 & 68.37 & 55.46 & 19.85 & 65.01 \\ 
    $\text{PRIMED}_{uni}$ &96.51 & 88.12 & 72.08 & 62.21 & 69.54 & 66.74 & 67.33 & 60.78 & 25.81 & 67.68\\  
    \rowcolor{RedOrange!8}
     $\text{PRIMED}_{dyn}$ & \textbf{97.00} &\textbf{89.06} & \textbf{72.19} & 69.06 & 69.34 & \textbf{67.68} & 66.46 & 61.24 & \textbf{20.32} & \textbf{68.04} \\  
    \midrule
   \textbf{Last} \\
    Continual FT  & 39.17 & 86.98 & 71.20 & 43.22 & 17.29 & 81.17 & 95.46 & 40.09 & \textbf{53.09} & 58.63\\
    $l_2$ baseline &78.62 & 86.47 & 66.02 & 81.80 & 38.91 & 66.23 & 91.59 & 41.18 & 50.45 & 66.81\\
    LwF~\cite{lwf} &59.47 & 67.89 & 63.92 & 40.52 & 48.05 & 74.68 & 95.46 & 59.69 & 45.65 & 61.70 \\
    iCaRL~\cite{icarl} & 96.79 & 84.07 & 64.72 & 86.23 & \textbf{77.33} & 62.99 & 94.82 & 49.67 & 48.26 & 73.88 \\
    WiSE-FT~\cite{wise_ft} & 76.65 & 87.10 & 74.76 & 67.10 & 23.61 & 81.17 & 86.00 & 44.23 & 52.46 & 65.90\\
    ZSCL~\cite{zscl} & 96.67 & 84.96 & 70.71 & 86.13 & 74.51 & 45.45 & 91.17 & 89.98 & 52.33 & 76.88 \\
    MoE-CL~\cite{MOE_CL} &81.34 & 85.95 & 70.27 & 76.56 & 64.75 & 71.58 & 91.47 & 75.99 & 50.02 &74.21\\  
    SND~\cite{snd} &47.02 & 87.99 & 72.33 & 26.00 & 9.71 & 55.19 & 95.55 & 33.33 & 52.76 & 53.32 \\  
    DIKI~\cite{DIKI}&90.07 & 85.97 & 73.11 & 78.35 & 68.66 & 75.32 & 90.84 & 80.31 & 48.03 &76.74\\  
    GIFT~\cite{GIFT}& 90.59 & 84.83 & 75.08 & 78.87 & 41.24 & 78.57 & 94.83 & 46.84 & 51.04 & 71.32 \\ 
    $\text{PRIMED}_{uni}$ & \textbf{96.90} & 88.37 & 75.24 & 78.33 & 70.66 & 78.57 & \textbf{96.40} & 57.12 & 51.31 & 76.99 \\  
    \rowcolor{RedOrange!8}
     $\text{PRIMED}_{dyn}$ &91.93 & \textbf{89.51} & \textbf{75.73} & \textbf{86.31} & 68.61 & \textbf{81.17} & 95.01 &\textbf{ 91.29} & 51.31 & \textbf{81.21} \\  
  \bottomrule
  \end{tabular}%
  }
  \end{table*}

\clearpage

\begin{table*}[t!] 
\renewcommand{\arraystretch}{1} 
\small 
\centering
\caption{Ablation study of different modules on HieraMedTransfer and MedXtreme. \colorbox{RedOrange!8}{\textbf{Red Background}} indicates the full model.}
\vspace{-1.5ex}

\begin{tabular}{@{}>{\centering\arraybackslash}m{0.85cm}
                  >{\centering\arraybackslash}m{0.85cm}
                  >{\centering\arraybackslash}m{0.85cm}|
                  c c c !{\vrule width 1.15pt} 
                  c c c !{\vrule width 1.15pt} 
                  c c c !{\vrule width 1.15pt} 
                  c c c @{}}
\toprule

\multirow{2}{*}{\small{\textbf{+CKT}}} & 
\multirow{2}{*}{\small{\textbf{+CMC}}} & 
\multirow{2}{*}{\small{\textbf{+DFG}}} & 

\multicolumn{3}{c!{\vrule width 1.2pt}}{\textbf{HieraMedTransfer I}} & 
\multicolumn{3}{c!{\vrule width 1.2pt}}{\textbf{HieraMedTransfer II}} &
\multicolumn{3}{c!{\vrule width 1.2pt}}{\textbf{MedXtreme I}} & 
\multicolumn{3}{c}{\textbf{MedXtreme II}} \\

\cmidrule(lr){4-6} \cmidrule(lr){7-9} \cmidrule(lr){10-12} \cmidrule(lr){13-15}

& & & Transfer & Avg. & Last 
& Transfer & Avg. & Last 
& ACC & AUC & BWT 
& ACC & AUC & BWT  \\
\midrule

$\checkmark$ & & 
&  56.2 & 71.8 & \textbf{82.6}
&  46.7 & 67.8 & 81.0
&  66.7 & 87.1 & -4.2
&  65.9 & 86.3 & -5.3 \\
& $\checkmark$ & 
&  54.3 & 68.9 & 78.9
&  49.5 & 67.2 & 76.4
&  64.9 & 85.4 & -7.6
&  60.4 & 83.7 & -13.2 \\
$\checkmark$ & $\checkmark$ & 
&  56.9 & 71.6 & 82.2
&  48.2 & 67.9 & 81.0
&  66.5 & 85.9 & -4.8
&  64.9 & 85.7 & -6.8\\
$\checkmark$ & & $\checkmark$ 
&  57.2 & 72.8 & 82.5
&  46.8 & 67.8 & 80.9
&  66.7 & 87.3 & -4.3
&  65.8 & 86.1 & -5.4\\
& $\checkmark$ & $\checkmark$ 
&  56.7 & 70.2 & 77.1
&  \textbf{50.0} & 67.4 & 77.2
&  65.2 & 85.2 & -7.1
&  60.4 &  82.8 & - 13.3 \\

\rowcolor{RedOrange!8}
$\checkmark$ & $\checkmark$ & $\checkmark$ 
&  \textbf{58.3} & \textbf{73.1} & 82.1
&  48.5 & \textbf{68.0} &\textbf{81.2}
&  \textbf{68.6} & \textbf{87.4} & \textbf{-2.7}
&  \textbf{68.1} & \textbf{86.3} & \textbf{-3.4} \\

\bottomrule
\end{tabular}%
\vspace{-1ex} 
\label{tab:module} 
\end{table*}

\section{Ablation Study}
To validate the effectiveness of our experimental settings at all levels, we performed extensive ablation studies involving four sequences on our two proposed benchmarks. The analysis is organized as follows: experimental setup, module-level ablation, hyperparameter and component-level ablation, and Dynamic Retrieval analysis.

\subsection{Full Settings}
To ensure reproducibility, we detail the key configurations and experimental settings for training our model as follows:
\begingroup
\setlist[itemize]{noitemsep, topsep=0pt, left=0pt}
\begin{itemize}
    \item \textbf{Batch Size and Label Smoothing:} We employ a batch size of 64 per GPU and apply label smoothing of 0.2. Notably, fine-tuning is fixed at 1,000 iterations across all datasets; for datasets with insufficient samples, the training data is cycled to meet this requirement.
    \item \textbf{Learning Rate:} A unified learning rate of $1\times10^{-5}$ is applied across all regularization, replay, and distillation methods~\cite{zscl,icarl,GIFT,wise_ft,lwf,snd}. For approaches based on LoRA~\cite{MOE_CL} or Prompt Tuning~\cite{DIKI}, we strictly adhere to the hyperparameter settings outlined in their papers.
    \item \textbf{Detailed Configurations:} In the following sections, we present a comprehensive ablation study covering all relevant components and hyperparameters. For clarity, the default settings adopted in our method are highlighted with a \colorbox{RedOrange!8}{\textbf{Red Background}}.
\end{itemize}
\endgroup

\subsection{Modular Level Analysis}
As shown in Tab.\ref{tab:module}, we conducted module-level ablation studies across all benchmarks and sequences. Encouragingly, the results remain fully consistent with the conclusions presented in the main text. This demonstrates the exceptional robustness of our method and the synergistic coupling between modules.

\subsection{Component and Hyperparameter Analysis}

Tab.\ref{tab:ablation_all} presents the ablation study demonstrating robustness at both the component and parameter levels, while consistently maintaining superior performance. Notably, under the challenging task scenarios simulated by MedXtreme, our retrieval method achieved a significantly larger performance margin compared to other approaches. This trend aligns with the behaviors observed in dynamic recall on uniformly distributed reference datasets. This suggests that in challenging scenarios or complex clinical settings, retrieval mechanisms with higher quality and finer granularity possess significant efficacy and potential.

\subsection{Dynamic Retrieval Dataset Analysis}
\begin{wrapfigure}{r}{0.7\textwidth}
  \centering

  \vspace{-10pt}

  \includegraphics[width=\linewidth, height=5cm, keepaspectratio]{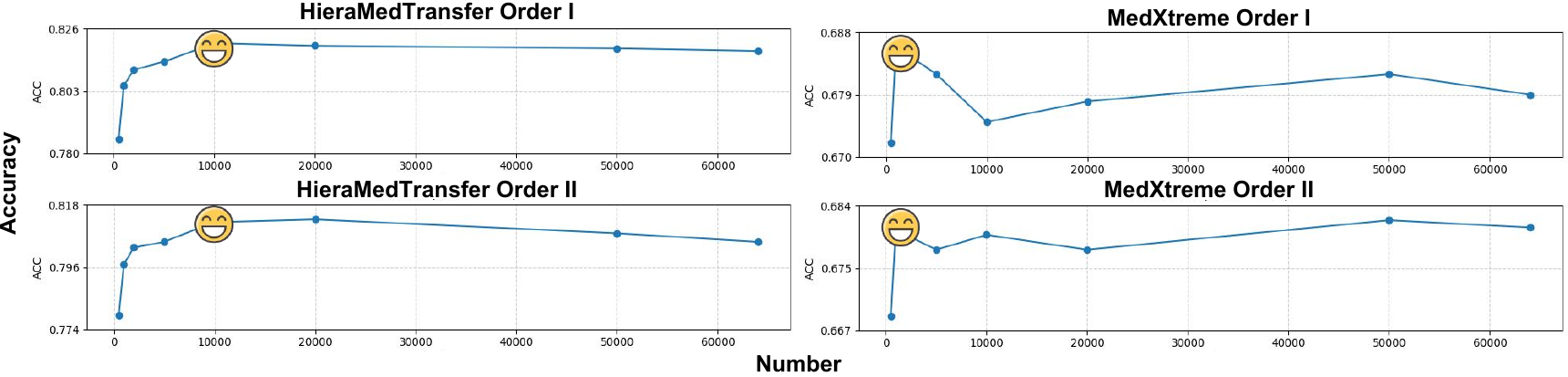}
  
  \caption{Peak Performance of Dynamic Retrieval across Datasets}
  \label{sup_fig5}

  \vspace{-10pt} 
\end{wrapfigure}

The Dynamic Retrieval component is the cornerstone of PRIMED and constitutes the fundamental difference between our method and existing approaches. Two specific aspects warrant further investigation. First, akin to other methods relying on reference datasets, the size of the dataset presents a critical trade-off. Insufficient capacity risks compromising generalization and diversity, while excessive size imposes a prohibitive computational and storage overhead. Since prior studies have demonstrated that performance tends to plateau beyond a certain threshold, our objective is to identify this optimal saturation point illustrated in Fig.\ref{sup_fig5}. Next, building on the determined peak number, we investigated the ratios for dynamic retrieval. Operating under the premise that task weights should exceed domain weights, which in turn should exceed general weights, we employed a grid search to identify the optimal ratios. The quantitative results regarding the dataset capacity saturation are detailed in Tab.~\ref{tab:ablation_RAG_distill_1}, while the outcomes of the grid search for optimal retrieval ratios are tabulated in Tab.~\ref{tab:ablation_RAG_distill_2}.

The experimental results highlight distinct requirements for recall versus generalization across different tasks. Specifically, for intra- and cross-domain transfer tasks such as HieraMedTransfer, enhanced generalization is pivotal for handling diverse scenarios. Conversely, in high-difficulty benchmarks like MedXtreme, models require an extensive, potentially iterative review of representative exemplars. This is an intriguing finding, as it parallels human cognitive processes that combine long-term retention with short-term intensive reinforcement. Indeed, many characteristics of model memory appear to mirror those inherent to human memory.

\begin{table*}[tp]
\renewcommand{\arraystretch}{1}
\small
\centering
\caption{Ablation study on different components and hyperparameters. \colorbox{RedOrange!8}{\textbf{Red Background}} indicates optimal settings.}
\label{tab:ablation_all}
\vspace{-1.5ex}

\begin{tabular}{@{} c c|
                c c c !{\vrule width 1.2pt}
                c c c !{\vrule width 1.2pt}
                c c c !{\vrule width 1.2pt}
                c c c @{}}
\toprule
\multicolumn{2}{c|}{\textbf{Comp./Hparam.}} &
\multicolumn{3}{c!{\vrule width 1.2pt}}{\textbf{HieraMedTransfer I}} &
\multicolumn{3}{c!{\vrule width 1.2pt}}{\textbf{HieraMedTransfer II}} &
\multicolumn{3}{c!{\vrule width 1.2pt}}{\textbf{MedXtreme I}} &
\multicolumn{3}{c}{\textbf{MedXtreme II}} \\

\cmidrule(lr){1-2}
\cmidrule(lr){3-5}
\cmidrule(lr){6-8}
\cmidrule(lr){9-11}
\cmidrule(lr){12-14}

\textbf{Aspect} & \textbf{Detail}
& Transfer & Avg. & Last
& Transfer & Avg. & Last
& ACC & AUC & BWT
& ACC & AUC & BWT \\

\midrule
& Initial CLIP
&  57.9 & 70.7 & 78.2
&  47.7 & 66.9 & 80.1
&  60.5 & 80.7 & -9.5
&  60.3 & 78.9 & -11.2 \\

\rowcolor{RedOrange!8}
\cellcolor{white}Teacher & Last CLIP
&  \textbf{58.3} & \textbf{73.1} & \textbf{82.1}
&  \textbf{48.5} & \textbf{68.0} & \textbf{81.2}
&  \textbf{68.6} & \textbf{87.4} &\textbf{-2.7}
&  \textbf{68.1} & \textbf{86.3} & \textbf{-3.4} \\

& WISE(0.5)
&  58.1 & 71.4 & 79.1
&  48.3 & 67.1 & 80.5
&  63.5 & 83.1 & -5.6
&  62.7 & 82.0 & -8.1 \\
\midrule

& Image-only
&  57.7 & 72.4 & 81.3
&  47.9 & 67.6 & 80.8
&  67.6 & 87.0 & -3.9
&  67.0 & 86.2 & -4.5 \\

KD Loss & Text-only
&  58.2 & 73.0 & 80.5
&  \textbf{49.8} & 67.9 & 80.4
&  65.5 & 85.8 & -6.3
&  63.2 & 85.3 & -9.2 \\

\rowcolor{RedOrange!8}
\cellcolor{white} & Contra.
&  \textbf{58.3} & \textbf{73.1} & \textbf{82.1}
&  48.5 & \textbf{68.0} & \textbf{81.2}
&  \textbf{68.6} & \textbf{87.4} &\textbf{-2.7}
&  \textbf{68.1} & \textbf{86.3} & \textbf{-3.4} \\

\midrule

& $\alpha=0.5$
&  58.3 & 72.8 & 81.3
&  \textbf{48.7} & 67.9 & 80.6
&  66.5 & 86.9 & -5.1
&  66.7 & 85.4 & -4.4 \\

\rowcolor{RedOrange!8}
\cellcolor{white}CKT Scale &$\alpha=1$
&  \textbf{58.3} & \textbf{73.1} & \textbf{82.1}
&  48.5 & \textbf{68.0} & \textbf{81.2}
&  \textbf{68.6} & \textbf{87.4} &\textbf{-2.7}
&  \textbf{68.1} & \textbf{86.3} & \textbf{-3.4} \\

& $\alpha=1.5$
&  58.3 & 73.0 & 81.9
&  48.0 & 67.8 & 81.0
&  66.2 & 87.4 & -4.9
&  66.3 & 85.6 & -5.1 \\
\midrule

& $\beta=0.0$
&  57.2 & 72.8 & \textbf{82.5}
&  46.8 & 67.8 & 80.9
&  66.7 & 87.3 & -4.3
&  65.8 & 86.1 & -5.4 \\

\rowcolor{RedOrange!8}
\cellcolor{white} CMC Scale &$\beta=0.25$
&  \textbf{58.3} & \textbf{73.1} & 82.1
&  \textbf{48.5} & \textbf{68.0} & \textbf{81.2}
&  \textbf{68.6} & \textbf{87.4} &\textbf{-2.7}
&  \textbf{68.1} & \textbf{86.3} & \textbf{-3.4} \\

\cellcolor{white} & $\beta=0.5$
&  58.1 & 72.5 & 81.4
& 48.2 & 67.4 & 77.3
&  66.8 & 85.7 & -4.6
&  64.2 & 84.8 & -7.9 \\

\midrule

& $l_2$
&  57.2 & 72.9 & 81.8
&  48.2 & 67.9 & 81.0
&  66.5 & 86.9 & -4.8
&  64.9 & 86.4 & -6.9 \\

Reg. & EWC
&  57.2 & 71.3 & 82.1
&  48.0 & 66.8 & \textbf{81.4}
&  67.3 & 87.2 & -3.7
&  67.2 & \textbf{86.9} & -3.5 \\

\rowcolor{RedOrange!8}
\cellcolor{white} & DFG
&  \textbf{58.3} & \textbf{73.1} & \textbf{82.1}
&  \textbf{48.5} & \textbf{68.0} & 81.2
&  \textbf{68.6} & \textbf{87.4} &\textbf{-2.7}
&  \textbf{68.1} & 86.3 & \textbf{-3.4} \\

\midrule

& BM25
&  55.1 & 72.2 & 81.8
&  47.5 & 67.5 & 79.4
&  67.4 & 86.8 & -3.4
&  64.8 & 85.5 & -6.5 \\

RAG & Embedding
&  56.1 & 72.5 & 81.5
&  47.3 & 67.2 & 76.7
&  66.3 & 86.9 & -4.8
&  63.9 & 85.4 & -7.5 \\

\rowcolor{RedOrange!8}
\cellcolor{white} & Hierarchical
& \textbf{58.3} & \textbf{73.1} & \textbf{82.1}
&  \textbf{48.5} & \textbf{68.0} & \textbf{81.2}
&  \textbf{68.6} & \textbf{87.4} &\textbf{-2.7}
&  \textbf{68.1} & \textbf{86.3} & \textbf{-3.4} \\

\bottomrule
\end{tabular}
\vspace{-2ex}
\end{table*}

\begin{table}[tp]
    \centering
    \small
    \renewcommand{\arraystretch}{1.1}

    \begin{minipage}[t]{0.49\textwidth}
        \centering
        \caption{Dynamic Retrieval Analysis in HieraMedTransfer.}
        \label{tab:ablation_RAG_distill_1}
        \vspace{-1.5ex}

        \resizebox{\linewidth}{!}{
            \begin{tabular}{@{} c| c c c !{\vrule width 0.6pt} c c c @{}}        
            \toprule
            \textbf{Ratio} &
            \multicolumn{3}{c!{\vrule width 0.6pt}}{\textbf{HieraMedTransfer I}} &
            \multicolumn{3}{c}{\textbf{HieraMedTransfer II }}  \\

            \cmidrule(lr){1-1}
            \cmidrule(lr){2-4}
            \cmidrule(lr){5-7}

            $b:c$& Trans. & Avg. & Last
            & Trans. & Avg. & Last\\
            \midrule
            10:90 &  58.0 & 72.8 & 81.7 &  48.3 & 67.8 & 80.8\\
            20:80 &  58.3 & 73.0 & 82.0 &  48.2 & 67.9 & 81.1 \\
            
            \rowcolor{RedOrange!8}
            30:80 &  \textbf{58.3} & \textbf{73.1} & \textbf{82.1} &  \textbf{48.5} & \textbf{68.0} & \textbf{81.2}\\
            
            40:80 &  58.3 & 73.0 & 82.0 &  48.4 & 67.8 & 79.8 \\
            50:50 &  58.3 & 73.0 & 81.7 &  48.5 & 68.0 & 80.2 \\
            60:40 &  58.2 & 72.9 & 82.1 &  48.0 & 67.9 & 80.8 \\
            80:20 &  58.2 & 73.0 & 81.4 &  48.4 & 68.0 & 81.2 \\
            90:10 &  58.2 & 73.1 & 81.5 &  48.1 & 67.6 & 78.6 \\
            \bottomrule
            \end{tabular}
        }
    \end{minipage}
    \hfill 
    \begin{minipage}[t]{0.49\textwidth}
        \centering
        \caption{Dynamic Retrieval Analysis in MedXtreme.}
        \label{tab:ablation_RAG_distill_2}
        \vspace{-1.5ex}
        
        \resizebox{\linewidth}{!}{
            \begin{tabular}{@{} c| c c c !{\vrule width 0.6pt} c c c @{}}        
            \toprule
            \textbf{Ratio} &
            \multicolumn{3}{c!{\vrule width 0.6pt}}{\textbf{MedXtreme I}} &
            \multicolumn{3}{c}{\textbf{MedXtreme II}}  \\

            \cmidrule(lr){1-1}
            \cmidrule(lr){2-4}
            \cmidrule(lr){5-7}

            $b:c$& ACC & AUC & BWT &  ACC & AUC & BWT\\
            \midrule
            100:100 &  68.4 & 87.0 & \textbf{-2.4} &  67.9 & 85.9 & -3.5\\
            0:100   &  68.2 & 86.5 & -3.2 &  67.7 & 85.7 & -3.9 \\
            70:70   &  68.6 & 86.9 & -2.7 &  68.1 & 85.7 & -3.4 \\
            70:90   &  68.4 & 87.0 & -2.9 &  67.8 & 85.8 & -3.5 \\

            \rowcolor{RedOrange!8}
            70:100  &  \textbf{68.6} & \textbf{87.4} &-2.7 &  \textbf{68.1} & \textbf{86.3} & \textbf{-3.4}\\

            50:100  &  68.5 & 87.3 & -2.8 &  68.1 & 86.3 & -3.4 \\
            50:75   &  68.3 & 87.1 & -3.0 &  67.7 & 86.0 & -4.2 \\
            90:100  &  68.2 & 86.9 & -2.5 &  67.9 & 85.7 & -3.7\\
            \bottomrule
            \end{tabular}
        }
    \end{minipage}
    \vspace{-2ex}
\end{table}

\section{Backbone Selection and Generalizability}
Prior to selecting the specific backbones, we briefly review the list of candidate models, providing a concise overview of these contrastive learning-based foundation models.

\begingroup
\setlist[itemize]{noitemsep, topsep=0pt, left=0pt}
\begin{itemize}
    \item \textbf{BiomedCLIP~\cite{biomedclip}} is a multimodal foundation model pretrained on PMC-15M, a large-scale dataset of 15 million image-text pairs sourced from 4 million scientific articles
     \item \textbf{MMKD-CLIP~\cite{MMKD-CLIP}} is a generalist biomedical foundation model developed via multi-teacher knowledge distillation, utilizing 19.2 million image-text feature pairs synthesized by 9 expert models from the PMC-OA dataset.
     \item \textbf{BMC-CLIP~\cite{biomedica}} is trained on the large-scale BIOMEDICA dataset, which includes over 24M image-text pairs from over 6M open-access scientific articles.
     \item \textbf{UniMed-CLIP~\cite{Unimed-clip}:} is a unified vision-language model trained on UniMed, a large-scale open-source dataset of 5.3 million image-text pairs spanning six imaging modalities (X-ray, CT, MRI, Ultrasound, Pathology, Fundus).
     \item \textbf{CLIP~\cite{CLIP}:} is a multimodal foundation model pretrained on WIT-400M, a large-scale dataset of 400 million image-text pairs collected from a variety of publicly available sources on the internet.

\end{itemize}
\endgroup

\begin{table*}[tp]
\renewcommand{\arraystretch}{1}
\small
\centering
\caption{Experimental Results of Continual Learning on Backbones Other than BiomedCLIP}
\label{tab:backbone_all}
\vspace{-1.5ex}

\resizebox{0.96\linewidth}{!}{

\begin{tabular}{@{} c c|
                c c c !{\vrule width 1.1pt}
                c c c !{\vrule width 1.1pt}
                c c c !{\vrule width 1.1pt}
                c c c @{}}
\toprule
\multicolumn{2}{c|}{\textbf{Architecture}} &
\multicolumn{3}{c!{\vrule width 1.1pt}}{\textbf{HieraMedTransfer I}} &
\multicolumn{3}{c!{\vrule width 1.1pt}}{\textbf{HieraMedTransfer II}} &
\multicolumn{3}{c!{\vrule width 1.1pt}}{\textbf{MedXtreme I}} &
\multicolumn{3}{c}{\textbf{MedXtreme II}} \\

\cmidrule(lr){1-2}
\cmidrule(lr){3-5}
\cmidrule(lr){6-8}
\cmidrule(lr){9-11}
\cmidrule(lr){12-14}

\textbf{Backbone} & \textbf{Method}
& Transfer & Avg. & Last
& Transfer & Avg. & Last
& ACC & AUC & BWT
& ACC & AUC & BWT \\
\midrule
& Continual FT 
& 45.8 & 66.9 & 79.4
& 45.7 & 66.8 & 79.5
& 65.7& 87.5 & -7.1
& 64.5 & 85.7 & -8.6  \\

& WiSE-FT~\cite{wise_ft}
& 46.4 & 67.1 & 80.2
& 46.4 & 67.1 & 80.0
& 64.8 & 87.3 & -4.0
& 62.4 & 86.0 & -6.9  \\

\cellcolor{white}MMKD~\cite{MMKD-CLIP} & ZSCL~\cite{zscl}
& \textbf{58.6} & 69.3 & 77.8
& 44.8 & 65.0 & 79.8
& 57.4 & 81.4 & -8.6
& 56.0 & 80.9 & -10.4  \\

& GIFT~\cite{GIFT}
& 49.6 & 68.3 & 80.5
& 49.6 & 68.3 & 80.4
& 69.7 & 88.2 & -1.8
& 68.4 & 88.1 & -3.1 \\

\rowcolor{RedOrange!8}
\cellcolor{white}& $\text{PRIMED}_{dyn}$
& 50.7 & \textbf{69.4} & \textbf{80.7}
& \textbf{50.7} & \textbf{69.4} & \textbf{80.6}
& \textbf{70.4} & \textbf{88.3} & \textbf{-1.5}
& \textbf{70.5} & \textbf{88.1} & \textbf{-1.3}  \\

\midrule
& Continual FT 
& 45.0 & 66.4 & 79.8
& 44.9 & 66.4 & 80.0
& 61.0 & 83.4 & -10.5
& 58.8 & 82.6 & -13.7  \\

& WiSE-FT~\cite{wise_ft}
& 44.0 & 66.5 & 81.5
& 44.0 & 66.5 & 81.4
& 57.9 & 83.2 & -9.1
& 61.6 & 86.5 & -5.4 \\

\cellcolor{white}UniMed~\cite{Unimed-clip} & ZSCL~\cite{zscl}
& 44.9 & 66.5 & 81.4
& 42.5 & 63.8 & 81.9
& 63.2 & 85.0 & -6.3
& 59.9 & 83.6 & -10.2 \\

& GIFT~\cite{GIFT}
& 44.2 & 66.7 & 82.5
& 44.2 & 66.7 & 82.4
& 67.1 & \textbf{88.7} & -3.6
& 67.0 & 87.6 & -3.5 \\

\rowcolor{RedOrange!8}
\cellcolor{white}& $\text{PRIMED}_{dyn}$
& \textbf{45.0} & \textbf{67.1} & \textbf{83.1}
& \textbf{45.1} & \textbf{67.1} & \textbf{83.2}
& \textbf{67.4} & 88.4 & -3.3
& \textbf{67.2} & \textbf{88.3} & \textbf{-3.3} \\

\midrule
& Continual FT 
& 34.9 & 59.1 & 71.1
& 34.2 & 58.6 & 54.4
& 61.5 & 83.7 & -17.2
& 47.3 & 82.5 & -34.1 \\

& WiSE-FT~\cite{wise_ft}
& 34.8 & 60.3 & 70.7
& 34.9 & 61.6 & 72.6
& 64.2 & 85.1 & -12.4
& 57.3 & 84.9 & -20.8 \\

\cellcolor{white}CLIP~\cite{CLIP} & ZSCL~\cite{zscl}
& 35.7 & 61.1 & 78.0
& 35.5 & 61.0 & 78.5
& 63.0 & 82.3 & -11.0
& 56.2 & 82.9 & -19.2 \\

& GIFT~\cite{GIFT}
& \textbf{36.2} & 61.7 & 73.9
& \textbf{36.2} & 62.6 & 75.5
& 71.0 & 87.5 & -6.0
& 70.9 & 88.7 & -5.9 \\

\rowcolor{RedOrange!8}
\cellcolor{white}& $\text{PRIMED}_{dyn}$
& 35.7 & \textbf{62.6} & \textbf{84.6}
& 35.6 & \textbf{62.6} & \textbf{84.0}
& \textbf{73.8} & \textbf{89.1} & \textbf{-2.6}
& \textbf{73.7} & \textbf{89.4} & \textbf{-2.8} \\

\bottomrule
\end{tabular}
}
\vspace{-1ex}
\end{table*}

\subsection{Justification for the BiomedCLIP Backbone}
Our choice of BiomedCLIP as the primary backbone is motivated by several factors, outlined below in descending order of significance. Crucially, we must underscore that this selection was not predicated solely on performance metrics. Indeed, considerations such as the guarantee against data contamination and the maturity of the architectural framework took precedence over raw performance.

\begin{wrapfigure}{r}{0.5\textwidth}
  \centering
  \vspace{-20pt}
  
  \includegraphics[width=0.85\linewidth, height=5cm, keepaspectratio]{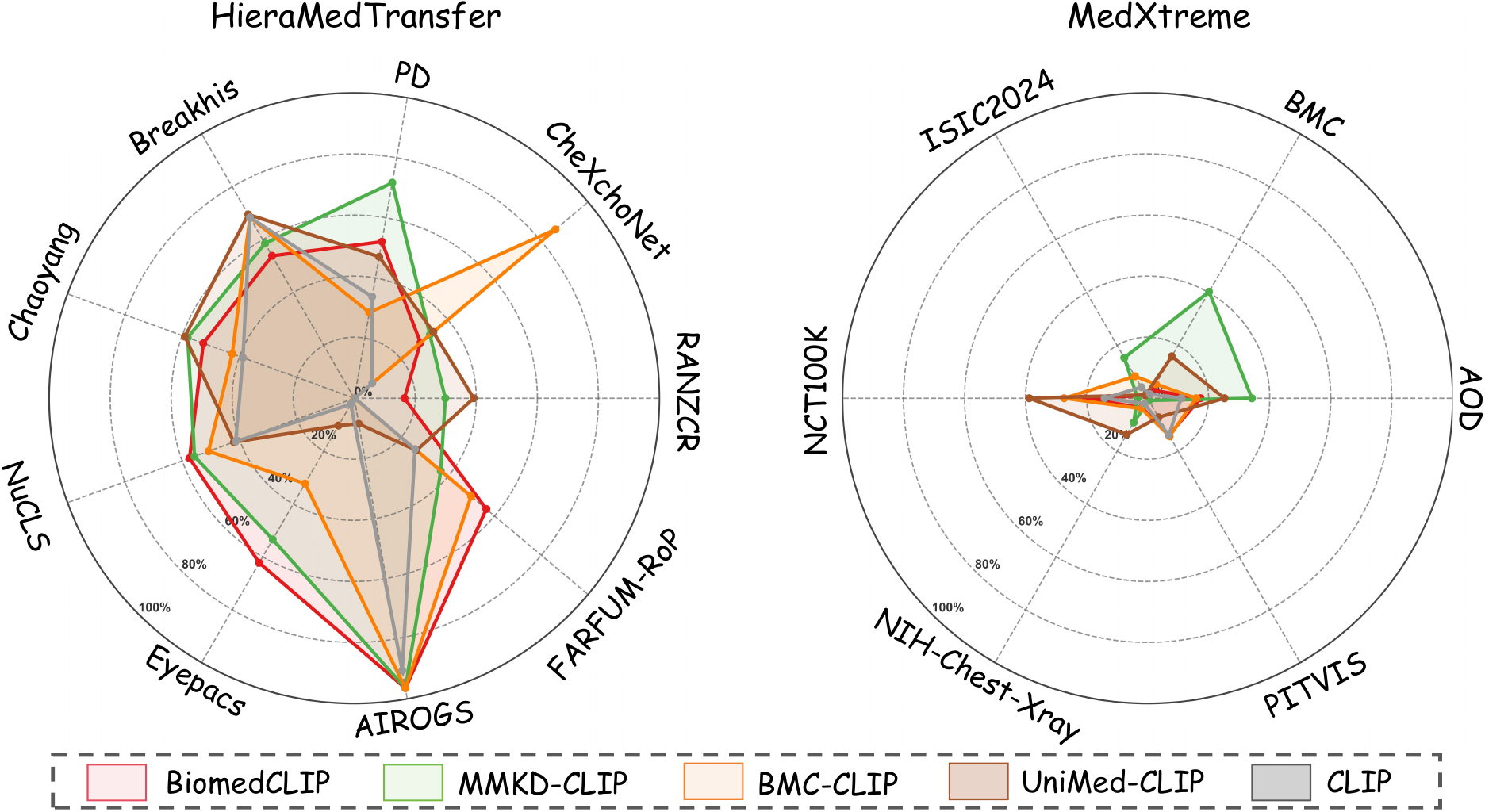}
  \vspace{-3pt}
  \caption{The zero-shot capabilities of 5 backbones across HieraMedTransfer and MedXtreme are depicted in radar chart format.}
  \label{sup_fig3}
  \vspace{-13pt}
\end{wrapfigure}

\vspace{2pt}\noindent\textbf{Data Security.} Admittedly, while all the aforementioned methods utilize open-source datasets, only BiomedCLIP and BMC-CLIP feature a comprehensive data acquisition architecture. This distinction fundamentally ensures data integrity and reproducibility while preventing data contamination. Although MMKD-CLIP exhibits impressive experimental performance, it is derived from multi-model distillation, making it difficult to fully verify its data provenance. Therefore, establishing a more controllable baseline is of paramount importance to our work.

\vspace{2pt}\noindent\textbf{Zero-shot performance.} Fig.\ref{sup_fig3} presents the zero-shot results of various models on the HieraMedTransfer and MedXtreme benchmarks. For HieraMedTransfer, it is essential that the backbone exhibits a reasonable baseline of zero-shot capability; otherwise, the subsequent transferability evaluation would lack validity. BiomedCLIP demonstrates consistent performance across all datasets without exhibiting anomalous outlier peaks, establishing it as a highly robust and reliable candidate. In contrast, all models yield suboptimal performance on MedXtreme. Consequently, absolute performance metrics are of secondary importance compared to the potential risk of data contamination. In this regard, BiomedCLIP serves as a good choice.

\vspace{2pt}\noindent\textbf{Model Architecture.} We favored a mature architecture that fits our specific demands; specifically, BiomedCLIP features a well-developed fine-tuning and post-training ecosystem. Additionally, we aimed to maximize experimental comparability by aligning with the ViT-B configuration used in natural image studies (e.g., ZSCL). Therefore, absent any distinct performance benefits, the ViT-L versions of BMC-CLIP and UniMed-CLIP were not selected as backbones for the main experiments.

\subsection{Generalizable SOTA Performance}
Although we consider BiomedCLIP to be the most intuitively suitable backbone, we also evaluated other ViT-B based backbones, as shown in Tab.\ref{tab:backbone_all}. Our method consistently achieved superior performance, demonstrating the effectiveness of our proposed strategy.

\begin{figure*}[tp]
\centering
 \includegraphics[width=1\linewidth]{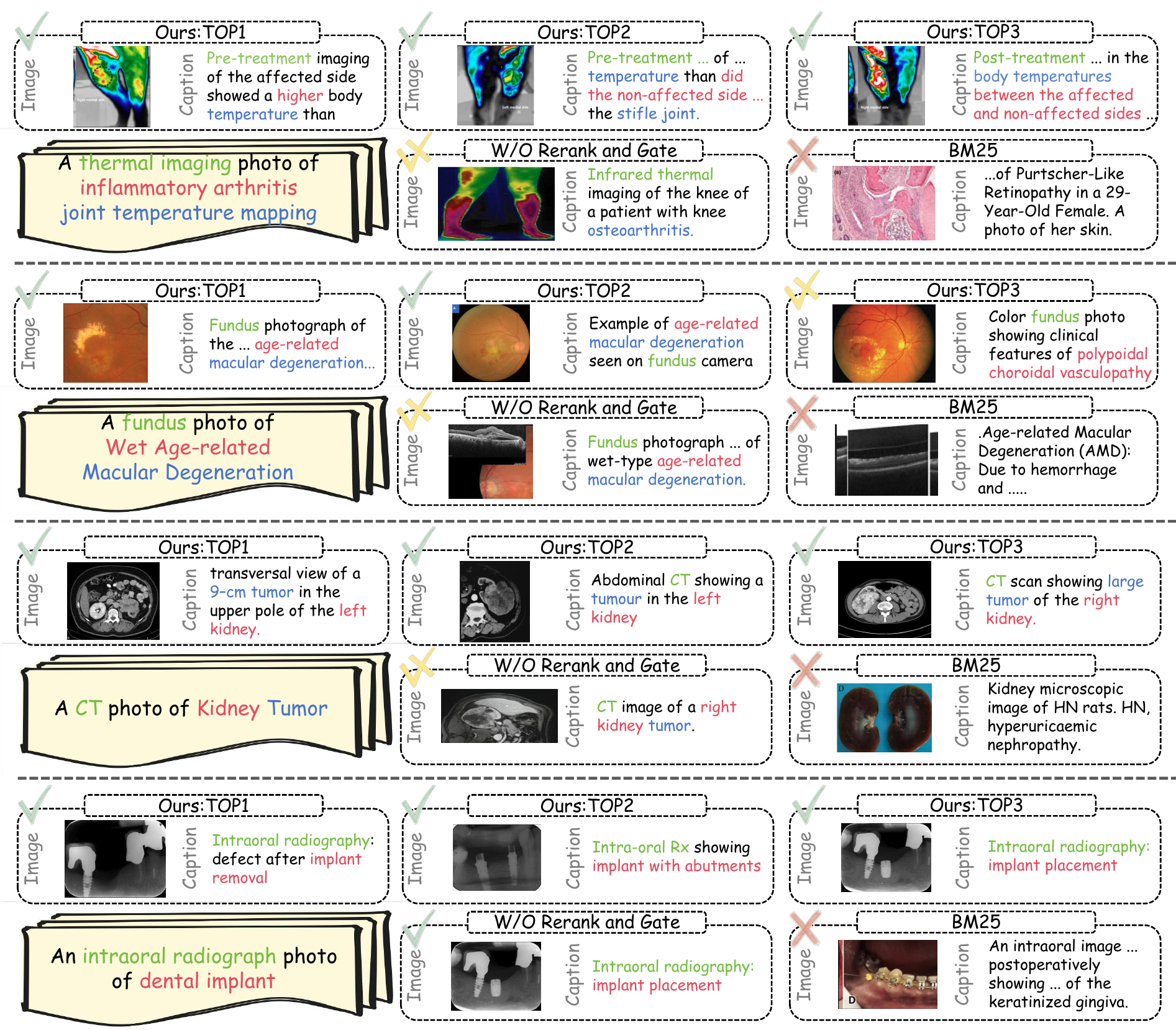}
\caption{Qualitative comparison with state-of-the-art methods. Our method achieves superior performance in visual correction and textual precision. By explicitly aligning the hierarchical content within questions with the retrieved data, our method achieves optimal fine-grained retrieval performance. Furthermore, leveraging visual-level retrieval capabilities allows our approach to prioritize complete and high-quality images rather than relying solely on textual cues. This capability enhances the dynamic retrieval database's distillation.}
\label{sup_fig4}
\end{figure*}

\section{Retrieval Visualization}

Echoing the main text, we underscore the distinct superiority of our retrieval approach in terms of intuitive visualization. Primarily, the intrinsic mechanism of multimodal retrieval endows our method with strong semantic disentanglement capabilities. A prime example is in dentistry, where our model clearly discriminates between OCT scans and natural images, despite their high textual semantic overlap. Furthermore, our approach exhibits enhanced recall precision, moving beyond the rigid constraints of keyword matching. Since our data source relies heavily on multi-subgraph disentanglement, as detailed above, we effectively filter out cases of failed disentanglement or conceptual ambiguity—providing a robust guarantee of effectiveness. Ultimately, we are encouraged to observe that the retrieved content demonstrates both generalizability and hierarchical progression. By moving beyond isolated disease categories to account for the holistic connections between diseases, lesions, and subtypes, we believe this property is pivotal in improving model memorization.

\end{document}